%% file: 0_arxiv_neurips_main.tex
\documentclass{article}

\usepackage[preprint,nonatbib]{neurips_2022}

\usepackage[utf8]{inputenc} 
\usepackage[T1]{fontenc}    
\usepackage{hyperref}       
\usepackage{url}            
\usepackage{booktabs}       
\usepackage{amsfonts}       
\usepackage{nicefrac}       
\usepackage{microtype}      
\usepackage{xcolor}         

\input{\includehome/supp_packages}
\input{\includehome/supp_macros}

\def\mytitle{Certifiably Robust Policy Learning against Adversarial Communication in Multi-agent Systems}

\usepackage{amsmath}
\usepackage{amssymb}
\usepackage{mathtools}
\usepackage{amsthm}

\usepackage{microtype}
\usepackage{graphicx}
\usepackage{booktabs} 
\usepackage{algorithm}
\usepackage{algorithmic}
\newtheorem{theorem}{Theorem}[section]

\newtheorem{corollary}[theorem]{Corollary}
\newtheorem{definition}[theorem]{Definition}
\newtheorem{assumption}[theorem]{Assumption}

\newtheorem{condition}[theorem]{Condition}

\title{\mytitle}

\usepackage{authblk}

\author{%
  Yanchao Sun${}^\dag$\thanks{Part of this work was done when the first author was an intern at JPMorgan AI Research. } 
  \quad
  Ruijie Zheng${}^\dag$
  \quad
  Parisa Hassanzadeh${}^\ddag$
  \quad
  Yongyuan Liang${}^\S$
  \quad
  Soheil Feizi${}^\dag$
  \quad
  Sumitra Ganesh${}^\ddag$
  \quad
  Furong Huang${}^\dag$
  \\
  ${}^\dag$ University of Maryland, College Park \quad \{ycs, rzheng12, sfeizi, furongh\}@umd.edu \\
  ${}^\ddag$JPMorgan AI Research \quad \{parisa.hassanzadeh, sumitra.ganesh\}@jpmchase.com \\
  ${}^\S$Sun Yat-sen University \quad liangyy58@mail2.sysu.edu.cn
}

\begin{document}

\maketitle

\begin{abstract}
Communication is important in many multi-agent reinforcement learning (MARL) problems for agents to share information and make good decisions. However, when deploying trained communicative agents in a real-world application where noise and potential attackers exist, the safety of communication-based policies becomes a severe issue that is underexplored. Specifically, if communication messages are manipulated by malicious attackers, agents relying on untrustworthy communication may take unsafe actions that lead to catastrophic consequences. Therefore, it is crucial to ensure that agents will not be misled by corrupted communication, while still benefiting from benign communication. In this work, we consider an environment with $N$ agents, where the attacker may arbitrarily change the communication from any $C<\frac{N-1}{2}$ agents to a victim agent. For this strong threat model, we propose a certifiable defense by constructing a message-ensemble policy that aggregates multiple randomly ablated message sets. Theoretical analysis shows that this message-ensemble policy can utilize benign communication while being certifiably robust to adversarial communication, regardless of the attacking algorithm. Experiments in multiple environments verify that our defense significantly improves the robustness of trained policies against various types of attacks.
\end{abstract}

\input{\arxivtexhome/1_intro}

\input{\texhome/3_setup_new}

\input{\arxivtexhome/4_algo_long}
\input{\arxivtexhome/2_relate}

\input{\texhome/5_exp}

\input{\arxivtexhome/6_discuss}

\section*{Acknowledgements}
This work is supported by JPMorgan Chase \& Co., National Science Foundation NSF-IIS-FAI Award 2147276, Office of Naval Research, Defense Advanced Research Projects Agency Guaranteeing AI Robustness against Deception (GARD), and Adobe, Capital One and JP Morgan faculty fellowships.

\paragraph{Disclaimer}
This paper was prepared for informational purposes in part by
the Artificial Intelligence Research group of JPMorgan Chase \& Co\. and its affiliates (``JP Morgan''),
and is not a product of the Research Department of JP Morgan.
JP Morgan makes no representation and warranty whatsoever and disclaims all liability,
for the completeness, accuracy or reliability of the information contained herein.
This document is not intended as investment research or investment advice, or a recommendation,
offer or solicitation for the purchase or sale of any security, financial instrument, financial product or service,
or to be used in any way for evaluating the merits of participating in any transaction,
and shall not constitute a solicitation under any jurisdiction or to any person,
if such solicitation under such jurisdiction or to such person would be unlawful.

\bibliographystyle{plain}

\input{0_arxiv_neurips_main.bbl}

\newpage
\appendix

{\centering{\Large Appendix}}


\input{\arxivtexhome/a2_theory}
\input{\texhome/a3_proof}
\newpage
\input{\texhome/a4_exp}
\input{a4_hyper}
\input{a5_detect}
\input{\texhome/a8_impact}

\end{document}

%% file: include/supp_packages.tex
\usepackage{url}
\usepackage{graphicx}
\usepackage{bbm}
\usepackage{xcolor}
\usepackage{scalerel}
\usepackage[outdir=./]{epstopdf}
\usepackage{graphicx,float,pgfplots,wrapfig,sidecap,lipsum}

\usepackage{colortbl}
\usepackage{tablefootnote}
\usepackage[font=small,labelfont=bf]{caption}
\usepackage{subcaption}
\usepackage{subfloat}
\usepackage{tikz}
\usetikzlibrary{fit}
\usetikzlibrary{calc,shapes}
\usetikzlibrary{decorations.pathmorphing} 
\usetikzlibrary{fit}					
\usetikzlibrary{backgrounds}	
\usetikzlibrary{pgfplots.groupplots}

\usepackage[utf8]{inputenc}
\usepackage{pgfplots}
\DeclareUnicodeCharacter{2212}{−}
\usepgfplotslibrary{groupplots,dateplot}
\usetikzlibrary{patterns,shapes.arrows}
\pgfplotsset{compat=newest}

\usepackage{xspace}
\usepackage{soul}

%% file: include/supp_macros.tex
\newcommand{\funmed}{\mathsf{Median}}
\newcommand{\commfunc}{\xi}

\newcommand{\agents}{\mathcal{D}}
\newcommand{\his}{\tau}
\newcommand{\hisspace}{\Gamma}

\newcommand{\step}{^{(t)}}

\newcommand{\ablpolicy}{\hat{\pi}}
\newcommand{\smpolicy}{\widetilde{\pi}}

\newcommand{\safeaction}{\mathcal{A}_{\mathrm{benign}}}

\newcommand{\safeactioncont}{\mathsf{Range}(\safeaction)}

\newcommand{\mtoi}{\mathbf{m}_{: \to i}}
\newcommand{\msg}{\mathbf{m}}
\newcommand{\msgk}{[\mathbf{m}]_k}
\newcommand{\allmsgk}{\mathcal{H}_{k}(\msg)}
\newcommand{\allmsgkd}{\mathcal{H}_{k,D}(\msg)}
\newcommand{\msgki}{[\mtoi]_k}
\newcommand{\allmsgki}{\mathcal{H}_{k}(\mtoi)}

\newcommand{\goodmsg}{\mathbf{m}_{\mathrm{benign}}}
\newcommand{\advmsg}{\mathbf{m}_{\mathrm{adv}}}
\newcommand{\goodmsgk}{[\mathbf{m}_{\mathrm{benign}}]_k}

\newcommand{\goodallmsgk}{\mathcal{H}_{k}(\goodmsg)}

\newcommand{\oursfull}{{Ablated Message Ensemble}\xspace}
\newcommand{\ours}{{AME}\xspace}
\newcommand{\ablname}{{message-ablation policy}\xspace}
\newcommand{\ablnames}{{message-ablation policies}\xspace}
\newcommand{\ensname}{{message-ensemble policy}\xspace}
\newcommand{\ablnamecap}{{Message-Ablation Policy}\xspace}
\newcommand{\ensnamecap}{{Message-Ensemble Policy}\xspace}
\newcommand{\safename}{{benign action}\xspace}
\newcommand{\ksample}{{k-sample}\xspace}
\newcommand{\ksamples}{{k-samples}\xspace}

\newcommand{\permute}{{Perm-Attack}\xspace}
\newcommand{\swap}{{Swap-Attack}\xspace}
\newcommand{\flip}{{Flip-Attack}\xspace}

\newcommand{\marketenv}{{InventoryManager}\xspace}

\definecolor{darkpastelgreen}{rgb}{0.01, 0.75, 0.24}

%% file: arxiv/1_intro.tex
\section{Introduction}
\label{sec:intro}
Neural network-based multi-agent reinforcement learning (MARL) has achieved significant advances in many real-world applications, such as autonomous driving~\cite{shalev2016safe,sallab2017deep}. In a multi-agent system, especially in a cooperative game, communication usually plays an important role. By feeding communication messages as additional inputs to the policy network, each agent can obtain more information about the environment and other agents, and thus can learn a better policy~\cite{foerster2016learning,hausknecht2016cooperation,sukhbaatar2016learning}. 

\begin{figure*}[!t]
    \centering
     \begin{subfigure}[b]{0.44\textwidth}
         \centering
         \includegraphics[width=0.9\textwidth]{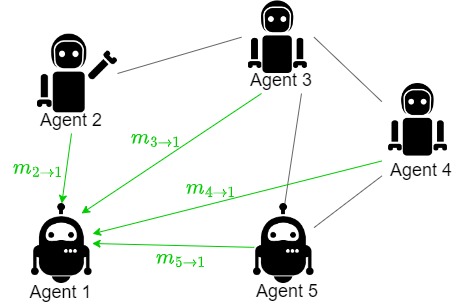}
         \vspace{-0.5em}
         \caption{Training time}
         \label{sfig:train}
     \end{subfigure}
     \hfill
     \begin{subfigure}[b]{0.44\textwidth}
         \centering
         \includegraphics[width=0.9\textwidth]{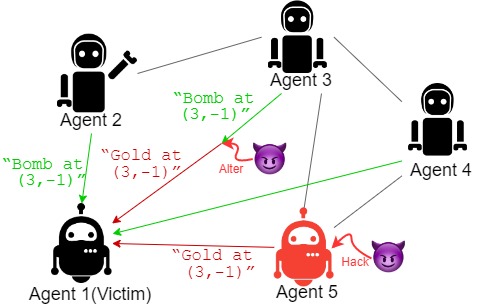}
         \vspace{-0.5em}
         \caption{Test time (in deployment)}
         \label{sfig:test}
     \end{subfigure}
     \vspace{-0.5em}
    \caption{An example of test-time communication attacks in a communicative MARL system.  (a) During training, agents are trained collaboratively in a safe environment, such as a simulated environment. (b) In deployment, agents execute pre-trained policies in the real world, where malicious attackers may modify the benign ({\color{darkpastelgreen}{green}}) messages into adversarial ({\color{red}{red}}) signals to mislead some victim agent(s). 
    }
    \label{fig:diagram}
\end{figure*}

However, neural networks are shown to be vulnerable to adversarial attacks~\cite{chakraborty2018adversarial}, i.e.,
a well-trained network may output a wrong answer if the input to the network is slightly perturbed~\cite{goodfellow2014explaining}.
As a result, although a policy can obtain a high reward by taking in the messages sent by other agents, it may also get drastically misled by inaccurate information or even adversarial communication from malicious attackers. 
For example, when several drones execute pre-trained policies and exchange information via wireless communication, it is possible that messages get noisy in a hostile environment, or even some malicious attacker eavesdrops on their communication and intentionally perturbs some messages to a victim agent via cyber attacks. 
Moreover, even if the communication channel is protected by advanced encryption algorithms, an attacker may also hack some agents and alter the messages before they are sent out (e.g. hacking IoT devices that usually lack sufficient protection~\cite{naik2017cyber}). Figure~\ref{fig:diagram} shows an example of communication attacks, where the agents are trained with benign communication, but attackers may perturb the communication during the test time. The attacker may lure a well-trained agent to a dangerous location through malicious message propagation and cause fatal damage. 
Although our paper focuses on adversarial perturbations of the communication messages, it also includes unintentional perturbations, such as misinformation due to malfunctioning sensors or communication failures; these natural perturbations are no worse than adversarial attacks.

Although adversarial attacks and defenses have been extensively studied in supervised learning~\cite{madry2018towards,cohen2019certified,zhang2019theoretically} and reinforcement learning~\cite{zhang2020robust,zhang2021robust,oikarinen2020robust}, there has been little discussion on the robustness issue against adversarial communication in MARL problems. 
Therefore, to safely improve decision-making with communication in high-stakes multi-agent scenarios, it is crucial to robustify MARL policies against adversarial communication.  
Achieving high performance in a multi-agent cooperative game through inter-agent communication while being robust to adversarial communication is a challenging problem due to the following reasons. 
\textbf{Challenge I}: 
Sometimes communication attacks are stealthier as misleading
messages  (which share the form of benign messages and are semantically meaningful in Figure~\ref{sfig:test}) can be hard to recognize, while attacking by abnormal actions~\cite{gleave2020adversarial} can be easier for humans to notice.
\textbf{Challenge II}: A strong attacker can even be \textit{adaptive} to the victim agent and significantly reduce the victim's total reward. For instance, for a victim agent who moves according to the average of GPS coordinates sent by others, the attacker may learn to send extreme coordinates to influence the average.
\textbf{Challenge III}: There can be more than one attacker (or an attacker can perturb more than one message at one step), such that they can collaborate to mislead a victim agent.

Tu et al.~\cite{tu2021adversarial} apply adversarial training~\cite{madry2018towards} to improve model robustness against communication perturbations. But they focus on perturbations with small $\ell_p$ distances, which do not cover many attack scenarios as discussed in Section~\ref{sec:setup_adv}.
Some recent works~\cite{blumenkamp2020emergence,xue2021mis,mitchell2020gaussian} take the first step to investigate adversarial communications in MARL and propose several defending methods, such as a learned message filter~\cite{xue2021mis}. However, these empirical defenses do not fully address the aforementioned challenges, and are not guaranteed to be robust, especially under adaptive attacks. 

In this paper, we address all aforementioned challenges by proposing {a certifiable defense, named \textbf{\oursfull (\ours)}, that can guarantee the performance of agents when a fraction of communication messages are perturbed}. 
The main idea of \ours is to make decisions based on multiple different subsets of communication messages (i.e., ablated messages). Specifically, for a list of messages coming from different agents, we train a \emph{\ablname} that takes in a subset of messages and outputs a \emph{base action}. Then, we construct an \emph{\ensname} by aggregating multiple base actions coming from multiple ablated message subsets. 
For a discrete action space, the ensemble policy takes the majority of the multiple base actions obtained, while for a continuous action space, the ensemble policy takes the median of these base actions. \ours is a generic framework that works for any agent that observes multiple messages from different sources, as long as the agent receives more benign communication messages than adversarial ones. We also show in Section~\ref{sec:algo_more} that \ours can be used in many other decision-making scenarios beyond defending against communication attacks. 
A similar randomized ablation idea is used by Levine et al.~\cite{Levine_Feizi_2020} to defend against $\ell_0$ attacks in image classification. However, they provide high-probability guarantee for a single-step decision, which is not suitable for sequential decision-making problems, as the guaranteed probability decreases when it propagates over timesteps. Moreover, their algorithm does not work when the model has a continuous output space. 
In contrast, \ours has robustness guarantees for both the immediate action and the long-term reward, and for both discrete and continuous actions. 


Our contributions can be summarized as below:\\
\textbf{(1)} We formulate the problem of adversarial attacks and defenses in communicative MARL (CMARL). \\
\textbf{(2)} We propose a novel defense method, \ours, 
that is certifiably robust against arbitrary perturbations of up to $C<\frac{N-1}{2}$ communications, where $N$ is the number of agents. \\
\textbf{(3)} Experiment in several multi-agent environments shows that \ours obtains significantly higher reward than baseline methods under both non-adaptive and adaptive attackers. 

%% file: neurips/3_setup_new.tex
\section{Communicative Muti-agent Reinforcement Learning (CMARL)}
\label{sec:setup_cmarl}
\textbf{Dec-POMDP Model}\quad
We consider a Decentralised Partially Observable Markov Decision Process (Dec-POMDP)~\cite{oliehoek2012decentralized,oliehoek2015concise,das2020tarmac} which is a multi-agent generalization of the single-agent POMDP models. 
A Dec-POMDP can be modeled as a tuple $\langle \mathcal{D}, \mathcal{S}, \mathcal{A}_{\agents}, \mathcal{O}_{\agents}, O, P, R, \gamma, \rho_0 \rangle$. 
$\mathcal{D}=\{1,\cdots,N\}$ is the set of $N$ agents.
$\mathcal{S}$ is the underlying state space. $\mathcal{A}_{\agents}=\times_{i\in\mathcal{D}}\mathcal{A}_i$ is the \emph{joint} action space. $\mathcal{O}_{\agents}=\times_{i\in\mathcal{D}}\mathcal{O}_i$ is the \emph{joint} observation space, with $O$ being the observation emission function.
$P: \mathcal{S}\times\mathcal{A}_1\times\cdots\mathcal{A}_N \to \Delta(\mathcal{S}) $ is the state transition function\footnote{$\Delta(\mathcal{X})$ denotes the space of probability distributions over space $\mathcal{X}$.}.
$R: \mathcal{S}\times\mathcal{A}_1\times\cdots\mathcal{A}_N \to \mathbb{R}$ is the shared reward function.
$\gamma$ is the shared discount factor, and $\rho_0$ is the initial state distribution. 

\textbf{Communication Policy $\commfunc$ in Dec-POMDP}\quad
Due to the partial observability, communication among agents is crucial for them to obtain high rewards. Consider a shared message space $\mathcal{M}$, where a message $m\in\mathcal{M}$ can be a scalar or a vector, e.g., signal of GPS coordinates. 
The communication policy of agent $i\in\mathcal{D}$, denoted as $\commfunc_i$, generates messages based on the agent's observation or interaction history. 
At every step $t$, agent $i$ sends a communication message $m_{i \to j}\step$ to agent $j$ for all $j\neq i$. 
(For notational simplicity, we will later omit ${}\step$ when there is no ambiguity.)
We assume that agents are fully connected for communication, although our algorithm directly applies to partially connected communication graphs. That is, at each step, every agent sends $N-1$ messages (one for each agent other than itself), and receives $N-1$ messages from others. \\
This paper studies a general defense under any given communication protocol, so we do not make any assumption on how the communication policy $\commfunc$ is obtained (e.g., from a pre-defined communication protocol, or a learning algorithm~\cite{foerster2016learning,das2020tarmac}). 

\textbf{Acting Policy $\pi$ with Communication}\quad
The goal of each agent $i\in\agents$ is to maximize the discounted cumulative reward $\sum_{t=0}^\infty \gamma^t r\step$ by learning an acting policy $\pi_i$. 
When there exists communication, the policy input contains both its own interaction history, denoted by $\his_i\in\hisspace_i$, and the communication messages $\mtoi:=\{m_{j \to i} | 1\leq j \leq N, j\neq i \}$. 
Similar to the communication policy $\commfunc$, we do not make any assumption on how the acting policy $\pi$ is formulated, as our defense mechanism introduced later can be plugged into any policy learning procedure.


\section{Problem Formulation: Communication Attacks in Deployment of CMARL}
\label{sec:setup_adv}

Communication is important for agents to obtain high rewards, but it can be a double-edged sword --- it benefits decision making but may make agents vulnerable to perturbations of messages.
Communication attacks in MARL has recently attracted increasing attention~\cite{blumenkamp2020emergence,tu2021adversarial,xue2021mis} with different focuses, as summarized in Section~\ref{sec:relate}. 
In this paper, we consider a practical and strong threat model where malicious attackers can arbitrarily perturb a subset of communication messages during test time.

\textbf{Formulation of Communication Attack: The Threat Model}\quad 
During test time, agents execute well-trained policies $\pi_1,\cdots, \pi_N$. 
As shown in Figure~\ref{sfig:test}, the attacker may perturb communication messages to mislead a specific victim agent. Without loss of generality, suppose $i\in\agents$ is the victim agent receiving $N-1$ communication messages from other agents. We consider the sparse attack setup where up to $C$ messages can be arbitrarily perturbed at every step, among all $N-1$ messages. 
Here $C$ is a reflection of the attacker's attacking power.
The victim agent has no knowledge of which messages are adversarial. 
We make the following mild assumption for the attacking power.
\begin{assumption} [Attacking Power]
\label{assum:num}
An attacker can arbitrarily manipulate fewer than a half of the communication messages, i.e.,
$C<\frac{N-1}{2}$.
\end{assumption}
This is a realistic assumption, as it takes the attacker's resources to hack or control each communication channel. It is less likely that an attacker can change the majority of communications among agents without being detected. Moreover, communications can be corrupted due to hardware failures, which usually affect a limited fraction of communications.
Note that this is a strong threat model as the $C$ attacked messages can be arbitrarily perturbed based on the attacker's attacking algorithm.

\textbf{Comparison with $\ell_p$ Threat Models}\quad 
Many existing works on adversarial attack and defense~\cite{goodfellow2014explaining,huang2017adversarial,zhang2020robust,tu2021adversarial} assume that the perturbation is small in $\ell_p$ norm. However, many realistic and stealthy attacks can not be covered by the $\ell_p$ threat model. For example, the attacker may replace a word in a sentence as in Figure~\ref{sfig:test}, add a patch to an image, or shift the signal by some bits. In these cases, the $\ell_p$ distance between the clean value and the perturbed value is large, such that $\ell_p$ defenses do not work.
In contrast, these attacks are covered by our threat model which allows arbitrary perturbations to $C$ messages. Therefore, our setting can work for broader applications. 

\textbf{Comparison between Communication Attacks and Other Attacks in MARL}\quad
Adversarial attacks and defenses in RL systems have recently attracted more and more attention, and are considered in many different scenarios. 
The majority of related work focuses on \textit{directly attacking a victim} by perturbing its observations~\cite{huang2017adversarial,gleave2020adversarial,oikarinen2020robust,sun2021strongest} or actions~\cite{tessler2019action,pinto2017robust}. 
However, an attacker may not have direct access to the specific victim's observation or action. 
In this case, \textit{indirect attacks via other agents} can be an alternative. For example, Gleave et al.~\cite{gleave2020adversarial} propose to attack the victim by changing the other agent's actions. Therefore, even if the victim agent has well-protected sensors, the attacker can still influence it by manipulating other under-protected agents. But the intermediary agent whose actions are altered will obtain sub-optimal reward, which makes the attack noticeable and less stealthy. 
In contrast, if an attacker alters the communication messages sent from the other agents (e.g., by man-in-the-middle attacks~\cite{mallik2019man}), the behaviors of other agents are not changed, and thus it is relatively hard to find who has sent adversarial messages and which messages are not trustworthy. \\
It is also worth pointing out that, since an acting policy $\pi$ takes in both its own observation and the communication messages, communication can be regarded as a subset of policy inputs. Therefore, the communication attack is related to traditional adversarial defenses of deep neural policies against $\ell_0$ perturbations to the input~\cite{Levine_Feizi_2020}, as detailed in Appendix~\ref{app:theory}.

\textbf{Goal: A General and Certifiable Defense}\quad
As formulated above, the attacker can perturb up to $C$ communication messages sent to any agent at every step. 
However, we do not make any assumption on what attack algorithm the attacker uses, i.e., how a message is perturbed. 
In practice, an attacker may randomly change the communication signal, or learn a function to perturb the communication based on the current situation. 
The attacker can either be white-box or black-box, based on whether it knows the victim agent's policy, as extensively studied in the field of adversarial supervised learning~\cite{chakraborty2018adversarial}. 
To achieve generic robustness under a wide range of practical scenarios, it is highly desirable for a defense to be agnostic to attack algorithms and work for arbitrary (including strongest/worst) perturbations. This is achieved by our defense which will be introduced in Section~\ref{sec:algo}.

%% file: arxiv/4_algo_long.tex
\section{Provably Robust Defense for CMARL}
\label{sec:algo}

In this section, we present our defense algorithm, \textit{\oursfull (\ours)}, against test-time communication attacks in CMARL. 
We first introduce the algorithm design in Section~\ref{sec:algo_overview}, then present the theoretical analysis in Section~\ref{sec:algo_theory}.
Section~\ref{sec:algo_more} discusses several extensions of \ours.

\subsection{\oursfull (\ours)}
\label{sec:algo_overview}


%
Our goal is to learn and execute a robust policy for any agent in the environment, so that the agent can perform well in both a non-adversarial environment and an adversarial environment. 
To ease the illustration, we focus on robustifying an arbitrary agent $i\in\agents$, and the same defense is applicable to all other agents.
We omit the agent subscript ${}_i$, and denote the agent's observation space, action space, and interaction history space as $\mathcal{O}$, $\mathcal{A}$, and $\hisspace$, respectively. 

Let $\msg \in \mathcal{M}^{N-1}$ denote a set of $N-1$ messages received by the agent. Then, we can build an ablated message subset of $\msg$ with $k$ randomly selected messages, as defined below.

\begin{definition}[$\boldsymbol{k}$-Ablation Message Sample (k-Sample)]
For a message set $\msg \in \mathcal{M}^{N-1}$ and any integer $1 \leq k \leq N-1$, define a $k$-ablation message sample (or \ksample for short), $\msgk \in \mathcal{M}^k$, as a set of $k$ randomly sampled messages from $\msg$. Let $\allmsgk$ be the collection of all unique \ksamples of $\msg$, and thus $|\allmsgk|=\binom{N-1}{k}$.
\end{definition}


Motivated by the fact that the benign messages sent from other agents usually contain overlapping information that may suggest similar decisions, the \textbf{main idea} of our defense is to make decisions based on the \emph{consensus} of the benign messages. 
To be more specific, we train a base policy that makes decisions with one \ksample at each step. During test time when communication messages can be perturbed, the agent collects multiple \ksamples at every step, and applies the trained base policy to each \ksample to get multiple resulting base actions. Then, the agent selects the action that reflects the majority opinion. By carefully designed ablation and ensemble strategies, we can ensure that the majority of these base actions is dominated by benign messages.

With the above intuition, we propose \oursfull (\ours), a generic defense framework that can be fused with any policy learning algorithm. 
\ours has two phases: the training phase where agents are trained in a clean environment, and a testing/defending phase where communications may be perturbed. 
The training and defending strategies of \ours are illustrated below, and the pseudocode is shown by Algorithm~\ref{alg:train} and Algorithm~\ref{alg:test}.

\input{algo_train}
\input{algo_test}

\textbf{Training Phase with \ablnamecap $\boldsymbol\ablpolicy$ (Algorithm~\ref{alg:train})}\quad
During training time, the agent learns a \textit{\ablname} $\ablpolicy: \hisspace \times \mathcal{M}^k \to \mathcal{A}$ which maps its own interaction history $\his$ and a random \ksample $\msgk\sim \mathrm{Uniform}(\allmsgk)$ to an action, where $\mathrm{Uniform}(\allmsgk)$ is the uniform distribution over all \ksamples from the message set $\msg$ it receives.
Here $k$ is a user-specified hyperparameter selected to satisfy certain conditions, as discussed in Section~\ref{sec:algo_theory}. 
The training objective is to maximize the cumulative reward of $\ablpolicy$ based on randomly sampled \ksamples in a non-adversarial environment. Any policy optimization algorithm can be used for training.


\textbf{Defending Phase with \ensnamecap $\boldsymbol\smpolicy$ (Algorithm~\ref{alg:test})}\quad
Once we obtain a reasonable \ablname $\ablpolicy$, we can execute it with ablation and aggregation during test time to defend against adversarial communication. 
The main idea is to collect all possible \ksamples from $\allmsgk$, and select an action suggested by the majority of those \ksamples. 
Specifically, we construct a \textit{\ensname} $\smpolicy: \hisspace \times \mathcal{M}^{N-1} \to \mathcal{A}$ that outputs an action by aggregating the base actions produced by $\ablpolicy$ on multiple \ksamples (Line 5 in Algorithm~\ref{alg:test}). 
The construction of the \ensname depends on whether the agent's action space $\mathcal{A}$ is discrete or continuous, which is given below by Definition~\ref{def:ensemble_dist} and Definition~\ref{def:ensemble_cont}, respectively. 


\begin{definition}[\ensnamecap for Discrete Action Space]
\label{def:ensemble_dist}
For a \ablname $\ablpolicy$ with observation history $\his$ and received message set $\msg$, define the \ensname $\smpolicy$ as
\begin{equation}
\label{eq:ensemble_pi_dist}
    \smpolicy(\his, \mathbf{m}) := \mathrm{argmax}_{a\in\mathcal{A}} \sum\nolimits_{\msgk\in\allmsgk} \mathbbm{1}[\ablpolicy(\his, \msgk)=a].
\end{equation}
\end{definition}


\begin{definition}[\ensnamecap for Continuous Action Space]
\label{def:ensemble_cont}
For a \ablname $\ablpolicy$ with observation history $\his$ and received message set $\msg$, define the \ensname $\smpolicy$ as
\begin{equation}
\label{eq:ensemble_pi_cont}
    \tilde{\pi}(\his, \mathbf{m}) = \funmed \{\ablpolicy(\his, \msgk): \msgk\in\allmsgk \},
\end{equation}
where $\funmed$ is a function that returns the element-wise median value of a set of vectors.
\end{definition}

Therefore, the \ensname $\smpolicy$ takes the action suggested by the consensus of all \ksamples (majority vote in a discrete action space, and median action for a continuous action space). We will show in the next section that, with some mild conditions, 
$\smpolicy$ under adversarial communications works similarly as the \ablname $\ablpolicy$ under all-benign communications.

\textbf{Computation Complexity of \ours}\quad
Different from many model-ensemble methods~\cite{kurutach2018modelensemble,harutyunyan2014shaping} that learn multiple network models, \emph{we learn a single policy $\ablpolicy$ during training, and use a single policy $\smpolicy$ during testing.}
Therefore, the training process does not require extra computations compared to the original policy learning method.
In the defending phase, we aggregate decisions from $\binom{N-1}{k}$ \ksamples, which require $\binom{N-1}{k}$ forward passes through $\ablpolicy$. 
Note that $\binom{N-1}{k}$ is generally small, because the number of agents in practical multi-agent problems is usually small (e.g., the number of drones in an area). 
For example, when $N=10,k=2$, $\binom{N-1}{k}=36$.
In the case where $\binom{N-1}{k}$ is large, we provide a variant of the algorithm in Section~\ref{sec:algo_more}, which takes $D$ \ksamples for $D\leq\binom{N-1}{k}$.
Then the certificate of \ours becomes a high-probability guarantee, depending on the value of $D$.


\subsection{Robustness Certificates of \ours}
\label{sec:algo_theory}

In this section, we provide theoretical guarantees for the robustness of \ours.
During test time, at any step, let $\his$ be the interaction history, $\goodmsg$ be the unperturbed benign message set, and $\advmsg$ be the perturbed message set. 
Note that $\goodmsg$ and $\advmsg$ both have $N-1$ messages while they differ by up to $C$ messages. 
With the above notation, we define a set of actions rendered by purely benign \ksamples in Definition~\ref{def:safeact_dist}. As the agent using a well-trained \ablname is likely to take these actions in a non-adversarial environment, they can be regarded as ``good'' actions to take. 

\begin{definition}[Benign Action Set $\safeaction$]
\label{def:safeact_dist}
For the execution of the \ablname $\ablpolicy$ at any step, define $\safeaction \subseteq \mathcal{A}$ as a set of actions that $\ablpolicy$ may select under benign \ksamples.
\setlength\abovedisplayskip{2pt}
\setlength\belowdisplayskip{2pt}
\begin{equation}
\label{eq:safeact_dist}
    \safeaction := \cup_{\goodmsgk\in\goodallmsgk} \left\{\ablpolicy(\his, \goodmsgk) \right\}.
\end{equation}
\end{definition}

\subsubsection{Certificates for Discrete Action Space}
\label{sec:algo_theory_dist}

We first characterize the action and reward certificate of \ours in a discrete action space, the \ablname takes the action with the most votes from all \ksamples as suggested by Equation~\eqref{eq:ensemble_pi_dist}. To ensure that this action stands for the consensus of benign messages, the following condition is needed.

\begin{condition}[Confident Consensus]
\label{cond:cons}
The highest number of votes among all actions $u_{\max}:=\max_{a\in\mathcal{A}} \sum\nolimits_{\msgk\in\allmsgk} \mathbbm{1}[\ablpolicy(\his, \msgk)=a]$ satisfies
\setlength\abovedisplayskip{2pt}
\setlength\belowdisplayskip{0pt}
\begin{equation}
    \label{eq:u_cond}
    u_{\max} > {\small{\binom{N-1}{k} - \binom{N-1-C}{k}}} =: u_{\mathrm{adv}},
\end{equation}
where $u_{\mathrm{adv}}$ is the number of votes that adversarial messages may affect (number of \ksamples that contain at least one adversarial message). 
\end{condition}

\textbf{Remarks}.
(1) This condition ensures the consensus has more votes than the votes that adversarial messages are involved in. Therefore, when $\smpolicy$ takes an action, there must exist some purely-benign \ksamples voting for this action. 
(2) This condition is easy to satisfy when $C \ll N$ as $\binom{N-1}{k} \approx \binom{N-1-C}{k}$. 
(3) This condition can be easily checked at every step of execution. 

Condition~\ref{cond:cons} considers the worst-case scenario when the adversarial messages collaborate to vote for a malicious action and outweigh benign messages in all \ksamples. 
However, such a worst-case attack is uncommon in practice as attackers are usually not omniscient. Therefore, Condition~\ref{cond:cons} is sufficient but not necessary for the robustness of $\smpolicy$ during execution. 
In real-world problems, our algorithm achieves strong robustness without requiring this condition, as verified in experiments.


\textbf{Action Certificate for Discrete Action Space}\quad
The following theorem suggests that the ensemble policy $\smpolicy$ always takes benign actions under the above conditions no matter whether attacks exist.
\begin{theorem}[Action Certificate for Discrete Action Space]
\label{thm:main_dist}
When Condition~\ref{cond:cons} holds, the ensemble policy $\smpolicy$ in Definition~\ref{def:ensemble_dist} produces benign actions under $\advmsg$, i.e.,
\begin{equation}
    \widetilde{a} = \smpolicy(\his, \advmsg) \in \safeaction.
    \label{eq:cert_dist}
\end{equation}
\end{theorem}
\vspace{-0.5em}

Under Condition~\ref{cond:cons} which are easy to check, Theorem~\ref{thm:main_dist} certifies that $\smpolicy$ ignores the malicious messages in $\mathbf{m}_{\mathrm{adv}}$ and executes a benign action that is suggested by some benign message combinations, even when the malicious messages are not identified. 
Then, we can further derive a reward certificate as introduced below. 

\textbf{Reward Certificate for Discrete Action Space}\quad
When Condition~\ref{cond:cons} holds, the \ensname $\smpolicy$'s action in every step falls into a benign action set, and thus \emph{its cumulative reward under adversarial communications is also guaranteed to be no lower than the worst natural reward the base policy $\ablpolicy$ can get under random benign message subsets.} Therefore, adversarial communication under Assumption~\ref{assum:num} cannot drop the reward of any agent trained with \ours. The formal reward certificate is shown in Corollary~\ref{cor:reward_dist} in Appendix~\ref{app:theory_dist}.

\subsubsection{Certificates in Continuous Action Space}
\label{sec:algo_theory_cont}

The conditions and theoretical results in a continuous action space is slightly different from the discrete case introduced above. We first introduce a condition for certificates to hold.

\begin{condition}[Dominating Benign Sample]
\label{cond:k}
The ablation size $k$ of \ours satisfies
\setlength\abovedisplayskip{2pt}
\setlength\belowdisplayskip{2pt}
\begin{equation}
    \label{eq:k_cond}
    \binom{N-1-C}{k} > \frac{1}{2} \binom{N-1}{k}.
\end{equation}
\end{condition}
\vspace{-0.5em}

\textbf{Remarks.}
(1) For the message set $\advmsg$ that has up to $C$ adversarial messages, Equation~\eqref{eq:k_cond} implies that among all \ksamples from $\advmsg$, there are more purely benign \ksamples than \ksamples that contain adversarial messages. 
(2) Under Assumption~\ref{assum:num}, this condition \textit{always has solutions }for $k$ as $C<\frac{N-1}{2}$, and $k=1$ is always a feasible solution. 

\textbf{Action Certificate for Continuous Action Space}\quad
As in the discrete action case, the continuous ensemble policy will output an action $\widetilde{a}$ that follows the consensus of benign messages. But in a continuous action space, it is hard to ensure that $\widetilde{a}$ is exactly in $\safeaction$. Instead, $\widetilde{a}$ is guaranteed to be within the range of $\safeaction$, as detailed in the following theorem.

\begin{theorem}[Action Certificate for Continuous Action Space]
\label{thm:main_cont}
Under Condition~\ref{cond:k}, the action $\widetilde{a} = \smpolicy(\his, \advmsg)$ generated by the ensemble policy $\smpolicy$ defined in Equation~\eqref{eq:ensemble_pi_cont} satisfies
\begin{equation}
    \widetilde{a} \in \mathsf{Range}(\safeaction) := \{a: \forall 1\leq l \leq L, \exists \underline{a}, \overline{a} \in \safeaction \text{ s.t } \underline{a}_l \leq a_l \leq \overline{a}_l \}.
    \label{eq:cert_cont}
\end{equation}
\end{theorem}

In other words, if the \ensname takes an action $\widetilde{a}$, then at each dimension $l$, there must exist $a_1$ and $a_2$ suggested by some purely benign \ksamples such that the $l$-th dimension of $\widetilde{a}$ is lower and upper bounded by the $l$-th dimension of $a_1$ and $a_2$, respectively. In many practical problems, it is reasonable to assume that actions in $\mathsf{Range}(\safeaction)$ are relatively safe, especially when benign actions in $\safeaction$ are concentrated. For example, if the action denotes the driving speed, and benign message combinations have suggested driving at 40 mph or driving at 50 mph, then driving at 45 mph is also safe. Appendix~\ref{app:theory_cont} provides more examples to explain $\mathsf{Range}(\safeaction)$ in practice.

\textbf{Reward Certificate for Continuous Action Space}\quad
We prove that under Condition~\ref{cond:k}, the reward of the \ensname under communication attacks is lower bounded by the natural reward of \ablname subtracting a environment-dependent constant. Intuitively, the gap between the natural performance and the  worst-case performance under attacks is smaller when the environment dynamics are relatively smooth and the benign \ksamples achieve good consensus. Formal theoretical results and analysis is in Appendix~\ref{app:theory_cont}.

\textbf{How to Select Ablation Size $k$: Trade-off between Performance and Robustness}\quad
The ablation size $k$ is an important hyperparameter for the guarantees to hold. For a fixed $C$, both Condition~\ref{cond:cons} and Condition~\ref{cond:k} prefer a relatively small $k$ as detailed in Appendix~\ref{app:hyper}. That is, a smaller $k$ can improve the robustness in general. Figure~\ref{fig:ablation_cnk} in Appendix~\ref{app:hyper} demonstrates the numerical relationships among $C$, $k$ and $N$, where we can see that a smaller $k$ enables guaranteed defense against more adversarial messages. 
However, we also point out that a smaller $k$ restricts the power of information sharing, as the \ablname can access fewer messages in one step. Therefore, the value of $k$ is related to the \emph{trade-off between robustness and natural performance}~\cite{tsipras2018robustness,zhang2019theoretically}. 
In experiments, we use \textit{the largest $k$ satisfying Equation~\eqref{eq:k_cond}}. 
In practice, one can also train several \ablnames with different $k$'s during training, and later adaptively select a \ablname to construct a \ensname during execution (decrease $k$ if higher robustness is needed).

\vspace{-0.5em}
\subsection{Extensions of \ours}
\label{sec:algo_more}

\textbf{Scaling Up: Ensemble with Partial Samples}\quad
So far we have discussed the proposed \ours defense and the constructed ensemble policy that aggregates all $\binom{N-1}{k}$ number of \ksamples out of $N-1$ messages. 
However, if $N$ is large, sampling all $\binom{N-1}{k}$ combinations of message subsets could be expensive. In this case, a smaller number of \ksamples can be used. That is, given a sample size $0<D\leq\binom{N-1}{k}$, we randomly select $D$ number of \ksamples from $\allmsgk$ (without replacement), and then we aggregate the \ablname's decisions on selected \ksamples. 
In this way, we define a partial-sample version of the ensemble policy, namely \textit{$D$-ensemble policy} $\smpolicy_D$, whose formal definition is provided in Appendix~\ref{app:theory_sample}. In this case, the robustness guarantee holds with high probability that increases as $D$ gets larger, as detailed in Theorem~\ref{thm:sample_cont} in Appendix~\ref{app:theory_sample}. 
Decreasing the hyperparameter $D$ trades some robustness guarantee for higher computational efficiency.
In our experiments, we empirically show that \ours usually works well even for a small $D$ (e.g. $D=1$).

\textbf{More Applications of \ours}\quad
The central idea of \ours is to make decisions based on the consensus instead of each individual.
Therefore, \ours can be combined with any decision-making model to work with possibly-untrustworthy information sources, not only restricted to communication attacks in CMARL. 
For example, when one collects data from multiple sources (e.g., different websites, multiple questionnaires, etc.) and some sources may contain malicious information, \ours can be used to make decisions based on the collected statistics without being significantly misled by false data.
Moreover, the idea of \ours can also be extended to detect attackers, by measuring the difference between the action suggested by any specific communication message and the ensemble policy's action, as detailed and empirically verified in Appendix~\ref{app:exp_detect}.

%% file: algo_train.tex
\begin{figure}[h]
\vspace{-1em}
\centering
\begin{algorithm}[H]
\caption{Training Phase of \ours}
\label{alg:train}
\begin{algorithmic}[1]
   \STATE {\bfseries Input:} ablation size $k$
   \STATE Initialize $\ablpolicy_i$ for every agent $i \in [N]$.
   \REPEAT
       \FOR{$i=1$ {\bfseries to} $N$}
            \STATE Receive a list of messages $\mtoi$, get local observation $o_i$ and update interaction history $\his_i$
            \STATE Randomly sample $\msgki\sim \mathrm{Uniform}(\allmsgki)$
            \STATE Take action based on $\his_i$ and the \ksample $\msgki$, i.e.,
            $a_i \leftarrow \ablpolicy_i (\his_i, \msgki)$ 
            \STATE Update the replay buffer and policy $\ablpolicy_i$
       \ENDFOR
   \UNTIL{end of training}
   \STATE {\bfseries Output:} \ablname $\ablpolicy_i, \forall i \in [N]$
\end{algorithmic}
\end{algorithm}
\vspace{-1em}
\end{figure}

%% file: algo_test.tex

\begin{figure}[h]
\vspace{-2em}
\centering
\begin{algorithm}[H]
   \caption{Defending Phase of \ours}
   \label{alg:test}
\begin{algorithmic}[1]
    \STATE {\bfseries Input:} ablation size $k$; trained \ablname $\ablpolicy_i, \forall i \in [N]$, \\ 
   \REPEAT
       \FOR{$i=1$ {\bfseries to} $N$}
            \STATE Receive a list of messages $\mtoi$ with at most $C$ malicious messages, get local observation $o_i$ and update interaction history $\his_i$
            \STATE Take $\widetilde{a}_i \leftarrow \smpolicy_i (\his_i, \mtoi)$, 
            where $\smpolicy_i$ is the \ensname defined with $\ablpolicy$ by Equation~\eqref{eq:ensemble_pi_dist} for discrete $\mathcal{A}_i$, and Equation~\eqref{eq:ensemble_pi_cont} for continuous $\mathcal{A}_i$
       \ENDFOR
   \UNTIL{end of test}
\end{algorithmic}
\end{algorithm}
\end{figure}

%% file: arxiv/2_relate.tex
\section{Related Work}
\label{sec:relate}

\textbf{Adversarial Robustness of RL Agents}\quad
Section~\ref{sec:setup_adv} introduces several adversarial attacks on single-agent and multi-agent problems.
To improve the robustness of agents, adversarial training (i.e., introducing adversarial agents to the system during training~\cite{pinto2017robust,phan2021resilient,zhang2021robust,sun2021strongest}) and network regularization~\cite{zhang2020robust,shen2020deep,oikarinen2020robust} are empirically shown to be effective under $\ell_p$ attacks, although such robustness is not theoretically guaranteed. 
Certifying the robustness of a network is an important research problem~\cite{raghunathan2018certified,cohen2019certified,gowal2018effectiveness}.
In an effort to certify RL agents' robustness, some approaches~\cite{lutjens2020certified,zhang2020robust,oikarinen2020robust,fischer2019online} apply network certification tools to bound the Q networks and improve the worst-case value of the policy. Kumar et al.~\cite{kumar2021policy} and Wu et al.~\cite{wu2021crop} follow the idea of randomized smoothing~\cite{cohen2019certified} and smooth out the policy by adding Gaussian noise to the input.

\textbf{Communication in MARL}\quad
Communication is crucial in solving collaborative MARL problems. There are many existing studies learning communication protocols across multiple agents. Foerster et al. \cite{foerster2016learning} are the first to learn differentiable communication protocols that is end-to-end trainable across agents. Another work by Sukhbaatar et al. \cite{sukhbaatar2016learning} proposes an efficient permutation-invariant centralized learning algorithm which learns a large feed-forward neural network to map inputs of all agents to their actions. 
It is also important to communicate selectively, since some communication may be less informative or unnecessarily expensive. To tackle this challenge, Das et al. \cite{das2020tarmac} propose an attention mechanism for agents to adaptively select which other agents to send messages to. Liu et al. \cite{liu2020who2com} introduce a handshaking procedure so that the agents communicate only when needed. Our work assumes a pre-trained communication protocol and does not consider the problem of learning communication mechanisms.

\textbf{Adversarial Attacks and Defenses in CMARL}\quad
Recently, the problem of adversarial communication in MARL has attracted increasing attention.  
Blumenkamp et al.~\cite{blumenkamp2020emergence} show that in a cooperative game, communication policies of some self-interest agents can hurt other agents' performance. 
To achieve robust communication, Mitchell et al.~\cite{mitchell2020gaussian} adopt a Gaussian Process-based probabilistic model to compute the posterior probabilities that whether each partner is truthful. 
Tu et al.~\cite{tu2021adversarial} investigate the vulnerability of multi-agent autonomous systems against communication attacks, with a focus on vision tasks.
Xue et al.~\cite{xue2021mis} propose an algorithm to defend against one adversarial communication message by an anomaly detector and a message reconstructor, which are trained with groundtruth labels and messages.

To the best of our knowledge, our \ours is the first certifiable defense in MARL against communication attacks. Moreover, we consider a strong threat model where up to half of the communication messages can be arbitrarily corrupted, capturing many realistic types of attacks.

%% file: neurips/5_exp.tex
\vspace{-0.5em}
\section{Empirical Study}
\label{sec:exp}

\input{\texhome/figs_all}

In this section, we verify the robustness of our \ours in multiple different CMARL environments against various communication attack algorithms. 
Then, we conduct hyperparameter tests for the ablation size $k$ and the sample size $D$ (for the variant of \ours introduced in Section~\ref{sec:algo_more}).

\textbf{Environments}\quad 
To evaluate the effectiveness of \ours, we consider the following environments.\\
$\bullet$ \textit{FoodCollector}: a 2D particle environment adapted from the WaterWorld task in PettingZoo~\cite{terry2020pettingzoo}. In our experiment, there are $N=9$ agents with different colors searching for foods with the same colors. Agents use their limited-range sensors to observe the surrounding objects, while communicating with each other to share their observations. The action space can be either discrete (9 moving directions and 1 no-move action), or continuous (2-dimensional vector denoting acceleration). Agents are penalized for having leftover foods at every step, so they seek to find all foods as fast as possible. \\
$\bullet$ \textit{\marketenv}: simulating a real-world inventory management problem. 
$N=10$ cooperative distributor agents carry inventory for 3 products. There are 300 buyers sharing an underlying demand distribution.
At every step, each buyer requests products from a randomly selected distributor. 
Then, each distributor agent observes random demand requests, takes restocking actions (continuous) to adjust its own inventory, and communicates with others by sending its demand observations.
Distributors are penalized for mismatches between their inventory and the demands. \\
The detailed description of each environment is in Appendix~\ref{app:exp_env}, where we empirically show that agents benefit a lot from communication in these environments.
Since our focus is to defend against adversarial communication over any communication policy, we let the benign agents communicate via fixed protocols detailed in Appendix~\ref{app:exp_env}. But our method also works for learned communications, as verified by additional experiments on an \textit{MARL image classification task} in Appendix~\ref{app:exp_mnist}.

\textbf{Implementation and Baselines}\quad
As discussed in Section~\ref{sec:algo}, our \ours is designed to defend against communication attacks, and can be fused with any policy learning algorithm. 
In FoodCollector and \marketenv, we use PPO~\cite{schulman2017proximal} with parameter sharing~\cite{terry2020parameter} as the base policy learning algorithm as it achieves relatively high natural performance. But our \ours can be applied to other MARL algorithms, such as QMIX~\cite{rashid2018qmix} that is evaluated in Appendix~\ref{app:exp_results}.
With the same policy learning method, we compare our \ours with two defense baselines: 
(1) \textbf{(Vanilla)}: vanilla training without defense, which learns a policy based on all benign messages. 
(2) \textbf{(AT)} adversarial training as in~\cite{zhang2021robust}, which alternately trains an adaptive RL attacker and an agent. 
During training and defending, we set the ablation size \ours as $k=2$, the largest solution to Equation~\eqref{eq:k_cond} for $C=2$ when $N=9$ (FoodCollector) or $N=10$ (\marketenv). For AT, we train the agent against $C=2$ learned adversaries. 
More implementation details are provided in Appendix~\ref{app:exp_imple}.

\textbf{Evaluation Metrics}\quad
To evaluate the effectiveness of our defense strategy, we test the performance of the trained policies under no attack and under various values of $C$ (for simplicity, we refer to $C$ as the \textit{number of adversaries}). 
Different from previous work~\cite{blumenkamp2020emergence} where the adversarial agent disrupts all the other agents, we consider the case where the attacker deliberately misleads a selected victim, which could evaluate the robustness of the victim under the harshest attacks. We report the victim's local reward under the following two types of attack methods: 
(1) \textit{Non-adaptive attack} that perturbs messages based on heuristics. In FoodCollector, we consider randomly generated messages within the valid range of communication messages. Note that this is already a strong attack since a randomly generated message could be arbitrary and far from the original benign message. For \marketenv where messages have clear physical meanings (demand of buyers), we consider 3 realistic attacks: \permute, \swap, \flip, which permute, swap or flip the demand observations, respectively, as detailed in Appendix~\ref{app:exp_imple:market}. 
(2) \textit{Learned adaptive attack} that learns the strongest/worst adversarial communication with an RL algorithm to minimize the victim's reward (a white-box attacker which knows the victim's reward). The learned attack can strategically mislead the victim.
As shown in prior works~\cite{zhang2020robust,zhang2021robust,sun2021strongest}, the theoretically optimal attack (which minimizes the victim's reward) can be formed as an RL problem and learned by RL algorithms. 
Therefore, we can regard this attack as a worst-case attack for the victim agents.
More details are in Appendix~\ref{app:exp_imple}.

\textbf{Experiment Results}\quad
The results are shown in Figure~\ref{fig:res_all}. 
We can see that the rewards of Vanilla and AT drastically drop under attacks. Under strong adaptive attackers, Vanilla and AT sometimes perform worse than a non-communicative agent shown by dashed red lines, which suggests that communication can be a double-edged sword. Although AT is usually effective for $\ell_p$ attacks~\cite{zhang2021robust}, we find that AT does not achieve better robustness than Vanilla, since it can not adapt to arbitrary perturbations to several messages.
(More analysis of AT is in Appendix~\ref{app:exp_imple}.)
In contrast, \textit{\ours can utilize benign communication well while being robust to adversarial communication. }
We use $k=2$ for our \ours, which in theory provides performance guarantees against up to $C=2$ adversaries for $N=9$ and $N=10$. 
We can see that the reward of \ours under $C=1$ or $C=2$ is similar to its reward under no attack, matching our theoretical analysis. 
Even under 3 adversaries where the theoretical guarantees no longer hold, \ours still obtains superior performance compared to Vanilla and AT.
Therefore, \textit{\ours makes agents robust under varying numbers of adversaries.} 

\begin{wrapfigure}{r}{0.3\textwidth}
\vspace{-1.5em}
\hspace{-1em}
 \centering
 \resizebox{0.26\columnwidth}{!}{\input{figs/mnist/MNIST_Adaptive_Attacker_Precision}}
 \vspace{-0.5em}
\caption{\textbf{(MARL-MNIST)}: Precision of MARL classification on MNIST without or with \ours, under learned adaptive attacks.}
\vspace{-1.5em}
\label{fig:res_mnist_sample}
\end{wrapfigure}

\textbf{Additional Results in MARL Image Classification}\quad
In addition to the above two environments, we also evaluate our proposed \ours defense in an MARL classification task in the MNIST dataset~\cite{lecun1998gradient}, which is proposed by Mousavi et al.~\cite{mousavi2019multi}. In this classification problem, every agent observes a local patch of an image and takes actions to move to adjacent patches. Agents are allowed to communicate their encoded local beliefs which are \textit{learned} by LSTM networks~\cite{hochreiter1997long}. Figure~\ref{fig:res_mnist_sample} shows the classification precision of \ours and baselines under adaptive attacks. We can see that the attacks significantly mislead Vanilla and AT, while \ours stays robust.
More details and results of this task are provided in Appendix~\ref{app:exp_mnist}.

\textbf{Hyperparameter Tests for Ablation Size $\boldsymbol k$}\quad
To see the empirical influence of $k$ on \ours, we evaluate \ours's natural reward and reward under $C=2$ attacks with different ablation sizes $k$. 
The result in discrete-action FoodCollector is shown in Figure~\ref{fig:res_k_dist}(left), while results in other environments are similar and are put in Appendix~\ref{app:exp_results}.
We observe that a larger $k$ leads to higher natural performance since each agent could gather more information from others. However, raising $k$ also increases the risk of making decisions based on communication messages sent from an adversary. 
Therefore, \textit{increasing ablation size $k$ trades off robustness for natural performance}, matching the analysis in Section~\ref{sec:algo_theory}.
Moreover, as the largest solution to Equation~\eqref{eq:k_cond} in Condition~\ref{cond:k}, ablation size $k=2$ achieves a good balance between performance and robustness. Even when $k>2$ which breaks Condition~\ref{cond:k}, \ours is still more robust than baselines.

\begin{wrapfigure}{r}{0.5\textwidth}
    \vspace{-2em}
     \centering
     \begin{subfigure}[t]{0.24\columnwidth}
      \resizebox{\columnwidth}{!}{\input{figs/FoodCollector_Ablation_k_2adv_dist}}
     \end{subfigure}
     \hfill
     \begin{subfigure}[t]{0.24\columnwidth}
      \resizebox{\columnwidth}{!}{\input{figs/FoodCollector_Ablation_k_sample_dist}}
     \end{subfigure}
     \vspace{-0.5em}
    \caption{Natural and robust performance of \ours with various values of \textbf{(left)} ablation size $k$ and \textbf{(right)} sample size $D$, in discrete-action FoodCollector under $C=2$. 
    }
    \vspace{-1em}
    \label{fig:res_k_dist}
    \end{wrapfigure}
    
\textbf{Hyper-parameter Test for Sample Size $\boldsymbol D$}\quad
We further evaluate the partial-sample variant of \ours introduced in Section~\ref{sec:algo_more} using $k=2$ under $C=2$ adversaries, with $D$ varying from $N-1 \choose k$ (the largest sample size) to 1 (the smallest sample size). Figure~\ref{fig:res_k_dist}(right) demonstrate the performance of different $D$'s in discrete-action FoodCollector, and Appendix~\ref{app:exp_results} shows other results. 
As $D$ goes down, \ours obtains lower reward under attackers, but it is still significantly more robust than baseline methods. For example, \ours in FoodCollector obtains much higher reward than Vanilla under attacks even when $D=1$ (using only 1 random \ksample in the \ensname). 

%% file: neurips/figs_all.tex
\begin{figure}[!b]
\vspace{-2em}
 \centering
\rotatebox{90}{\scriptsize{\qquad \quad \textbf{FoodCollector}}}
 \begin{subfigure}[t]{0.24\columnwidth}
  \resizebox{\textwidth}{!}{\input{figs/FoodCollector_Attacker_Dist_naive}}
  \vspace{-1.5em}
  \caption{Disc. \& Non-adaptive}
 \end{subfigure}
 \hfill
 \begin{subfigure}[t]{0.24\columnwidth}
  \resizebox{\textwidth}{!}{\input{figs/FoodCollector_Attacker_Dist}}
  \vspace{-1.5em}
  \caption{Disc. \& Adaptive}
 \end{subfigure}
 \hfill
 \begin{subfigure}[t]{0.24\columnwidth}
  \resizebox{\textwidth}{!}{\input{figs/FoodCollector_Attacker_Cont_naive}}
  \vspace{-1.5em}
  \caption{Cont. \& Non-adaptive}
 \end{subfigure}
 \hfill
 \begin{subfigure}[t]{0.24\columnwidth}
  \resizebox{\textwidth}{!}{ \input{figs/FoodCollector_Attacker_Cont}}
  \vspace{-1.5em}
  \caption{Cont. \& Adaptive}
 \end{subfigure}

\rotatebox{90}{\scriptsize{\enspace\textbf{\marketenv}}}
 \begin{subfigure}[t]{0.24\textwidth}
 \centering
  \resizebox{\linewidth}{!}{\input{figs/MarketEnv/MarketEnv_Attacker_Dist_adaptive}}
  \vspace{-1.5em}
  \caption{Adaptive Attack}
 \end{subfigure}
 \begin{subfigure}[t]{0.24\textwidth}
 \centering
   \resizebox{\linewidth}{!}{\input{figs/MarketEnv/MarketEnv_Attacker_Dist_permute}}
  \vspace{-1.5em}
  \caption{\permute}
 \end{subfigure}
 \hfill
 \begin{subfigure}[t]{0.24\textwidth}
 \centering
   \resizebox{\linewidth}{!}{\input{figs/MarketEnv/MarketEnv_Attacker_Dist_flip}}
  \vspace{-1.5em}
  \caption{\swap}
 \end{subfigure}
 \begin{subfigure}[t]{0.24\textwidth}
 \centering
  \resizebox{\linewidth}{!}{ \input{figs/MarketEnv/MarketEnv_Attacker_Dist_swap}}
  \vspace{-1.5em}
  \caption{\flip}
 \end{subfigure}
 \vspace{-0.5em}
\caption{Rewards of our \ours and baselines in FoodCollector and \marketenv, under no attacker and varying numbers of adversaries for adaptive and various non-adaptive attacks. The dashed red lines stand for the average performance of a non-communicative agent. Results are averaged over 5 random seeds.
}
\vspace{-2em}
\label{fig:res_all}
\end{figure}

%% file: figs/FoodCollector_Attacker_Dist_naive.tex
\begin{tikzpicture}

\definecolor{color0}{rgb}{0.56078431372549,0.737254901960784,0.56078431372549}
\definecolor{color1}{rgb}{0,0.749019607843137,1}
\definecolor{color2}{rgb}{1,0.388235294117647,0.27843137254902}

\begin{axis}[
axis line style={white!15!black},
legend cell align={left},
legend style={fill opacity=1, draw opacity=1, text opacity=1, draw=none, fill=none, at={(1,1)}, anchor=north east},
tick align=outside,
x grid style={white!80!black},
xmin=-0.48, xmax=3.48,
xtick style={color=white!15!black},
xtick={0,1,2,3},
xtick pos = lower,
xticklabels={No Attacker,$C=1$,$C=2$,$C=3$},
y grid style={white!80!black},
ymin=0, ymax=100,
ytick style={color=white!15!black},
ytick={0,20,40,60,80,100},
ytick pos = left,
ylabel = {\textbf{Reward}},
yticklabels={-100,-80,-60,-40,-20,0}
]
\draw[draw=black,fill=color0] (axis cs:-0.3,0) rectangle (axis cs:-0.1,86.55);
\draw[draw=black,fill=color0] (axis cs:0.7,0) rectangle (axis cs:0.9,60.98);
\draw[draw=black,fill=color0] (axis cs:1.7,0) rectangle (axis cs:1.9,51.93);
\draw[draw=black,fill=color0] (axis cs:2.7,0) rectangle (axis cs:2.9,44.03);
\draw[draw=black,fill=color1] (axis cs:-0.1,0) rectangle (axis cs:0.1,77.53);
\draw[draw=black,fill=color1] (axis cs:0.9,0) rectangle (axis cs:1.1,27.92);
\draw[draw=black,fill=color1] (axis cs:1.9,0) rectangle (axis cs:2.1,17.15);
\draw[draw=black,fill=color1] (axis cs:2.9,0) rectangle (axis cs:3.1,16.89);
\draw[draw=black,fill=color2] (axis cs:0.1,0) rectangle (axis cs:0.3,74.19);
\draw[draw=black,fill=color2] (axis cs:1.1,0) rectangle (axis cs:1.3,75.66);
\draw[draw=black,fill=color2] (axis cs:2.1,0) rectangle (axis cs:2.3,74.39);
\draw[draw=black,fill=color2] (axis cs:3.1,0) rectangle (axis cs:3.3,76.57);
\path [draw=black, semithick]
(axis cs:-0.2,85.15)
--(axis cs:-0.2,87.95);

\path [draw=black, semithick]
(axis cs:0.8,59.27)
--(axis cs:0.8,62.69);

\path [draw=black, semithick]
(axis cs:1.8,47.79)
--(axis cs:1.8,56.07);

\path [draw=black, semithick]
(axis cs:2.8,40.54)
--(axis cs:2.8,47.52);

\path [draw=black, semithick]
(axis cs:0,76.72)
--(axis cs:0,78.34);

\path [draw=black, semithick]
(axis cs:1,22.91)
--(axis cs:1,32.93);

\path [draw=black, semithick]
(axis cs:2,10.65)
--(axis cs:2,23.65);

\path [draw=black, semithick]
(axis cs:3,10.7)
--(axis cs:3,23.08);

\path [draw=black, semithick]
(axis cs:0.2,73.38)
--(axis cs:0.2,75);

\path [draw=black, semithick]
(axis cs:1.2,70.65)
--(axis cs:1.2,80.67);

\path [draw=black, semithick]
(axis cs:2.2,67.89)
--(axis cs:2.2,80.89);

\path [draw=black, semithick]
(axis cs:3.2,70.38)
--(axis cs:3.2,82.76);

\addplot [semithick, black, mark=-, mark size=6, mark options={solid}, only marks, forget plot]
table {%
-0.2 85.15
0.8 59.27
1.8 47.79
2.8 40.54
};
\addplot [semithick, black, mark=-, mark size=6, mark options={solid}, only marks, forget plot]
table {%
-0.2 87.95
0.8 62.69
1.8 56.07
2.8 47.52
};
\addplot [semithick, black, mark=-, mark size=6, mark options={solid}, only marks, forget plot]
table {%
0 76.72
1 22.91
2 10.65
3 10.7
};
\addplot [semithick, black, mark=-, mark size=6, mark options={solid}, only marks, forget plot]
table {%
0 78.34
1 32.93
2 23.65
3 23.08
};
\addplot [semithick, black, mark=-, mark size=6, mark options={solid}, only marks, forget plot]
table {%
0.2 73.38
1.2 70.65
2.2 67.89
3.2 70.38
};
\addplot [semithick, black, mark=-, mark size=6, mark options={solid}, only marks, forget plot]
table {%
0.2 75
1.2 80.67
2.2 80.89
3.2 82.76
};
\addplot [semithick, green!50!black, mark=*, mark size=3, mark options={solid}]
table {%
-0.2 86.55
0.8 60.98
1.8 51.93
2.8 44.03
};
\addlegendentry{Vanilla}
\addplot [semithick, blue, mark=x, mark size=3, mark options={solid}]
table {%
0 77.53
1 27.92
2 17.15
3 16.89
};
\addlegendentry{AT}
\addplot [semithick, red, mark=square*, mark size=3, mark options={solid}]
table {%
0.2 74.19
1.2 75.66
2.2 74.39
3.2 76.57
};
\addlegendentry{\ours (Ours)}
\end{axis}

\end{tikzpicture}

%% file: figs/FoodCollector_Attacker_Dist.tex
\begin{tikzpicture}

\definecolor{color0}{rgb}{0.56078431372549,0.737254901960784,0.56078431372549}
\definecolor{color1}{rgb}{0,0.749019607843137,1}
\definecolor{color2}{rgb}{1,0.388235294117647,0.27843137254902}

\begin{axis}[
axis line style={white!15!black},
legend cell align={left},
legend style={fill opacity=1, draw opacity=1, text opacity=1, draw=none, fill=none},
tick align=outside,
x grid style={white!80!black},
xmajorticks=true,
xmin=-0.48, xmax=3.48,
xtick style={color=white!15!black},
xtick={0,1,2,3},
xtick pos = lower,
xticklabels={No Attacker,$C=1$, $C=2$, $C=3$},
y grid style={white!80!black},
ymajorticks=true,
ymin=0, ymax=100,
ytick style={color=white!15!black},
ytick={0,20,40,60,80,100},
ytick pos = left,
ylabel = {\textbf{Reward}},
yticklabels={-100,-80,-60,-40,-20,0}
]
\draw[draw=black,fill=color0] (axis cs:-0.3,0) rectangle (axis cs:-0.1,86.55);
\draw[draw=black,fill=color0] (axis cs:0.7,0) rectangle (axis cs:0.9,42.87);
\draw[draw=black,fill=color0] (axis cs:1.7,0) rectangle (axis cs:1.9,20.9);
\draw[draw=black,fill=color0] (axis cs:2.7,0) rectangle (axis cs:2.9,19.31);
\draw[draw=black,fill=color1] (axis cs:-0.1,0) rectangle (axis cs:0.1,77.53);
\draw[draw=black,fill=color1] (axis cs:0.9,0) rectangle (axis cs:1.1,19.74);
\draw[draw=black,fill=color1] (axis cs:1.9,0) rectangle (axis cs:2.1,12.05);
\draw[draw=black,fill=color1] (axis cs:2.9,0) rectangle (axis cs:3.1,7.98);
\draw[draw=black,fill=color2] (axis cs:0.1,0) rectangle (axis cs:0.3,74.19);
\draw[draw=black,fill=color2] (axis cs:1.1,0) rectangle (axis cs:1.3,71.66);
\draw[draw=black,fill=color2] (axis cs:2.1,0) rectangle (axis cs:2.3,71.39);
\draw[draw=black,fill=color2] (axis cs:3.1,0) rectangle (axis cs:3.3,46.57);
\path [draw=black, semithick]
(axis cs:-0.2,85.15)
--(axis cs:-0.2,87.95);

\path [draw=black, semithick]
(axis cs:0.8,41.16)
--(axis cs:0.8,44.58);

\path [draw=black, semithick]
(axis cs:1.8,16.76)
--(axis cs:1.8,25.04);

\path [draw=black, semithick]
(axis cs:2.8,15.82)
--(axis cs:2.8,22.8);

\path [draw=black, semithick]
(axis cs:0,76.72)
--(axis cs:0,78.34);

\path [draw=black, semithick]
(axis cs:1,16.73)
--(axis cs:1,22.75);

\path [draw=black, semithick]
(axis cs:2,11.24)
--(axis cs:2,12.86);

\path [draw=black, semithick]
(axis cs:3,7.09)
--(axis cs:3,8.87);

\path [draw=black, semithick]
(axis cs:0.2,73.38)
--(axis cs:0.2,75);

\path [draw=black, semithick]
(axis cs:1.2,68.65)
--(axis cs:1.2,74.67);

\path [draw=black, semithick]
(axis cs:2.2,70.58)
--(axis cs:2.2,72.2);

\path [draw=black, semithick]
(axis cs:3.2,45.68)
--(axis cs:3.2,47.46);

\addplot [semithick, black, mark=-, mark size=6, mark options={solid}, only marks, forget plot]
table {%
-0.2 85.15
0.8 41.16
1.8 16.76
2.8 15.82
};
\addplot [semithick, black, mark=-, mark size=6, mark options={solid}, only marks, forget plot]
table {%
-0.2 87.95
0.8 44.58
1.8 25.04
2.8 22.8
};
\addplot [semithick, black, mark=-, mark size=6, mark options={solid}, only marks, forget plot]
table {%
0 76.72
1 16.73
2 11.24
3 7.09
};
\addplot [semithick, black, mark=-, mark size=6, mark options={solid}, only marks, forget plot]
table {%
0 78.34
1 22.75
2 12.86
3 8.87
};
\addplot [semithick, black, mark=-, mark size=6, mark options={solid}, only marks, forget plot]
table {%
0.2 73.38
1.2 68.65
2.2 70.58
3.2 45.68
};
\addplot [semithick, black, mark=-, mark size=6, mark options={solid}, only marks, forget plot]
table {%
0.2 75
1.2 74.67
2.2 72.2
3.2 47.46
};
\addplot [semithick, green!50!black, mark=*, mark size=3, mark options={solid}]
table {%
-0.2 86.55
0.8 42.87
1.8 20.9
2.8 19.31
};
\addlegendentry{Vanilla}
\addplot [semithick, blue, mark=x, mark size=3, mark options={solid}]
table {%
0 77.53
1 19.74
2 12.05
3 7.98
};
\addlegendentry{AT}
\addplot [semithick, red, mark=square*, mark size=3, mark options={solid}]
table {%
0.2 74.19
1.2 71.66
2.2 71.39
3.2 46.57
};
\addlegendentry{\ours Ours}
\addplot [semithick, red, dashed, forget plot]
table {%
-0.48 28.12
3.48 28.12
};
\end{axis}

\end{tikzpicture}

%% file: cmarl/figs/FoodCollector_Attacker_Cont_Naive.tex
\begin{tikzpicture}

\definecolor{color0}{rgb}{0.56078431372549,0.737254901960784,0.56078431372549}
\definecolor{color1}{rgb}{0,0.749019607843137,1}
\definecolor{color2}{rgb}{1,0.388235294117647,0.27843137254902}

\begin{axis}[
axis line style={white!15!black},
legend cell align={left},
legend style={fill opacity=0, draw opacity=1, text opacity=1, draw=none, fill=none, at={(1,1)}, anchor=north east},
tick align=outside,
x grid style={white!80!black},
xmajorticks=true,
xmin=-0.48, xmax=3.48,
xtick style={color=white!15!black},
xtick={0,1,2,3},
xtick pos = lower,
xticklabels={No Attacker,$C=1$, $C=2$, $C=3$},
y grid style={white!80!black},
ymajorticks=true,
ymin=0, ymax=100,
ytick style={color=white!15!black},
ytick={0,20,40,60,80,100},
yticklabels={-100,-80,-60,-40,-20,0},
ytick pos = left,
ylabel = {\textbf{Reward}},
]
\draw[draw=black,fill=color0] (axis cs:-0.3,0) rectangle (axis cs:-0.1,72.21);
\draw[draw=black,fill=color0] (axis cs:0.7,0) rectangle (axis cs:0.9,61.52);
\draw[draw=black,fill=color0] (axis cs:1.7,0) rectangle (axis cs:1.9,60.25);
\draw[draw=black,fill=color0] (axis cs:2.7,0) rectangle (axis cs:2.9,54.54);
\draw[draw=black,fill=color1] (axis cs:-0.1,0) rectangle (axis cs:0.1,68.74);
\draw[draw=black,fill=color1] (axis cs:0.9,0) rectangle (axis cs:1.1,57.02);
\draw[draw=black,fill=color1] (axis cs:1.9,0) rectangle (axis cs:2.1,52.6);
\draw[draw=black,fill=color1] (axis cs:2.9,0) rectangle (axis cs:3.1,51.62);
\draw[draw=black,fill=color2] (axis cs:0.1,0) rectangle (axis cs:0.3,64.79);
\draw[draw=black,fill=color2] (axis cs:1.1,0) rectangle (axis cs:1.3,63.21);
\draw[draw=black,fill=color2] (axis cs:2.1,0) rectangle (axis cs:2.3,63.31);
\draw[draw=black,fill=color2] (axis cs:3.1,0) rectangle (axis cs:3.3,61.89);
\path [draw=black, semithick]
(axis cs:-0.2,67.32)
--(axis cs:-0.2,77.1);

\path [draw=black, semithick]
(axis cs:0.8,57.69)
--(axis cs:0.8,65.35);

\path [draw=black, semithick]
(axis cs:1.8,55.95)
--(axis cs:1.8,64.55);

\path [draw=black, semithick]
(axis cs:2.8,48.69)
--(axis cs:2.8,60.39);

\path [draw=black, semithick]
(axis cs:0,67.93)
--(axis cs:0,69.55);

\path [draw=black, semithick]
(axis cs:1,51.28)
--(axis cs:1,62.76);

\path [draw=black, semithick]
(axis cs:2,49.47)
--(axis cs:2,55.73);

\path [draw=black, semithick]
(axis cs:3,48.89)
--(axis cs:3,54.35);

\path [draw=black, semithick]
(axis cs:0.2,63.98)
--(axis cs:0.2,65.6);

\path [draw=black, semithick]
(axis cs:1.2,57.47)
--(axis cs:1.2,68.95);

\path [draw=black, semithick]
(axis cs:2.2,60.18)
--(axis cs:2.2,66.44);

\path [draw=black, semithick]
(axis cs:3.2,59.16)
--(axis cs:3.2,64.62);

\addplot [semithick, black, mark=-, mark size=6, mark options={solid}, only marks, forget plot]
table {%
-0.2 67.32
0.8 57.69
1.8 55.95
2.8 48.69
};
\addplot [semithick, black, mark=-, mark size=6, mark options={solid}, only marks, forget plot]
table {%
-0.2 77.1
0.8 65.35
1.8 64.55
2.8 60.39
};
\addplot [semithick, black, mark=-, mark size=6, mark options={solid}, only marks, forget plot]
table {%
0 67.93
1 51.28
2 49.47
3 48.89
};
\addplot [semithick, black, mark=-, mark size=6, mark options={solid}, only marks, forget plot]
table {%
0 69.55
1 62.76
2 55.73
3 54.35
};
\addplot [semithick, black, mark=-, mark size=6, mark options={solid}, only marks, forget plot]
table {%
0.2 63.98
1.2 57.47
2.2 60.18
3.2 59.16
};
\addplot [semithick, black, mark=-, mark size=6, mark options={solid}, only marks, forget plot]
table {%
0.2 65.6
1.2 68.95
2.2 66.44
3.2 64.62
};
\addplot [semithick, green!50!black, mark=*, mark size=3, mark options={solid}]
table {%
-0.2 72.21
0.8 61.52
1.8 60.25
2.8 54.54
};
\addlegendentry{Vanilla}
\addplot [semithick, blue, mark=x, mark size=3, mark options={solid}]
table {%
0 68.74
1 57.02
2 52.6
3 51.62
};
\addlegendentry{AT}
\addplot [semithick, red, mark=square*, mark size=3, mark options={solid}]
table {%
0.2 64.79
1.2 63.21
2.2 63.31
3.2 61.89
};
\addlegendentry{\ours (Ours)}
\addplot [semithick, red, dashed, forget plot]
table {%
-0.48 40.14
3.48 40.14
};
\end{axis}

\end{tikzpicture}

%% file: figs/FoodCollector_Attacker_Cont.tex
\begin{tikzpicture}

\definecolor{color0}{rgb}{0.56078431372549,0.737254901960784,0.56078431372549}
\definecolor{color1}{rgb}{0,0.749019607843137,1}
\definecolor{color2}{rgb}{1,0.388235294117647,0.27843137254902}

\begin{axis}[
axis line style={white!15!black},
legend cell align={left},
legend style={fill opacity=1, draw opacity=1, text opacity=1, draw=none, fill=none},
tick align=outside,
x grid style={white!80!black},
xmajorticks=true,
xmin=-0.48, xmax=3.48,
xtick style={color=white!15!black},
xtick={0,1,2,3},
xtick pos = lower,
xticklabels={No Attacker, $C=1$, $C=2$, $C=3$},
y grid style={white!80!black},
ymajorticks=true,
ymin=0, ymax=100,
ytick style={color=white!15!black},
ytick pos = left,
ytick={0,20,40,60,80,100},
yticklabels={-100,-80,-60,-40,-20,0},
ylabel = {\textbf{Reward}},
]
\draw[draw=black,fill=color0] (axis cs:-0.3,0) rectangle (axis cs:-0.1,72.21);
\draw[draw=black,fill=color0] (axis cs:0.7,0) rectangle (axis cs:0.9,50.22);
\draw[draw=black,fill=color0] (axis cs:1.7,0) rectangle (axis cs:1.9,25.45);
\draw[draw=black,fill=color0] (axis cs:2.7,0) rectangle (axis cs:2.9,24.94);
\draw[draw=black,fill=color1] (axis cs:-0.1,0) rectangle (axis cs:0.1,68.74);
\draw[draw=black,fill=color1] (axis cs:0.9,0) rectangle (axis cs:1.1,42.74);
\draw[draw=black,fill=color1] (axis cs:1.9,0) rectangle (axis cs:2.1,41.4);
\draw[draw=black,fill=color1] (axis cs:2.9,0) rectangle (axis cs:3.1,24.09);
\draw[draw=black,fill=color2] (axis cs:0.1,0) rectangle (axis cs:0.3,64.79);
\draw[draw=black,fill=color2] (axis cs:1.1,0) rectangle (axis cs:1.3,60.52);
\draw[draw=black,fill=color2] (axis cs:2.1,0) rectangle (axis cs:2.3,54.25);
\draw[draw=black,fill=color2] (axis cs:3.1,0) rectangle (axis cs:3.3,49.08);
\path [draw=black, semithick]
(axis cs:-0.2,67.32)
--(axis cs:-0.2,77.1);

\path [draw=black, semithick]
(axis cs:0.8,46.77)
--(axis cs:0.8,53.67);

\path [draw=black, semithick]
(axis cs:1.8,23.46)
--(axis cs:1.8,27.44);

\path [draw=black, semithick]
(axis cs:2.8,19.41)
--(axis cs:2.8,30.47);

\path [draw=black, semithick]
(axis cs:0,67.93)
--(axis cs:0,69.55);

\path [draw=black, semithick]
(axis cs:1,39.73)
--(axis cs:1,45.75);

\path [draw=black, semithick]
(axis cs:2,40.59)
--(axis cs:2,42.21);

\path [draw=black, semithick]
(axis cs:3,23.2)
--(axis cs:3,24.98);

\path [draw=black, semithick]
(axis cs:0.2,63.98)
--(axis cs:0.2,65.6);

\path [draw=black, semithick]
(axis cs:1.2,57.51)
--(axis cs:1.2,63.53);

\path [draw=black, semithick]
(axis cs:2.2,53.44)
--(axis cs:2.2,55.06);

\path [draw=black, semithick]
(axis cs:3.2,48.19)
--(axis cs:3.2,49.97);

\addplot [semithick, black, mark=-, mark size=6, mark options={solid}, only marks, forget plot]
table {%
-0.2 67.32
0.8 46.77
1.8 23.46
2.8 19.41
};
\addplot [semithick, black, mark=-, mark size=6, mark options={solid}, only marks, forget plot]
table {%
-0.2 77.1
0.8 53.67
1.8 27.44
2.8 30.47
};
\addplot [semithick, black, mark=-, mark size=6, mark options={solid}, only marks, forget plot]
table {%
0 67.93
1 39.73
2 40.59
3 23.2
};
\addplot [semithick, black, mark=-, mark size=6, mark options={solid}, only marks, forget plot]
table {%
0 69.55
1 45.75
2 42.21
3 24.98
};
\addplot [semithick, black, mark=-, mark size=6, mark options={solid}, only marks, forget plot]
table {%
0.2 63.98
1.2 57.51
2.2 53.44
3.2 48.19
};
\addplot [semithick, black, mark=-, mark size=6, mark options={solid}, only marks, forget plot]
table {%
0.2 65.6
1.2 63.53
2.2 55.06
3.2 49.97
};
\addplot [semithick, green!50!black, mark=*, mark size=3, mark options={solid}]
table {%
-0.2 72.21
0.8 50.22
1.8 25.45
2.8 24.94
};
\addlegendentry{Vanilla}
\addplot [semithick, blue, mark=x, mark size=3, mark options={solid}]
table {%
0 68.74
1 42.74
2 41.4
3 24.09
};
\addlegendentry{AT}
\addplot [semithick, red, mark=square*, mark size=3, mark options={solid}]
table {%
0.2 64.79
1.2 60.52
2.2 54.25
3.2 49.08
};
\addlegendentry{\ours (Ours)}
\addplot [semithick, red, dashed, forget plot]
table {%
-0.48 40.14
3.48 40.14
};
\end{axis}

\end{tikzpicture}

%% file: figs/MarketEnv/MarketEnv_Attacker_Dist_adaptive.tex
\begin{tikzpicture}

\definecolor{darkgray176}{RGB}{176,176,176}
\definecolor{darkseagreen}{RGB}{143,188,143}
\definecolor{deepskyblue}{RGB}{0,191,255}
\definecolor{green01270}{RGB}{0,127,0}
\definecolor{tomato}{RGB}{255,99,71}

\begin{axis}[
legend cell align={left},
legend style={fill opacity=1, draw opacity=1, text opacity=1, draw=none, fill=none},
tick align=outside,
tick pos=left,
x grid style={darkgray176},
xmin=-0.48, xmax=3.48,
xtick style={color=black},
xtick={0,1,2,3},
xticklabels={No Attacker,\(\displaystyle C\)=1,\(\displaystyle C\)=2,\(\displaystyle C\)=3},
y grid style={darkgray176},
ylabel={Reward},
ymin=0, ymax=5,
ytick style={color=black},
ytick={0,1,2,3,4,5},
yticklabels={-10,-9,-8,-7,-6,-5}
]
\draw[draw=black,fill=darkseagreen,very thin] (axis cs:-0.3,0) rectangle (axis cs:-0.1,3.35261747367);
\draw[draw=black,fill=darkseagreen,very thin] (axis cs:0.7,0) rectangle (axis cs:0.9,2.74375780840471);
\draw[draw=black,fill=darkseagreen,very thin] (axis cs:1.7,0) rectangle (axis cs:1.9,2.08063143484891);
\draw[draw=black,fill=darkseagreen,very thin] (axis cs:2.7,0) rectangle (axis cs:2.9,1.48938500469201);
\draw[draw=black,fill=deepskyblue,very thin] (axis cs:-0.1,0) rectangle (axis cs:0.1,3.4500648402125);
\draw[draw=black,fill=deepskyblue,very thin] (axis cs:0.9,0) rectangle (axis cs:1.1,2.99917325928728);
\draw[draw=black,fill=deepskyblue,very thin] (axis cs:1.9,0) rectangle (axis cs:2.1,2.52488931148059);
\draw[draw=black,fill=deepskyblue,very thin] (axis cs:2.9,0) rectangle (axis cs:3.1,1.90610342499193);
\draw[draw=black,fill=tomato,very thin] (axis cs:0.1,0) rectangle (axis cs:0.3,3.40985975819437);
\draw[draw=black,fill=tomato,very thin] (axis cs:1.1,0) rectangle (axis cs:1.3,3.35216496201073);
\draw[draw=black,fill=tomato,very thin] (axis cs:2.1,0) rectangle (axis cs:2.3,3.21343708102598);
\draw[draw=black,fill=tomato,very thin] (axis cs:3.1,0) rectangle (axis cs:3.3,2.653874062927);
\path [draw=black, semithick]
(axis cs:-0.2,3.25868214378157)
--(axis cs:-0.2,3.44655280355844);

\path [draw=black, semithick]
(axis cs:0.8,2.03877443644111)
--(axis cs:0.8,3.44874118036832);

\path [draw=black, semithick]
(axis cs:1.8,0.932569971338992)
--(axis cs:1.8,3.22869289835882);

\path [draw=black, semithick]
(axis cs:2.8,0.0407605691640502)
--(axis cs:2.8,2.93800944021997);

\addplot [semithick, black, mark=-, mark size=6, mark options={solid}, only marks, forget plot]
table {%
-0.2 3.25868214378157
0.8 2.03877443644111
1.8 0.932569971338992
2.8 0.0407605691640502
};
\addplot [semithick, black, mark=-, mark size=6, mark options={solid}, only marks, forget plot]
table {%
-0.2 3.44655280355844
0.8 3.44874118036832
1.8 3.22869289835882
2.8 2.93800944021997
};
\path [draw=black, semithick]
(axis cs:0,3.34655849761189)
--(axis cs:0,3.5535711828131);

\path [draw=black, semithick]
(axis cs:1,2.50047030175051)
--(axis cs:1,3.49787621682405);

\path [draw=black, semithick]
(axis cs:2,1.73462471441415)
--(axis cs:2,3.31515390854703);

\path [draw=black, semithick]
(axis cs:3,0.568479023512618)
--(axis cs:3,3.24372782647124);

\addplot [semithick, black, mark=-, mark size=6, mark options={solid}, only marks, forget plot]
table {%
0 3.34655849761189
1 2.50047030175051
2 1.73462471441415
3 0.568479023512618
};
\addplot [semithick, black, mark=-, mark size=6, mark options={solid}, only marks, forget plot]
table {%
0 3.5535711828131
1 3.49787621682405
2 3.31515390854703
3 3.24372782647124
};
\path [draw=black, semithick]
(axis cs:0.2,3.27311029544467)
--(axis cs:0.2,3.54660922094408);

\path [draw=black, semithick]
(axis cs:1.2,3.20976437687065)
--(axis cs:1.2,3.49456554715081);

\path [draw=black, semithick]
(axis cs:2.2,2.97023184045904)
--(axis cs:2.2,3.45664232159291);

\path [draw=black, semithick]
(axis cs:3.2,1.60816453726033)
--(axis cs:3.2,3.69958358859367);

\addplot [semithick, black, mark=-, mark size=6, mark options={solid}, only marks, forget plot]
table {%
0.2 3.27311029544467
1.2 3.20976437687065
2.2 2.97023184045904
3.2 1.60816453726033
};
\addplot [semithick, black, mark=-, mark size=6, mark options={solid}, only marks, forget plot]
table {%
0.2 3.54660922094408
1.2 3.49456554715081
2.2 3.45664232159291
3.2 3.69958358859367
};
\addplot [semithick, green01270, mark=*, mark size=2.5, mark options={solid}]
table {%
-0.2 3.35261747367
0.8 2.74375780840471
1.8 2.08063143484891
2.8 1.48938500469201
};
\addlegendentry{Vanilla}
\addplot [semithick, blue, mark=x, mark size=2.5, mark options={solid}]
table {%
0 3.4500648402125
1 2.99917325928728
2 2.52488931148059
3 1.90610342499193
};
\addlegendentry{AT}
\addplot [semithick, red, mark=square*, mark size=2.5, mark options={solid}]
table {%
0.2 3.40985975819437
1.2 3.35216496201073
2.2 3.21343708102598
3.2 2.653874062927
};
\addlegendentry{AME (Ours)}
\end{axis}

\end{tikzpicture}

%% file: figs/MarketEnv/MarketEnv_Attacker_Dist_permute.tex
\begin{tikzpicture}

\definecolor{darkgray176}{RGB}{176,176,176}
\definecolor{darkseagreen}{RGB}{143,188,143}
\definecolor{deepskyblue}{RGB}{0,191,255}
\definecolor{green01270}{RGB}{0,127,0}
\definecolor{tomato}{RGB}{255,99,71}

\begin{axis}[
legend cell align={left},
legend style={fill opacity=1, draw opacity=1, text opacity=1, draw=none, fill=none},
tick align=outside,
tick pos=left,
x grid style={darkgray176},
xmin=-0.48, xmax=3.48,
xtick style={color=black},
xtick={0,1,2,3},
xticklabels={No Attacker,\(\displaystyle C\)=1,\(\displaystyle C\)=2,\(\displaystyle C\)=3},
y grid style={darkgray176},
ylabel={Reward},
ymin=0, ymax=5,
ytick style={color=black},
ytick={0,1,2,3,4,5},
yticklabels={-10,-9,-8,-7,-6,-5}
]
\draw[draw=black,fill=darkseagreen,very thin] (axis cs:-0.3,0) rectangle (axis cs:-0.1,3.35261747367);
\draw[draw=black,fill=darkseagreen,very thin] (axis cs:0.7,0) rectangle (axis cs:0.9,3.28434203029404);
\draw[draw=black,fill=darkseagreen,very thin] (axis cs:1.7,0) rectangle (axis cs:1.9,3.13748179722622);
\draw[draw=black,fill=darkseagreen,very thin] (axis cs:2.7,0) rectangle (axis cs:2.9,2.89991624715801);
\draw[draw=black,fill=deepskyblue,very thin] (axis cs:-0.1,0) rectangle (axis cs:0.1,3.4500648402125);
\draw[draw=black,fill=deepskyblue,very thin] (axis cs:0.9,0) rectangle (axis cs:1.1,3.37014332286767);
\draw[draw=black,fill=deepskyblue,very thin] (axis cs:1.9,0) rectangle (axis cs:2.1,3.21814469906741);
\draw[draw=black,fill=deepskyblue,very thin] (axis cs:2.9,0) rectangle (axis cs:3.1,2.99655476173624);
\draw[draw=black,fill=tomato,very thin] (axis cs:0.1,0) rectangle (axis cs:0.3,3.40985975819437);
\draw[draw=black,fill=tomato,very thin] (axis cs:1.1,0) rectangle (axis cs:1.3,3.39434917682172);
\draw[draw=black,fill=tomato,very thin] (axis cs:2.1,0) rectangle (axis cs:2.3,3.35078436712512);
\draw[draw=black,fill=tomato,very thin] (axis cs:3.1,0) rectangle (axis cs:3.3,3.26456770226101);
\path [draw=black, semithick]
(axis cs:-0.2,3.25868214378157)
--(axis cs:-0.2,3.44655280355844);

\path [draw=black, semithick]
(axis cs:0.8,3.15646236819114)
--(axis cs:0.8,3.41222169239693);

\path [draw=black, semithick]
(axis cs:1.8,2.90209896874265)
--(axis cs:1.8,3.37286462570979);

\path [draw=black, semithick]
(axis cs:2.8,2.43071301169277)
--(axis cs:2.8,3.36911948262325);

\addplot [semithick, black, mark=-, mark size=6, mark options={solid}, only marks, forget plot]
table {%
-0.2 3.25868214378157
0.8 3.15646236819114
1.8 2.90209896874265
2.8 2.43071301169277
};
\addplot [semithick, black, mark=-, mark size=6, mark options={solid}, only marks, forget plot]
table {%
-0.2 3.44655280355844
0.8 3.41222169239693
1.8 3.37286462570979
2.8 3.36911948262325
};
\path [draw=black, semithick]
(axis cs:0,3.34655849761189)
--(axis cs:0,3.5535711828131);

\path [draw=black, semithick]
(axis cs:1,3.23976895231343)
--(axis cs:1,3.50051769342192);

\path [draw=black, semithick]
(axis cs:2,2.97143789149265)
--(axis cs:2,3.46485150664217);

\path [draw=black, semithick]
(axis cs:3,2.5556927509884)
--(axis cs:3,3.43741677248408);

\addplot [semithick, black, mark=-, mark size=6, mark options={solid}, only marks, forget plot]
table {%
0 3.34655849761189
1 3.23976895231343
2 2.97143789149265
3 2.5556927509884
};
\addplot [semithick, black, mark=-, mark size=6, mark options={solid}, only marks, forget plot]
table {%
0 3.5535711828131
1 3.50051769342192
2 3.46485150664217
3 3.43741677248408
};
\path [draw=black, semithick]
(axis cs:0.2,3.27311029544467)
--(axis cs:0.2,3.54660922094408);

\path [draw=black, semithick]
(axis cs:1.2,3.27014195681272)
--(axis cs:1.2,3.51855639683071);

\path [draw=black, semithick]
(axis cs:2.2,3.21808978275711)
--(axis cs:2.2,3.48347895149312);

\path [draw=black, semithick]
(axis cs:3.2,3.06923114487888)
--(axis cs:3.2,3.45990425964315);

\addplot [semithick, black, mark=-, mark size=6, mark options={solid}, only marks, forget plot]
table {%
0.2 3.27311029544467
1.2 3.27014195681272
2.2 3.21808978275711
3.2 3.06923114487888
};
\addplot [semithick, black, mark=-, mark size=6, mark options={solid}, only marks, forget plot]
table {%
0.2 3.54660922094408
1.2 3.51855639683071
2.2 3.48347895149312
3.2 3.45990425964315
};
\addplot [semithick, green01270, mark=*, mark size=2.5, mark options={solid}]
table {%
-0.2 3.35261747367
0.8 3.28434203029404
1.8 3.13748179722622
2.8 2.89991624715801
};
\addlegendentry{Vanilla}
\addplot [semithick, blue, mark=x, mark size=2.5, mark options={solid}]
table {%
0 3.4500648402125
1 3.37014332286767
2 3.21814469906741
3 2.99655476173624
};
\addlegendentry{AT}
\addplot [semithick, red, mark=square*, mark size=2.5, mark options={solid}]
table {%
0.2 3.40985975819437
1.2 3.39434917682172
2.2 3.35078436712512
3.2 3.26456770226101
};
\addlegendentry{AME (Ours)}
\addplot [semithick, red, dashed, forget plot]
table {%
-0.48 2.46209120815972
3.48 2.46209120815972
};
\end{axis}

\end{tikzpicture}

%% file: figs/MarketEnv/MarketEnv_Attacker_Dist_flip.tex
\begin{tikzpicture}

\definecolor{darkgray176}{RGB}{176,176,176}
\definecolor{darkseagreen}{RGB}{143,188,143}
\definecolor{deepskyblue}{RGB}{0,191,255}
\definecolor{green01270}{RGB}{0,127,0}
\definecolor{tomato}{RGB}{255,99,71}

\begin{axis}[
legend cell align={left},
legend style={fill opacity=1, draw opacity=1, text opacity=1, draw=none, fill=none},
tick align=outside,
tick pos=left,
x grid style={darkgray176},
xmin=-0.48, xmax=3.48,
xtick style={color=black},
xtick={0,1,2,3},
xticklabels={No Attacker,\(\displaystyle C\)=1,\(\displaystyle C\)=2,\(\displaystyle C\)=3},
y grid style={darkgray176},
ylabel={Reward},
ymin=0, ymax=5,
ytick style={color=black},
ytick={0,1,2,3,4,5},
yticklabels={-10,-9,-8,-7,-6,-5}
]
\draw[draw=black,fill=darkseagreen,very thin] (axis cs:-0.3,0) rectangle (axis cs:-0.1,3.35261747367);
\draw[draw=black,fill=darkseagreen,very thin] (axis cs:0.7,0) rectangle (axis cs:0.9,3.14737255002094);
\draw[draw=black,fill=darkseagreen,very thin] (axis cs:1.7,0) rectangle (axis cs:1.9,2.71399369105815);
\draw[draw=black,fill=darkseagreen,very thin] (axis cs:2.7,0) rectangle (axis cs:2.9,2.08030285437827);
\draw[draw=black,fill=deepskyblue,very thin] (axis cs:-0.1,0) rectangle (axis cs:0.1,3.4500648402125);
\draw[draw=black,fill=deepskyblue,very thin] (axis cs:0.9,0) rectangle (axis cs:1.1,3.21284598463934);
\draw[draw=black,fill=deepskyblue,very thin] (axis cs:1.9,0) rectangle (axis cs:2.1,2.7996878157636);
\draw[draw=black,fill=deepskyblue,very thin] (axis cs:2.9,0) rectangle (axis cs:3.1,2.22517185195052);
\draw[draw=black,fill=tomato,very thin] (axis cs:0.1,0) rectangle (axis cs:0.3,3.40985975819437);
\draw[draw=black,fill=tomato,very thin] (axis cs:1.1,0) rectangle (axis cs:1.3,3.36995294614631);
\draw[draw=black,fill=tomato,very thin] (axis cs:2.1,0) rectangle (axis cs:2.3,3.24288030834742);
\draw[draw=black,fill=tomato,very thin] (axis cs:3.1,0) rectangle (axis cs:3.3,2.76326187919382);
\path [draw=black, semithick]
(axis cs:-0.2,3.25868214378157)
--(axis cs:-0.2,3.44655280355844);

\path [draw=black, semithick]
(axis cs:0.8,2.91798357685843)
--(axis cs:0.8,3.37676152318345);

\path [draw=black, semithick]
(axis cs:1.8,2.07127851487178)
--(axis cs:1.8,3.35670886724451);

\path [draw=black, semithick]
(axis cs:2.8,0.840738584938843)
--(axis cs:2.8,3.3198671238177);

\addplot [semithick, black, mark=-, mark size=6, mark options={solid}, only marks, forget plot]
table {%
-0.2 3.25868214378157
0.8 2.91798357685843
1.8 2.07127851487178
2.8 0.840738584938843
};
\addplot [semithick, black, mark=-, mark size=6, mark options={solid}, only marks, forget plot]
table {%
-0.2 3.44655280355844
0.8 3.37676152318345
1.8 3.35670886724451
2.8 3.3198671238177
};
\path [draw=black, semithick]
(axis cs:0,3.34655849761189)
--(axis cs:0,3.5535711828131);

\path [draw=black, semithick]
(axis cs:1,2.95409321709894)
--(axis cs:1,3.47159875217975);

\path [draw=black, semithick]
(axis cs:2,2.17498579264861)
--(axis cs:2,3.4243898388786);

\path [draw=black, semithick]
(axis cs:3,1.08494571153595)
--(axis cs:3,3.36539799236509);

\addplot [semithick, black, mark=-, mark size=6, mark options={solid}, only marks, forget plot]
table {%
0 3.34655849761189
1 2.95409321709894
2 2.17498579264861
3 1.08494571153595
};
\addplot [semithick, black, mark=-, mark size=6, mark options={solid}, only marks, forget plot]
table {%
0 3.5535711828131
1 3.47159875217975
2 3.4243898388786
3 3.36539799236509
};
\path [draw=black, semithick]
(axis cs:0.2,3.27311029544467)
--(axis cs:0.2,3.54660922094408);

\path [draw=black, semithick]
(axis cs:1.2,3.23747547537105)
--(axis cs:1.2,3.50243041692157);

\path [draw=black, semithick]
(axis cs:2.2,3.02436330283669)
--(axis cs:2.2,3.46139731385816);

\path [draw=black, semithick]
(axis cs:3.2,1.90147406729153)
--(axis cs:3.2,3.62504969109611);

\addplot [semithick, black, mark=-, mark size=6, mark options={solid}, only marks, forget plot]
table {%
0.2 3.27311029544467
1.2 3.23747547537105
2.2 3.02436330283669
3.2 1.90147406729153
};
\addplot [semithick, black, mark=-, mark size=6, mark options={solid}, only marks, forget plot]
table {%
0.2 3.54660922094408
1.2 3.50243041692157
2.2 3.46139731385816
3.2 3.62504969109611
};
\addplot [semithick, green01270, mark=*, mark size=2.5, mark options={solid}]
table {%
-0.2 3.35261747367
0.8 3.14737255002094
1.8 2.71399369105815
2.8 2.08030285437827
};
\addlegendentry{Vanilla}
\addplot [semithick, blue, mark=x, mark size=2.5, mark options={solid}]
table {%
0 3.4500648402125
1 3.21284598463934
2 2.7996878157636
3 2.22517185195052
};
\addlegendentry{AT}
\addplot [semithick, red, mark=square*, mark size=2.5, mark options={solid}]
table {%
0.2 3.40985975819437
1.2 3.36995294614631
2.2 3.24288030834742
3.2 2.76326187919382
};
\addlegendentry{AME (Ours)}
\addplot [semithick, red, dashed, forget plot]
table {%
-0.48 2.46209120815972
3.48 2.46209120815972
};
\end{axis}

\end{tikzpicture}

%% file: figs/MarketEnv/MarketEnv_Attacker_Dist_swap.tex
\begin{tikzpicture}

\definecolor{darkgray176}{RGB}{176,176,176}
\definecolor{darkseagreen}{RGB}{143,188,143}
\definecolor{deepskyblue}{RGB}{0,191,255}
\definecolor{green01270}{RGB}{0,127,0}
\definecolor{tomato}{RGB}{255,99,71}

\begin{axis}[
legend cell align={left},
legend style={fill opacity=1, draw opacity=1, text opacity=1, draw=none, fill=none},
tick align=outside,
tick pos=left,
x grid style={darkgray176},
xmin=-0.48, xmax=3.48,
xtick style={color=black},
xtick={0,1,2,3},
xticklabels={No Attacker,\(\displaystyle C\)=1,\(\displaystyle C\)=2,\(\displaystyle C\)=3},
y grid style={darkgray176},
ylabel={Reward},
ymin=0, ymax=5,
ytick style={color=black},
ytick={0,1,2,3,4,5},
yticklabels={-10,-9,-8,-7,-6,-5}
]
\draw[draw=black,fill=darkseagreen,very thin] (axis cs:-0.3,0) rectangle (axis cs:-0.1,3.35261747367);
\draw[draw=black,fill=darkseagreen,very thin] (axis cs:0.7,0) rectangle (axis cs:0.9,3.17090483036452);
\draw[draw=black,fill=darkseagreen,very thin] (axis cs:1.7,0) rectangle (axis cs:1.9,2.78225879975945);
\draw[draw=black,fill=darkseagreen,very thin] (axis cs:2.7,0) rectangle (axis cs:2.9,2.21266128472267);
\draw[draw=black,fill=deepskyblue,very thin] (axis cs:-0.1,0) rectangle (axis cs:0.1,3.4500648402125);
\draw[draw=black,fill=deepskyblue,very thin] (axis cs:0.9,0) rectangle (axis cs:1.1,3.23551278573123);
\draw[draw=black,fill=deepskyblue,very thin] (axis cs:1.9,0) rectangle (axis cs:2.1,2.8655802589046);
\draw[draw=black,fill=deepskyblue,very thin] (axis cs:2.9,0) rectangle (axis cs:3.1,2.34848948927625);
\draw[draw=black,fill=tomato,very thin] (axis cs:0.1,0) rectangle (axis cs:0.3,3.40985975819437);
\draw[draw=black,fill=tomato,very thin] (axis cs:1.1,0) rectangle (axis cs:1.3,3.3759057670942);
\draw[draw=black,fill=tomato,very thin] (axis cs:2.1,0) rectangle (axis cs:2.3,3.27030511472095);
\draw[draw=black,fill=tomato,very thin] (axis cs:3.1,0) rectangle (axis cs:3.3,2.87237030776049);
\path [draw=black, semithick]
(axis cs:-0.2,3.25868214378157)
--(axis cs:-0.2,3.44655280355844);

\path [draw=black, semithick]
(axis cs:0.8,2.96320133653874)
--(axis cs:0.8,3.3786083241903);

\path [draw=black, semithick]
(axis cs:1.8,2.20534936625974)
--(axis cs:1.8,3.35916823325916);

\path [draw=black, semithick]
(axis cs:2.8,1.10016439246527)
--(axis cs:2.8,3.32515817698006);

\addplot [semithick, black, mark=-, mark size=6, mark options={solid}, only marks, forget plot]
table {%
-0.2 3.25868214378157
0.8 2.96320133653874
1.8 2.20534936625974
2.8 1.10016439246527
};
\addplot [semithick, black, mark=-, mark size=6, mark options={solid}, only marks, forget plot]
table {%
-0.2 3.44655280355844
0.8 3.3786083241903
1.8 3.35916823325916
2.8 3.32515817698006
};
\path [draw=black, semithick]
(axis cs:0,3.34655849761189)
--(axis cs:0,3.5535711828131);

\path [draw=black, semithick]
(axis cs:1,2.99883402936828)
--(axis cs:1,3.47219154209418);

\path [draw=black, semithick]
(axis cs:2,2.30522409078178)
--(axis cs:2,3.42593642702742);

\path [draw=black, semithick]
(axis cs:3,1.32272432674797)
--(axis cs:3,3.37425465180452);

\addplot [semithick, black, mark=-, mark size=6, mark options={solid}, only marks, forget plot]
table {%
0 3.34655849761189
1 2.99883402936828
2 2.30522409078178
3 1.32272432674797
};
\addplot [semithick, black, mark=-, mark size=6, mark options={solid}, only marks, forget plot]
table {%
0 3.5535711828131
1 3.47219154209418
2 3.42593642702742
3 3.37425465180452
};
\path [draw=black, semithick]
(axis cs:0.2,3.27311029544467)
--(axis cs:0.2,3.54660922094408);

\path [draw=black, semithick]
(axis cs:1.2,3.24390253407565)
--(axis cs:1.2,3.50790900011275);

\path [draw=black, semithick]
(axis cs:2.2,3.07441352795941)
--(axis cs:2.2,3.4661967014825);

\path [draw=black, semithick]
(axis cs:3.2,2.14792237417153)
--(axis cs:3.2,3.59681824134944);

\addplot [semithick, black, mark=-, mark size=6, mark options={solid}, only marks, forget plot]
table {%
0.2 3.27311029544467
1.2 3.24390253407565
2.2 3.07441352795941
3.2 2.14792237417153
};
\addplot [semithick, black, mark=-, mark size=6, mark options={solid}, only marks, forget plot]
table {%
0.2 3.54660922094408
1.2 3.50790900011275
2.2 3.4661967014825
3.2 3.59681824134944
};
\addplot [semithick, green01270, mark=*, mark size=2.5, mark options={solid}]
table {%
-0.2 3.35261747367
0.8 3.17090483036452
1.8 2.78225879975945
2.8 2.21266128472267
};
\addlegendentry{Vanilla}
\addplot [semithick, blue, mark=x, mark size=2.5, mark options={solid}]
table {%
0 3.4500648402125
1 3.23551278573123
2 2.8655802589046
3 2.34848948927625
};
\addlegendentry{AT}
\addplot [semithick, red, mark=square*, mark size=2.5, mark options={solid}]
table {%
0.2 3.40985975819437
1.2 3.3759057670942
2.2 3.27030511472095
3.2 2.87237030776049
};
\addlegendentry{AME (Ours)}
\addplot [semithick, red, dashed, forget plot]
table {%
-0.48 2.46209120815972
3.48 2.46209120815972
};
\end{axis}

\end{tikzpicture}

%% file: figs/mnist/MNIST_Adaptive_Attacker_Precision.tex
\begin{tikzpicture}

\definecolor{darkgray176}{RGB}{176,176,176}
\definecolor{darkseagreen}{RGB}{143,188,143}
\definecolor{deepskyblue}{RGB}{0,191,255}
\definecolor{green01270}{RGB}{0,127,0}
\definecolor{lightgray204}{RGB}{204,204,204}
\definecolor{tomato}{RGB}{255,99,71}

\begin{axis}[
legend cell align={left},
legend style={fill opacity=1, draw opacity=1, text opacity=1, draw=lightgray204},
tick align=outside,
ytick pos = left,
xtick pos = lower,
x grid style={darkgray176},
xmin=-0.48, xmax=3.48,
xtick style={color=black},
xtick={0,1,2,3},
xticklabels={No Attacker,C=1,C=2,C=3},
y grid style={darkgray176},
ylabel={Precision},
ymin=0.7, ymax=1.03,
ytick style={color=black}
]
\draw[draw=black,fill=darkseagreen] (axis cs:-0.3,0) rectangle (axis cs:-0.1,0.9428);
\draw[draw=black,fill=darkseagreen] (axis cs:0.7,0) rectangle (axis cs:0.9,0.9182);
\draw[draw=black,fill=darkseagreen] (axis cs:1.7,0) rectangle (axis cs:1.9,0.8568);
\draw[draw=black,fill=darkseagreen] (axis cs:2.7,0) rectangle (axis cs:2.9,0.7796);
\draw[draw=black,fill=deepskyblue] (axis cs:-0.1,0) rectangle (axis cs:0.1,0.935);
\draw[draw=black,fill=deepskyblue] (axis cs:0.9,0) rectangle (axis cs:1.1,0.9);
\draw[draw=black,fill=deepskyblue] (axis cs:1.9,0) rectangle (axis cs:2.1,0.822);
\draw[draw=black,fill=deepskyblue] (axis cs:2.9,0) rectangle (axis cs:3.1,0.744);
\draw[draw=black,fill=tomato] (axis cs:0.1,0) rectangle (axis cs:0.3,0.9324);
\draw[draw=black,fill=tomato] (axis cs:1.1,0) rectangle (axis cs:1.3,0.9278);
\draw[draw=black,fill=tomato] (axis cs:2.1,0) rectangle (axis cs:2.3,0.9222);
\draw[draw=black,fill=tomato] (axis cs:3.1,0) rectangle (axis cs:3.3,0.9046);
\path [draw=black, semithick]
(axis cs:-0.2,0.937925576957219)
--(axis cs:-0.2,0.947674423042782);

\path [draw=black, semithick]
(axis cs:0.8,0.899096073701985)
--(axis cs:0.8,0.937303926298015);

\path [draw=black, semithick]
(axis cs:1.8,0.825800645168004)
--(axis cs:1.8,0.887799354831996);

\path [draw=black, semithick]
(axis cs:2.8,0.741329384640432)
--(axis cs:2.8,0.817870615359568);

\path [draw=black, semithick]
(axis cs:0,0.927624364434166)
--(axis cs:0,0.942375635565834);

\path [draw=black, semithick]
(axis cs:1,0.893675444679663)
--(axis cs:1,0.906324555320337);

\path [draw=black, semithick]
(axis cs:2,0.782301133517442)
--(axis cs:2,0.861698866482558);

\path [draw=black, semithick]
(axis cs:3,0.733801960972814)
--(axis cs:3,0.754198039027186);

\path [draw=black, semithick]
(axis cs:0.2,0.924460226703488)
--(axis cs:0.2,0.940339773296512);

\path [draw=black, semithick]
(axis cs:1.2,0.916338760974485)
--(axis cs:1.2,0.939261239025516);

\path [draw=black, semithick]
(axis cs:2.2,0.910226696362323)
--(axis cs:2.2,0.934173303637676);

\path [draw=black, semithick]
(axis cs:3.2,0.896077089698935)
--(axis cs:3.2,0.913122910301065);

\addplot [semithick, black, mark=-, mark size=6, mark options={solid}, only marks, forget plot]
table {%
-0.2 0.937925576957219
0.8 0.899096073701985
1.8 0.825800645168004
2.8 0.741329384640432
};
\addplot [semithick, black, mark=-, mark size=6, mark options={solid}, only marks, forget plot]
table {%
-0.2 0.947674423042782
0.8 0.937303926298015
1.8 0.887799354831996
2.8 0.817870615359568
};
\addplot [semithick, black, mark=-, mark size=6, mark options={solid}, only marks, forget plot]
table {%
0 0.927624364434166
1 0.893675444679663
2 0.782301133517442
3 0.733801960972814
};
\addplot [semithick, black, mark=-, mark size=6, mark options={solid}, only marks, forget plot]
table {%
0 0.942375635565834
1 0.906324555320337
2 0.861698866482558
3 0.754198039027186
};
\addplot [semithick, black, mark=-, mark size=6, mark options={solid}, only marks, forget plot]
table {%
0.2 0.924460226703488
1.2 0.916338760974485
2.2 0.910226696362323
3.2 0.896077089698935
};
\addplot [semithick, black, mark=-, mark size=6, mark options={solid}, only marks, forget plot]
table {%
0.2 0.940339773296512
1.2 0.939261239025516
2.2 0.934173303637676
3.2 0.913122910301065
};
\addplot [semithick, green01270, mark=*, mark size=3, mark options={solid}]
table {%
-0.2 0.9428
0.8 0.9182
1.8 0.8568
2.8 0.7796
};
\addlegendentry{Vanilla}
\addplot [semithick, blue, mark=x, mark size=3, mark options={solid}]
table {%
0 0.935
1 0.9
2 0.822
3 0.744
};
\addlegendentry{AT}
\addplot [semithick, red, mark=square*, mark size=3, mark options={solid}]
table {%
0.2 0.9324
1.2 0.9278
2.2 0.9222
3.2 0.9046
};
\addlegendentry{AME (ours)}
\end{axis}

\end{tikzpicture}

%% file: figs/FoodCollector_Ablation_k_2adv_dist.tex
\begin{tikzpicture}

\definecolor{color0}{rgb}{0.172549019607843,0.627450980392157,0.172549019607843}
\definecolor{color1}{rgb}{0.12156862745098,0.466666666666667,0.705882352941177}
\definecolor{color2}{rgb}{0.83921568627451,0.152941176470588,0.156862745098039}

\begin{axis}[
axis line style={white!15!black},
legend cell align={left},
legend style={
  fill opacity=1,
  draw opacity=1,
  text opacity=1,
  at={(0.32,0.38)},
  anchor=south west,
  draw=white!80!black
},
tick align=outside,
x grid style={white!80!black},
xlabel={Ablation Size \(\displaystyle k\)},
xmajorticks=true,
xmin=0.65, xmax=8.35,
xtick style={color=white!15!black},
xtick pos = lower,
y grid style={white!80!black},
ylabel={Reward},
ymajorticks=true,
ytick pos = left,
ymin=-90, ymax=-9.632,
ytick style={color=white!15!black}
]
\path [draw=color0, fill=color0, opacity=0.2]
(axis cs:1,-47.59)
--(axis cs:1,-49.41)
--(axis cs:2,-27.32)
--(axis cs:3,-22.21)
--(axis cs:4,-24.1)
--(axis cs:5,-15.56)
--(axis cs:6,-18.23)
--(axis cs:7,-18.34)
--(axis cs:8,-13.83)
--(axis cs:8,-13.07)
--(axis cs:8,-13.07)
--(axis cs:7,-17.26)
--(axis cs:6,-15.53)
--(axis cs:5,-14.3)
--(axis cs:4,-22.32)
--(axis cs:3,-19.91)
--(axis cs:2,-24.3)
--(axis cs:1,-47.59)
--cycle;

\path [draw=color1, fill=color1, opacity=0.2]
(axis cs:1,-46.17)
--(axis cs:1,-51.73)
--(axis cs:2,-26.51)
--(axis cs:3,-24.99)
--(axis cs:4,-26.21)
--(axis cs:5,-35.98)
--(axis cs:6,-34.02)
--(axis cs:7,-40.93)
--(axis cs:8,-51.51)
--(axis cs:8,-44.63)
--(axis cs:8,-44.63)
--(axis cs:7,-36.67)
--(axis cs:6,-30.7)
--(axis cs:5,-32.5)
--(axis cs:4,-23.59)
--(axis cs:3,-18.59)
--(axis cs:2,-22.79)
--(axis cs:1,-46.17)
--cycle;

\path [draw=color2, fill=color2, opacity=0.2]
(axis cs:1,-44.39)
--(axis cs:1,-52.61)
--(axis cs:2,-32.9)
--(axis cs:3,-67.19)
--(axis cs:4,-66.92)
--(axis cs:5,-68.79)
--(axis cs:6,-76.31)
--(axis cs:7,-79.92)
--(axis cs:8,-81.83)
--(axis cs:8,-76.37)
--(axis cs:8,-76.37)
--(axis cs:7,-77.74)
--(axis cs:6,-68.53)
--(axis cs:5,-62.25)
--(axis cs:4,-58.9)
--(axis cs:3,-59.39)
--(axis cs:2,-23.28)
--(axis cs:1,-44.39)
--cycle;

\addplot [semithick, color0, mark=x, mark size=3, mark options={solid}]
table {%
1 -48.5
2 -25.81
3 -21.06
4 -23.21
5 -14.93
6 -16.88
7 -17.8
8 -13.45
};
\addlegendentry{No Attacker}
\addplot [semithick, color1, mark=*, mark size=3, mark options={solid}]
table {%
1 -48.95
2 -24.65
3 -21.79
4 -24.9
5 -34.24
6 -32.36
7 -38.8
8 -48.07
};
\addlegendentry{Non-adaptive Attack}
\addplot [semithick, color2, mark=square*, mark size=3, mark options={solid}]
table {%
1 -48.5
2 -28.09
3 -63.29
4 -62.91
5 -65.52
6 -72.42
7 -78.83
8 -79.1
};
\addlegendentry{Adaptive Attack}
\addplot [semithick, green!50!black, dashed, forget plot]
table {%
0.75 -79.1
8.75 -79.1
};
\end{axis}
\node[above,green!50!black] at (3,0.1) {Vanilla under Adaptive Attack};

\end{tikzpicture}

%% file: figs/FoodCollector_Ablation_k_sample_dist.tex
\begin{tikzpicture}

\definecolor{color0}{rgb}{0.83921568627451,0.152941176470588,0.156862745098039}
\definecolor{color1}{rgb}{0.12156862745098,0.466666666666667,0.705882352941177}

\begin{axis}[
axis line style={white!15!black},
legend cell align={left},
legend style={
  fill opacity=1,
  draw opacity=1,
  text opacity=1,
  at={(0.09,0.4)},
  anchor=west,
  draw=white!80!black
},
tick align=outside,
x grid style={white!80!black},
xlabel={Ablation Sample Size \(\displaystyle D\)},
xmin=0.75, xmax=6.25,
xtick style={color=white!15!black},
xtick pos = left, 
xtick={6,5,4,3,2,1},
xticklabels={1,9,15,21,25,28},
y grid style={white!80!black},
ylabel={Reward under Attack},
ytick pos = left,
ymin=-90, ymax=-20,
ytick style={color=white!15!black}
]
\path [draw=color0, semithick]
(axis cs:6,-49.86)
--(axis cs:6,-47.98);

\path [draw=color0, semithick]
(axis cs:5,-47.59)
--(axis cs:5,-41.21);

\path [draw=color0, semithick]
(axis cs:4,-52.48)
--(axis cs:4,-43.44);

\path [draw=color0, semithick]
(axis cs:3,-42.55)
--(axis cs:3,-35.95);

\path [draw=color0, semithick]
(axis cs:2,-36.79)
--(axis cs:2,-31.69);

\path [draw=color0, semithick]
(axis cs:1,-31.57)
--(axis cs:1,-25.93);

\path [draw=color1, semithick]
(axis cs:6,-30.24)
--(axis cs:6,-28.36);

\path [draw=color1, semithick]
(axis cs:5,-30.25)
--(axis cs:5,-23.87);

\path [draw=color1, semithick]
(axis cs:4,-30.18)
--(axis cs:4,-25.8);

\path [draw=color1, semithick]
(axis cs:3,-30.16)
--(axis cs:3,-25.76);

\path [draw=color1, semithick]
(axis cs:2,-26.17)
--(axis cs:2,-21.75);

\path [draw=color1, semithick]
(axis cs:1,-26.19)
--(axis cs:1,-21.95);

\addplot [semithick, color0, mark=square*, mark size=4.5, mark options={solid}]
table {%
6 -48.92
5 -44.4
4 -47.96
3 -39.25
2 -34.24
1 -28.75
};
\addlegendentry{Adaptive Attack}
\addplot [semithick, color1, mark=*, mark size=4.5, mark options={solid}]
table {%
6 -29.3
5 -27.06
4 -27.99
3 -27.96
2 -23.96
1 -24.07
};
\addlegendentry{Non-adaptive Attack}

\addplot [semithick, color0, mark=-, mark size=5, mark options={solid}, only marks]
table {%
6 -49.86
5 -47.59
4 -52.48
3 -42.55
2 -36.79
1 -31.57
};
\addplot [semithick, color0, mark=-, mark size=5, mark options={solid}, only marks]
table {%
6 -47.98
5 -41.21
4 -43.44
3 -35.95
2 -31.69
1 -25.93
};
\addplot [semithick, color1, mark=-, mark size=5, mark options={solid}, only marks]
table {%
6 -30.24
5 -30.25
4 -30.18
3 -30.16
2 -26.17
1 -26.19
};
\addplot [semithick, color1, mark=-, mark size=5, mark options={solid}, only marks]
table {%
6 -28.36
5 -23.87
4 -25.8
3 -25.76
2 -21.75
1 -21.95
};
\addplot [semithick, green!50!black, dashed, forget plot]
table {%
0.75 -79.1
6.25 -79.1
};

\end{axis}
\node[above,green!50!black] at (3,0.2) {Vanilla under Adaptive Attack};
\end{tikzpicture}

%% file: arxiv/6_discuss.tex
\section{Conclusion}
\label{sec:discuss}

This paper formulates communication attacks in MARL problems, and proposes a defense framework \ours, which is certifiably robust against multiple arbitrarily perturbed adversarial communication messages, based on randomized ablation and aggregation of messages. 
We show in theory and by experiments that \ours maintains its performance under various communication perturbations.
Even when theoretical conditions do not hold, \ours is still moderately robust and significantly outperforms other defenses baselines.
\ours is a generic defense that can be applied to any policy learning workflow, and can easily scale up to many real-world tasks. 
In Appendix~\ref{app:hyper}, we provide detailed analysis on hyperparameter selection and its relation to the trade-off between natural performance and robustness. 
The idea of \ours can also be extended to attack detection as discussed in Appendix~\ref{app:exp_detect}.

%% file: arxiv/a2_theory.tex
\section{Additional Theoretical Definitions and Analysis}
\label{app:theory}

\textbf{Relation between Communication Attack and $\boldsymbol \ell_0$ Observation Attack}\quad
The communication threat model described in Section~\ref{sec:setup_adv} is analogous to a constrained $\ell_0$ attack on policy inputs. When agent $i$ is attacked, the input space of its acting policy $\pi_i$ is $\mathcal{X}_i := \hisspace_i \times \mathcal{M}^{N-1}$. Therefore, when $C$ communication messages are corrupted, the original input $x_i$ gets perturbed to $\tilde{x}_i$. Let $d$ be the dimension of a communication message, then $x_i$ and $\tilde{x}_i$ differ by up to $dC$ dimensions, which is similar to an $\ell_0$ attack constrained in certain dimensions.

\subsection{Additional Analysis of \ours in Discrete Action Space}
\label{app:theory_dist}

\textbf{Reward Certificate}\quad
Theorem~\ref{thm:main_dist} justifies that under sufficient majority votes, the \ensname $\smpolicy$ ignores the malicious messages in $\mathbf{m}_{\mathrm{adv}}$ and executes a benign action that is suggested by some benign message combinations, even when the malicious messages are not identified.
It is important to note that, when the \ensname $\smpolicy$ selects an action $\widetilde{a}$, there must exist at least one purely benign \ksample that let the \ablname $\ablpolicy$ produce $\widetilde{a}$. Therefore, as long as $\ablpolicy$ can obtain high reward with randomly selected benign \ksamples, $\smpolicy$ can also obtain high reward with ablated adversarial communication due to its design.

Specifically, we consider a specific agent with \ablname $\ablpolicy$ and \ensname $\smpolicy$ (suppose other agents are executing fixed policies). 
Let $\nu:\mathcal{M}^{N-1}\to\mathcal{M}^{N-1}$ be an attack algorithm that perturbs at most $C$ messages in a message set. 
Let $\zeta \sim Z(P,\ablpolicy)$ be a trajectory of policy $\ablpolicy$ under no attack, i.e., $\zeta=(o^{(0)},\msg^{(0)},a^{(0)},r^{(0)},o^{(1)},\msg^{(1)},a^{(0)},r^{(0)},\cdots)$. (Recall that a \ablname $\ablpolicy$ takes in a random size-$k$ subset of $\msg\step$ and outputs action $a\step$.)
When there exists attack with $\nu$, let $\zeta_\nu \sim Z(P,\smpolicy;\nu)$ be a trajectory of policy $\smpolicy$ under communication attacks, i.e., $\zeta_\nu=(o^{(0)},\nu(\msg^{(0)}),a^{(0)},r^{(0)},o^{(1)},\nu(\msg^{(1)}),a^{(0)},r^{(0)},\cdots)$. 
For any trajectory $\zeta$, let $r(\zeta)$ be the discounted cumulative reward of this trajectory.


With the above notations, we propose the following reward certificate.

\begin{corollary}[Reward Certificate for Discrete Action Space]
\label{cor:reward_dist}
When Condition~\ref{cond:k} and Condition~\ref{cond:cons} hold at every step of execution, the cumulative reward of ensemble policy $\smpolicy$ defined in Equation~\eqref{eq:ensemble_pi_dist} under adversarial communication is no lower than the lowest cumulative reward that the ablation policy $\ablpolicy$ can obtain with randomly selected \ksamples under no attacks, i.e.,
\begin{equation}
\begin{aligned}
    \min_{\zeta_\nu\sim Z(P,\smpolicy;\nu)} r(\zeta_\nu) \geq \min_{\zeta\sim Z(P,\ablpolicy)} r(\zeta),
\end{aligned}
\end{equation}
for any attacker $\nu$ satisfying Assumption~\ref{assum:num} ($C<\frac{N-1}{2}$).
\end{corollary}

\textbf{Remarks.} 
(1) The certificate holds for any attack algorithm $\nu$ with $C<\frac{N-1}{2}$. 
(2) The \ablname $\ablpolicy$ has extra randomness from the sampling of \ksamples. That is, at every step, $\ablpolicy$ takes a uniformly randomly selected \ksample from $\allmsgk$. Therefore, the $\min_\zeta$ in the RHS considers the worst-case message sampling in the clean environment without attacks. Since $\smpolicy$ always takes actions selected by some purely-benign message combinations, the trajectory generated by $\smpolicy$ can also be produced by the \ablname.
(3) Note that the RHS ($\min_{\zeta\sim Z(P,\ablpolicy)} r(\zeta)$) can be approximately estimated by executing $\ablpolicy$ during training time, so that the test-time performance of $\smpolicy$ is guaranteed to be no lower than this value, even if there are up to $C$ corrupted messages at every step.


Appendix~\ref{app:proof} provides detailed proofs for the above theoretical results.

\subsection{Additional Analysis of \ours in Continuous Action Space}
\label{app:theory_cont}



Note that Theorem~\ref{thm:main_cont} certifies that the selected action is in a set of actions that are close to benign actions $\mathsf{Range}(\safeaction)$, but does not make any assumption on this set. Next we interpret this result in details.

\textbf{How to Understand $\mathsf{Range}(\safeaction)$?}\quad
Theoretically, $\mathsf{Range}(\safeaction)$ is a set of actions that are coordinate-wise bounded by base actions resulted from purely benign \ksamples. In many practical problems, it is reasonable to assume that actions in $\mathsf{Range}(\safeaction)$ are relatively safe, especially when benign actions in $\safeaction$ are concentrated. The following examples illustrate some scenarios where actions in $\mathsf{Range}(\safeaction)$ are relatively good.

\begin{enumerate}
    \item If the action denotes the price a seller sells its product in a market, and the communication messages are the transaction signals in an information pool, then $\mathsf{Range}(\safeaction)$ is a price range that is determined based on purely benign messages. Therefore, the seller will set a reasonable price without being influenced by a malicious message. 
    \item If the action denotes the driving speed, and benign message combinations have suggested driving at 40 mph or driving at 50 mph, then driving at 45 mph is also safe. 
    \item If the action is a vector denoting movements of all joints of a robot (as in many MuJoCo tasks), and two slightly different joint movements are suggested by two benign message combinations, then an action that does not exceed the range of the two benign movements at every joint is likely to be safe as well.
\end{enumerate}  

The above examples show by intuition why the \ensname can be regarded as a relatively robust policy. 
However, in extreme cases where there exists ``caveat'' in $\mathsf{Range}(\safeaction)$, taking an action in this set may also be unsafe. To quantify the influence of $\mathsf{Range}(\safeaction)$ on the long-term reward, we next analyze the cumulative reward of the \ensname in the continuous-action case.

\textbf{How Does $\mathsf{Range}(\safeaction)$ Lead to A Reward Certificate?}\quad
Different from the discrete-action case, the \ensname $\smpolicy$ in a continuous action space may take actions not in $\safeaction$ such that it generates trajectories not seen by the \ablname $\ablpolicy$. However, since the action of $\smpolicy$ is guaranteed to stay in $\mathsf{Range}(\safeaction)$, we can bound the difference between the value of $\smpolicy$ and the value of $\ablpolicy$, and how large the different is depends on some properties of $\mathsf{Range}(\safeaction)$.

Concretely, Let $R$ and $P$ be the reward function and transition probability function of the current agent when the other agents execute fixed policies. So $R(s,a)$ is the immediate reward of taking action $a$ at state $s$, and $P(s^\prime|s,a)$ is the probability of transitioning to state $s^\prime$ from $s$ by taking action $a$. (Note that $s$ is the underlying state which may not be observed by the agent.)

\begin{definition}[Dynamics Discrepancy of $\ablpolicy$]
\label{def:disc}
A \ablname $\ablpolicy$ is called $\epsilon_R$,$\epsilon_P$-discrepant if $\epsilon_R$, $\epsilon_P$ are the smallest values such that for any $s\in\mathcal{S}$ and the corresponding \safename set $\safeaction$, we have $\forall a_1, a_2 \in \mathsf{Range}(\safeaction)$, 
\begin{align}
    & |R(s,a_1) - R(s,a_2)| \leq \epsilon_R, \\
    & \int | P(s^\prime| s, a_1) - P(s^\prime| s, a_2) | \mathrm{d}s^\prime \leq \epsilon_P. \label{eq:p_disc}
\end{align}
\end{definition}

\textbf{Remarks.}
(1) Equation~\eqref{eq:p_disc} is equivalent to that the total variance distance between $P(\cdot| s, a_1)$ and $P(\cdot| s, a_2)$ is less than or equal to $\epsilon_P/2$.
(2) For any environment with bounded reward, $\epsilon_R$ and $\epsilon_P$ always exist. 

Definition~\ref{def:disc} characterizes how different the local dynamics of actions in $\mathsf{Range}(\safeaction)$ are, over all possible states. If $\mathsf{Range}(\safeaction)$ is small and the environment is relatively smooth, then taking different actions within this range will not result in very different future rewards. The theorem below shows a reward certificate for the \ensname $\smpolicy$.


\begin{theorem}[Reward Certificate for Continuous Action Space]
\label{thm:reward_cont}
Let $V^{\ablpolicy}(s)$ be the clean value (discounted cumulative reward) of $\ablpolicy$ starting from state $s$ under no attack; let $\tilde{V}^{\smpolicy}_\nu(s)$ be the value of $\smpolicy$ starting from state $s$ under attack algorithm $\nu$, where $\nu$ satisfies Assumption~\ref{assum:num}; let $k$ be an ablation size satisfying Condition~\ref{cond:k}.
If $\ablpolicy$ is $\epsilon_R$,$\epsilon_P$-discrepant, then for any state $s\in\mathcal{S}$, we have
\begin{equation}
    \min_{\nu} \tilde{V}^{\smpolicy}_\nu(s) \geq V^{\ablpolicy}(s) - \frac{\epsilon_R+\gamma V_{\max}\epsilon_P}{1-\gamma},
\end{equation}
where $V_{\max} := \sup_{s,\pi}  |V^\pi(s)|$. 
\end{theorem}

\textbf{Remarks.}
(1) The certificate holds for any attack algorithm $\nu$ with $C<\frac{N-1}{2}$. 
(2) If $\epsilon_R$ and $\epsilon_P$ are small, then the performance of \ensname $\smpolicy$ under attacks is similar to the performance of the \ablname $\ablpolicy$ under no attack. 

It is important to note that $\epsilon_R$ and $\epsilon_P$ are intrinsic properties of $\ablpolicy$, independent of the attacker. Therefore, one can approximately measure $\epsilon_R$ and $\epsilon_P$ during training. 
Similar to Condition~\ref{cond:cons} required for a discrete action space, the gap between the attacked reward of $\smpolicy$ and the natural reward of $\ablpolicy$ depends on how well the benign messages are reaching a consensus. (Smaller $\epsilon_R$ and $\epsilon_P$ imply that the actions in $\safeaction$ are relatively concentrated and the environment dynamics are relatively smooth.)

Moreover, one can optimize $\ablpolicy$ during training such that $\epsilon_R$ and $\epsilon_P$ are as small as possible, to further improve the robustness guarantee of $\smpolicy$. This can be a future extension of this work.

Technical proofs of all theoretical results can be found in Appendix~\ref{app:proof}.

\subsection{Extension of \ours with Partial Samples}
\label{app:theory_sample}

As motivated in Section~\ref{sec:algo_more}, our \ours can be extended to a partial-sample version, where the ensemble policy is constructed by $D$ instead of $\binom{N-1}{k}$ samples. Let $\allmsgkd$ be a subset of $\allmsgk$ that contains $D$ random \ksamples from $\allmsgk$. Then the $D$-ensemble policy $\pi_D$ is defined as
\begin{equation}
\label{eq:ensemble_pi_dist_d}
    \smpolicy_D(\his, \mathbf{m}) := \mathrm{argmax}_{a\in\mathcal{A}} \sum_{\msgk\in\allmsgkd} \mathbbm{1}[\ablpolicy(o, \msgk)=a],
\end{equation}
for a discrete action space, and
\begin{equation}
\label{eq:ensemble_pi_cont_d}
    \tilde{\pi}_D(\his, \mathbf{m}) = \funmed \{\ablpolicy(\his, \msgk) \}_{\msgk\in\allmsgkd}.
\end{equation}\normalsize
for a continuous action space. 

In the partial-sample version of \ours, we can still provide high-probability robustness guarantees.

For notation simplicity, let $n_1=\binom{N-1}{k}$, $n_2=\binom{N-C-1}{k}$. 
Define the majority vote as 
\begin{equation}
\label{eq:vote}
    u_{\max}:=\max_{a\in\mathcal{A}} \sum\nolimits_{\msgk\in\allmsgkd} \mathbbm{1}[\ablpolicy(\his, \msgk)=a].
\end{equation}
The following theorem shows a general guarantee for $D$-ensemble policy in a discrete action space. 

\begin{theorem}[General Action Guarantee for Discrete Action Space]
\label{thm:sample_dist}
Given an arbitrary sample size $0<D\leq\binom{N-1}{k}$, for the $D$-ensemble policy $\smpolicy_D$ defined in Equation~\eqref{eq:ensemble_pi_dist_d},
Relation~\eqref{eq:cert_dist} holds deterministically if the majority vote $u_{\max}>n_1-n_2$. 
Otherwise it holds with probability at least
\begin{equation}
\label{eq:p_sample_dist}
    p_D=\frac{\sum_{j=0}^{u_{\max}-1}\binom{n_1-n_2}{j}\binom{n_2}{D-j}}{\binom{n_1}{D}}.
\end{equation}
\end{theorem}
Note that Theorem~\ref{thm:main_dist} is a special case of Theorem~\ref{thm:sample_dist}, since it assumes $u_{\max}>n_1-n_2$.

Theorem~\ref{thm:sample_cont} below further shows the theoretical result for a continuous action space.

\begin{theorem}[General Action Guarantee for Continuous Action Space]
\label{thm:sample_cont}
Given an arbitrary sample size $0<D\leq\binom{N-1}{k}$, for the $D$-ensemble policy $\smpolicy_D$ defined in Equation~\eqref{eq:ensemble_pi_cont_d} with an ablation size $k$ satisfying Condition~\ref{cond:k}, Relation~\eqref{eq:cert_cont} holds with probability at least
\begin{equation}
\label{eq:p_sample_cont}
    p_D=\frac{\sum_{j=\tilde{D}}^{D}\binom{n_2}{j}\binom{n_1-n_2}{D-j}}{\binom{n_1}{D}},
\end{equation}
where $\tilde{D} = {\lfloor \frac{D}{2} \rfloor}+1$. 
\end{theorem}

The larger $D$ is, the higher the probability $p_D$ is, the more likely that the \ensname selects an action in $\safeactioncont$.
In Theorem~\ref{thm:sample_cont}, when $D=\binom{N-1}{k}$,  the probability $p_D$ is 1 and the result matches Theorem~\ref{thm:main_cont}.

Technical proofs of all theoretical results can be found in Appendix~\ref{app:proof}.

%% file: neurips/a3_proof.tex
\section{Technical Proofs}
\label{app:proof}

For the simplicity of the proof, we make the following definition.

\begin{definition} (Purely Benign \ksample and contaminated \ksample)
A \ksample $\msgk\in\allmsgk$ is purely benign if every message in $\msgk$ comes from a benign agent and is unperturbed. 
A \ksample $\msgk\in\allmsgk$ is contaminated if there exists some message in $\msgk$ that is perturbed.
\end{definition}

For notation simplicity, let $n_1:=|\allmsgk|=\binom{N-1}{k}$ be the total number of \ksamples from a message set $\msg$.
Note that the total number of purely benign \ksamples is $n_2:=\binom{N-C-1}{k}$, and the total number of contaminated \ksamples is $n_1-n_2=\binom{N-1}{k}-\binom{N-C-1}{k}$.

\subsection{Proofs in Discrete Action Space}

\textbf{Action Certificates}\quad
We first prove the action certificates in the discrete action. 
Note that Theorem~\ref{thm:main_dist} is a special case of Theorem~\ref{thm:sample_dist} ($u_{\max}>n_1-n_2$ and $D=\binom{N-1}{k}$), so we first prove the general version Theorem~\ref{thm:sample_dist} and then Theorem~\ref{thm:main_dist} holds as a result.

\begin{proof}[Proof of Theorem~\ref{thm:sample_dist} and Theorem~\ref{thm:main_dist}]


The majority voted action $\tilde{a}$ is a benign action, i.e., $\tilde{a}\in \safeaction$, if the ablation policy $\hat{\pi}$ renders action $\tilde{a}$ for
at least one purely benign \ksample.  
If $u_{\max}>n_1-n_2$, since $n_1-n_2$ is exactly the total number of contaminated \ksamples, then it is guaranteed that there is at least one purely benign \ksample for which $\hat{\pi}$ renders $\tilde{a}$. Thus, $\tilde{a}\in \safeaction$, and Theorem~\ref{thm:main_dist} holds.

On the other hand, if $u_{\max}\leq n_1-n_2$, then in order for $\tilde{a}$ to be in $\safeaction$, among the $u_{\max}$ \ksamples resulting in $\tilde{a}$ there can be at most $u_{\max}-1$ contaminated \ksamples.  
There are $ \sum_{j=0}^{u_{\max}-1}\binom{n_1-n_2}{j}\binom{n_2}{D-j}$ such combinations in terms of the sampling of $D$, and the total number of combinations are $\binom{n_1}{D}$. Therefore, we get Equation \eqref{eq:p_sample_dist}.

\end{proof}




\textbf{Reward Certificate.}\quad
Next, following Theorem~\ref{thm:main_dist}, we proceed to prove the reward certificate.

\begin{proof}[Proof of Corollary~\ref{cor:reward_dist}]

Based on the definition of benign action set $\safeaction$, $\smpolicy$ selects an action $\widetilde{a}$ if and only if there exists a purely benign \ksample $\msgk\in\allmsgk$ such that the \ablname $\ablpolicy$ selects $\widetilde{a}=\ablpolicy(\tau, \msgk)$. 
Therefore, for any trajectory generated by $\smpolicy$ under attacks, there is a trajectory of $\ablpolicy$ with a list of \ksamples $\msgk^{(1)},\msgk^{(2)},\cdots$ that renders the same cumulative reward under no attack.

\end{proof}

\subsection{Proofs in Continuous Action Space}

\textbf{Action Certificate.}\quad
Similar to the discrete-action case, we first prove Theorem~\ref{thm:sample_cont}, and then prove Theorem~\ref{thm:main_cont} as a special case of Theorem~\ref{thm:sample_cont}.

\begin{proof}[Proof of Theorem~\ref{thm:sample_cont}]


To understand the intuition of element-wise median operation in continuous action space, let us first start with an intuitive example: 
consider 5 arbitrary numbers $x_1$, ..., $x_5$, if we already know 3 of them $x_1$, $x_2$, $x_3$, then it is certain that 
$\min(x_1, x_2, x_3) \leq \funmed(x_1, \cdots, x_5) \leq \max(x_1, x_2, x_3)$.
Therefore, when purely benign \ksamples form the majority (Condition~\ref{cond:k}), the element-wise median action falls into the range of actions produced by safe messages. 

To be more general, in a continuous action space,  $\tilde{a}\in \safeactioncont$ is equivalent to the condition that out of the $D$ sampled \ksamples,  purely benign \ksamples make up the majority. There are $\sum_{j=\tilde{D}}^{D}\binom{n_2}{j}\binom{n1-n2}{D-j}$ such combinations in terms of the sampling of $D$, where $\tilde{D}={\lfloor \frac{D}{2} \rfloor}+1$. Once again the total number of combinations is $\binom{n_1}{D}$. Therefore, we get Equation \eqref{eq:p_sample_cont}.
\end{proof}

\begin{proof}[Proof of Theorem~\ref{thm:main_cont}]
The proof of Theorem~\ref{thm:main_cont} follows as a special case of Theorem~\ref{thm:sample_cont} when $D=\binom{N-1}{k}=n_1$. In this case, the only non-zero term left in the numerator of $p_D$ is $\binom{n_2}{j}\binom{n_1-n_2}{n_1-j}=\binom{n_2}{n_2}\binom{n_1-n_2}{n_1-n_2}=1$ (we need $n_2\geq j$ and $n_1-n_2 \geq n_1-j$ to keep the numerator from vanishing, which implies $j=n_2$, which is no lower than $\tilde{D}$ since $n_2>n_1/2$ due to Condition~\ref{cond:k}). Hence we have $p_D=1$. 
\end{proof}

\textbf{Reward Certificate.}\quad
Next, we derive the reward guarantee for the continuous-action case.

\begin{proof}[Proof of Theorem~\ref{thm:reward_cont}]


We let $\mathbb{P}(a|s;\pi)$ be the probability of the \ablname $\ablpolicy$ taking action $a$ at state $s$, where $\pi$ can be either the \ablname $\ablpolicy$ or the \ensname $\smpolicy$. Note that this is a conditional probability function, and the policy does not necessarily observe $s$.



Without loss of generality, let $\nu^*$ be the optimal attacking algorithm that minimizes $\tilde{V}^{\smpolicy}_\nu$. 
Let $\mathcal{A}_s$ denote the range of \safename at state $s$ induced by the current \ablname $\ablpolicy$.
Then we have

\begin{equation}
    \begin{aligned}
    & \sup_{s\in\mathcal{S}} \left| V^{\ablpolicy}(s) - \tilde{V}^{\smpolicy}_{\nu^*}(s) \right| \\
    &= \sup_{s\in\mathcal{S}} \left| \mathbb{E}_{a\sim\mathbb{P}(a|s;\ablpolicy)} \left[R(s,a) + \gamma \int P(s^\prime|s,a) V^{\ablpolicy}(s^\prime) \mathrm{d}s^\prime \right] - \mathbb{E}_{a\sim\mathbb{P}(a|s;\smpolicy)} \left[R(s,a) + \gamma \int P(s^\prime|s,a) \tilde{V}^{\smpolicy}_{\nu^*}(s^\prime) \mathrm{d}s^\prime \right]   \right| \\
    &\leq \sup_{s\in\mathcal{S}} \sup_{a_1,a_2 \in \mathcal{A}_s} \left| R(s,a_1) + \gamma \int P(s^\prime|s,a_1) V^{\ablpolicy}(s^\prime) \mathrm{d}s^\prime - R(s,a_2) - \gamma \int P(s^\prime|s,a_2) \tilde{V}^{\smpolicy}_{\nu^*}(s)(s^\prime) \mathrm{d}s^\prime \right| \\
    &\leq \sup_{s\in\mathcal{S}} \sup_{a_1,a_2 \in \mathcal{A}_s} \left| R(s,a_1) - R(s,a_2) \right| + \sup_{a_1,a_2 \in \mathcal{A}_s} \left| \gamma \int P(s^\prime|s,a_1) V^{\ablpolicy}(s^\prime) \mathrm{d}s^\prime - \gamma \int P(s^\prime|s,a_2) \tilde{V}^{\smpolicy}_{\nu^*}(s^\prime) \mathrm{d}s^\prime \right| \\
    &\leq \epsilon_R + \gamma \sup_{s\in\mathcal{S}} \sup_{a_1,a_2 \in \mathcal{A}_s} \left| \int P(s^\prime|s,a_1) V^{\ablpolicy}(s^\prime) \mathrm{d}s^\prime - \int P(s^\prime|s,a_2) \tilde{V}^{\smpolicy}_{\nu^*}(s^\prime) \mathrm{d}s^\prime \right| \\
    &\leq \epsilon_R + \gamma \sup_{s\in\mathcal{S}} \sup_{a_1,a_2 \in \mathcal{A}_s} \left| \int P(s^\prime|s,a_1) V^{\ablpolicy}(s^\prime) \mathrm{d}s^\prime - \int P(s^\prime|s,a_1) \tilde{V}^{\smpolicy}_{\nu^*}(s^\prime) \mathrm{d}s^\prime \right| \\
    & \hspace{3em} + \gamma \sup_{s\in\mathcal{S}} \sup_{a_1,a_2 \in \mathcal{A}_s}\left|\int P(s^\prime|s,a_1) \tilde{V}^{\smpolicy}_{\nu^*}(s^\prime) \mathrm{d}s^\prime - \int P(s^\prime|s,a_2) \tilde{V}^{\smpolicy}_{\nu^*}(s^\prime) \mathrm{d}s^\prime \right| \\
    &\leq \epsilon_R + \gamma \sup_{s\in\mathcal{S}}\left| V^{\ablpolicy}(s) - \tilde{V}^{\smpolicy}_{\nu^*}(s) \right| + \gamma \left|\int P(s^\prime|s,a_1) \tilde{V}^{\smpolicy}_{\nu^*}(s^\prime) \mathrm{d}s^\prime - \int P(s^\prime|s,a_2) \tilde{V}^{\smpolicy}_{\nu^*}(s^\prime) \mathrm{d}s^\prime \right| \\
    &\leq \epsilon_R + \gamma \sup_{s\in\mathcal{S}}\left| V^{\ablpolicy}(s) - \tilde{V}^{\smpolicy}_{\nu^*}(s) \right| + \gamma V_{\max} \epsilon_P.
\end{aligned}
\end{equation}

By solving for the recurrence relation over $\sup_{s\in\mathcal{S}}\left| V^{\ablpolicy}(s) - \tilde{V}^{\smpolicy}_{\nu^*}(s) \right|$, we obtain
\begin{equation}
    \sup_{s\in\mathcal{S}}\left| V^{\ablpolicy}(s) - \tilde{V}^{\smpolicy}_{\nu^*}(s) \right| \leq \frac{\epsilon_R+\gamma V_{\max} \epsilon_P}{1-\gamma}.
\end{equation}
which leads to the desired relation in Theorem~\ref{thm:reward_cont}.



\end{proof}

%% file: neurips/a4_exp.tex
\section{Experiment Details and Additional Results in FoodCollector and \marketenv}
\label{app:exp}

\subsection{More Environment Description}
\label{app:exp_env}

\subsubsection{FoodCollector}
\label{app:exp_env:food}

\begin{figure}[htbp]
\vspace{-1em}
    \centering
    \includegraphics[width=0.4\columnwidth]{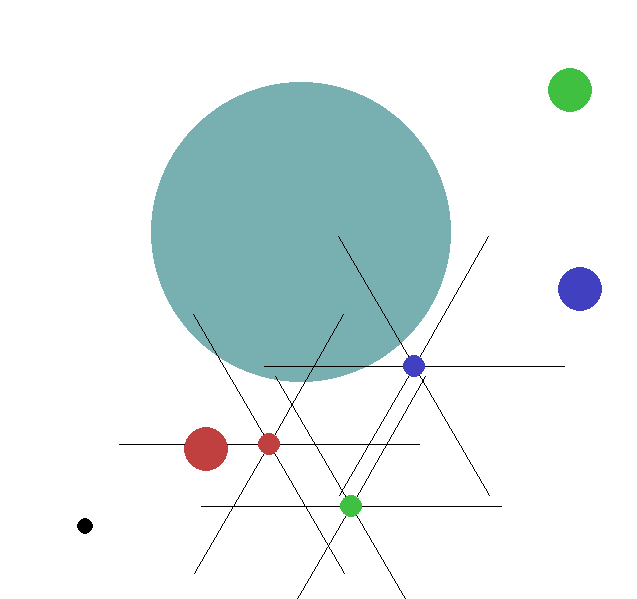}
    \vspace{-0.5em}
    \caption{The FoodCollector Environment. For figure readability, we only show 3 agents colored as red/green/blue. In our experiments, there are 9 agents. }
    \label{fig:food}
\vspace{-1em}
\end{figure}

The FoodCollector environment is a 2D particle world shown by Figure~\ref{fig:food}(left). There are $N=9$ agents with different colors, and $N$ foods with colors corresponding to the $N$ agents. Agents are rewarded when eating foods with the same color. A big round obstacle is located in the center of the map, which the agent cannot go through. There are some poisons (shown as black dots) in the environment, and the agents get penalized whenever they touch the poison. Each agent has 6 sensors that detect the objects around it, including the poisons and the colored foods. 
The game is episodic, with horizon set to be 200. In the beginning of each episode, the agents, foods and poisons are randomly generated in the world. 

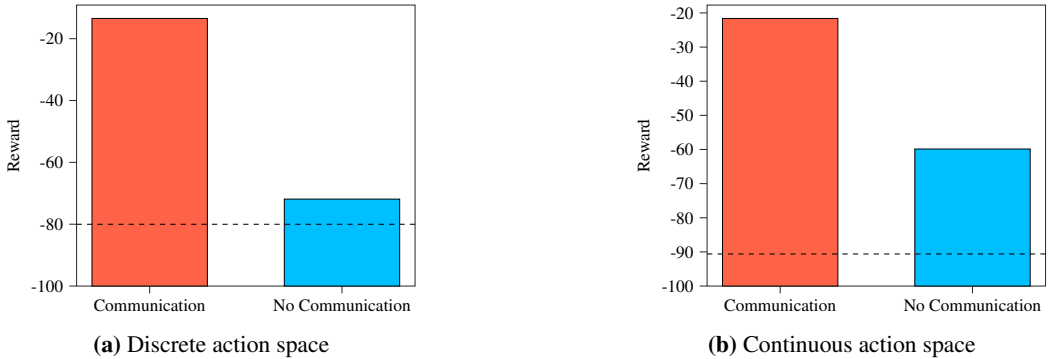
\begin{figure}[htbp]
    \centering
    \begin{subfigure}[t]{0.4\textwidth}
        \centering
        \resizebox{\linewidth}{!}{\input{figs/comm_dist}}
        \caption{Discrete action space}
    \end{subfigure}
    \hfill
    \begin{subfigure}[t]{0.4\textwidth}
        \centering
        \resizebox{\linewidth}{!}{\input{figs/comm_cont}}
        \caption{Continuous action space}
    \end{subfigure}
\caption{\textbf{FoodCollector}: Reward of agents trained by PPO with communication v.s. without communication. Black dashed line stands for the mean performance of the agent when selecting actions uniformly randomly.}
\label{fig:food_comm}
\end{figure}

\textbf{State Observation}\quad
Each of the 6 sensors can detect the following values when the corresponding element is within the sensor's range (the corresponding dimensions are 0's if nothing is detected):\\
(1) (if detects a food) the distance to a food (real-valued); \\
(2) (if detects a food) the color of the food (one-hot); \\
(3) (if detects an obstacle) the distance to the obstacle (real-valued); \\
(4) (if detects the boundary) the distance to the boundary (real-valued); \\ (5) (if detects another agent) the distance to another agent (real-valued).\\
The observation of an agent includes an agent-identifier (one-hot encoding), its own location (2D coordinates), its own velocity, two flags of colliding with its food and colliding with poison, and the above sensory inputs. Therefore, the observation space is a ($7N+30$)-dimensional vector space. 

\textbf{Agent Action}\quad
The action space can be either \textit{discrete} or \textit{continuous}. For the discrete version, there are 9 actions including 8 moving directions (north, northwest, west, southwest, south, southeast, east, northeast) and 1 no-move action. For the continuous version, the action is an acceleration decision, denoted by a 2-dimensional real-valued vector, with each coordinate taking values in $[-0.01,0.01]$. 

\textbf{Reward Function}\quad
At every step, each agent will receives a negative reward $-0.5$ if it has not eaten all its food. In addition, it receives extra $-1$ reward if it collides with a poison. 
Therefore, every agent is expected to explore the environment and eat all food as fast as possible. The team reward is calculated by the average of all agents' local rewards. Note that the agents' actions do not affect each other, because they have different target foods to collect. \emph{Agents collaborate only via communication introduced below.}

\textbf{Communication Protocol}\quad
Due to the limited sensory range, every agent can only see the objects around it and thus only partially observes the world. Therefore, communication among agents can help them find their foods much faster. Since our focus is to defend against adversarially perturbed communications, we first define a valid and beneficial communication protocol, where an agent sends a message to a receiver once it observes a food with the receiver's color. For example, if a red agent encounters a blue food, it can then send a message to the blue agent so that the blue agent knows where to find its food. 
To remember the up-to-date communication, every agent maintains a list of most recent $N-1$ messages sent from other $N-1$ agents.
A message contains the sender's current location and the relative distance to the food (recorded by the 6 sensors), which are bounded between -1 and 1. Therefore, a message is a $8$-dimensional vector, and each agent's communication list has $8(N-1)$ dimensions in total.

\textbf{Communication Gain}\quad
During training with communication, we concatenate the observation and the communication list of the agent to an MLP-based policy, compared to the non-communicative case where the policy only takes in local observations. More implementation details are in Appendix~\ref{app:exp_imple:food}.
As verified in Figure~\ref{fig:food_comm}, communication does help the agent to obtain a much higher reward in both discrete action and continuous action cases, which suggests that the agents tend to rely heavily on the communication messages for finding their food.


\subsubsection{\marketenv}
\label{app:exp_env:market}

The \textit{\marketenv} environment is an inventory management setup, where $N=10$ cooperative heterogeneous distributors carry inventory for $M=3$ products. A population of $B=300$ buyers request a product from a randomly selected distributor agent according to a demand distribution $\mathbf{p}=[p_1,\dots,p_M]$. We denote the demand realization for distributor $i$ with $\mathbf{d_i}=[d_{i,1},\dots,d_{i,M}]$. Distributor agents manage their inventory by restocking products through interacting with the buyers. The game is episodic with horizon set to $50$. At the beginning of each episode, a realization of the demand distribution $\mathbf{p}$ is randomly generated and the distributors' inventory for each product is randomly initialized from $[0, \frac{B}{N}]$, where $\frac{B}{N}$ is the expected number of buyers per distributor. 
The distributor agents are penalized for mismatch between their inventory and the demand for a product, and they aim to restock enough units of a product at each step to prevent insufficient inventory without accruing a surplus at the end of each step.  

    \textbf{State Observation}\quad
    A distributor agent's observation includes its inventory for each product, and the products that were requested by buyers during the previous step. The observation space is a $2M$-dimensional vector.
    
    \textbf{Agent Action}\quad
    Distributors manage their inventory by restocking new units of each product or discarding part of the leftover inventory at the beginning of each step. Hence, agents take both positive and negative actions denoted by an $M$-dimensional vector, and the action space can be either discrete or continuous. In our experiments, we use continuous actions assuming that products are divisible and distributors can restock and hold fractions of a product unit.
    
    \textbf{Reward Function}\quad
    During each step, agent $i$'s reward is defined as $r_i = - || \max( \mathbf{I}_i+\mathbf{a}_i, 0) - \mathbf{d}_i||_2$, where $\mathbf{I}_i$ denotes the agents initial inventory vector, and $\mathbf{a}_i$ denotes the inventory restock vector from action policy $\pi_i$. Note that the agents' actions do not affect each other. \emph{Agents collaborate only via communication introduced below.}

    \textbf{Communication Protocol}\quad
    Distributors learn the demand distribution and optimize their inventory by interacting with their own customers (i.e., portion of buyers that request a product from that distributor). Distributors would benefit from sharing their observed demands with each other, so that they could estimate the demand distribution more accurately for managing inventory. At the end of each step, a distributor communicates an $M$-dimensional vector reporting its observed demands to all other agents. In the case of adversarial communications, this message may differ from the agents' truly observed demands.
    
\begin{figure}[h]
\vspace{-0.1cm}
 \centering
\resizebox{.4\linewidth}{!}{\input{figs/MarketEnv/MarketEnv_comm_dist}}
\vspace{-0.1cm}
\caption{\textbf{\marketenv}: Reward of agents trained with communication v.s. without communication. 
} 
\label{fig:marketenv_comm}
\end{figure}
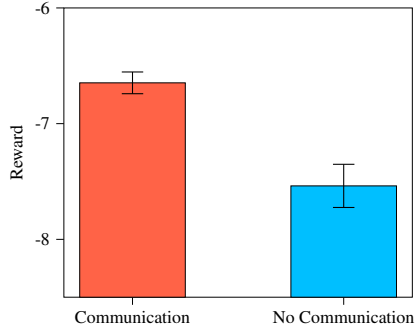

    \textbf{Communication Gain}\quad
    During training with communication, messages received from all other agents are concatenated to the agent's observation, which is used to train an MLP-based policy, compared to the non-communicative case where the policy is trained using only local observations. As observed from Figure \ref{fig:marketenv_comm}, communication helps agents obtain  higher rewards since they are able to manage their inventory based on the overall demands observed across the population of buyers rather than their local observation. Results are reported by averaging rewards corresponding to $5$ experiments run with different training seeds.
 
\subsection{Implementation Details}
\label{app:exp_imple}

\subsubsection{FoodCollector}
\label{app:exp_imple:food}

\textbf{Implementation of Trainer}\quad
In our experiments, we use the Proximal Policy Optimization (PPO) \cite{schulman2017proximal} algorithm to train all agents (with parameter sharing among agents) as well as the  attackers. Specifically, we adapt from the elegant OpenAI Spinning Up \cite{SpinningUp2018} Implementation for PPO training algorithm. On top of the Spinning Up PPO implementation, we also keep track of the running average and standard deviation of the observation and normalize the observation. All experiments are conducted on NVIDIA GeForce RTX 2080 Ti GPUs.

For the policy network, we use a multi-layer perceptron (MLP) with two hidden layers of size 64. For a discrete action space, a categorical output distribution is used. For a continuous action space, since the valid action is bounded within a small range [-0.01,0.01], we parameterize the policy as a Beta distribution, which has been proposed in previous works to better solve reinforcement learning problems with bounded actions  \cite{pmlr-v70-chou17a}. In particular, we parameterize the Beta distribution by $\alpha_\theta$ and $\beta_\theta$, such that $\alpha=\log(1+e^{\alpha_{\theta}(s)})+1$ and $\beta=\log(1+e^{\beta_{\theta}(s)})+1$ (1 is added to make sure that $\alpha, \beta\geq 1$). Then, $\pi(a|s)=f(\frac{a-h}{2h};\alpha, \beta)$, where $h=0.01$, and $f(x;\alpha, \beta)=\frac{\Gamma(\alpha+\beta)}{\Gamma(\alpha)\Gamma(\beta)}x^{\alpha-1}(1-x)^{\beta-1}$ is the density function of the Beta Distribution.
For the value network, we also use an MLP with two hidden layers of size 64. 

In terms of other hyperparameters used in the experiments, we use a learning rate of 0.0003 for the policy network, and a learning rate of 0.001 is used for the value network. We use the Adam optimizer with $\beta_1=0.99$ and  $\beta_2 = 0.999$. For every training epoch, the PPO agent interacts with the environment for 4000 steps, and it is trained for 500 epochs in our experiments.


\textbf{Attacker}\quad
An attacker maps its own observation to the malicious communication messages that it will send to the victim agent. Thus, the action space of the attacker is the communication space of a benign agent, which is bounded between -1 and 1. 
\begin{itemize}
\item \emph{Non-adaptive Attacker}\quad
We implement a fast and naive attacking method for the adversary. At every dimension, the naive attacker randomly picks 1 or -1 as its action, and then sends the perturbed message which consists entirely of 1 or -1 to the victim agent. 
\item \emph{Adaptive RL Attacker}\quad
We use the PPO algorithm to train the attacker, where we set the reward of the attacker to be the negative reward of the victim. The attacker uses a Gaussian policy, where the action is clipped to be in the valid communication range. The network architecture and all other hyperparameter settings follow the exact same from the clean agent training.
\end{itemize}

\textbf{Implementation of Baselines}\quad
\begin{itemize}
\item \emph{Vanilla Learning}\quad
For Vanilla method, we train a shared policy network to map observations and communication to actions. 
\item \emph{Adversarial Training (AT)}\quad
For adversarial training, we alternate between the training of attacker and the training of the victim agent. Both the victim and attackers are trained by PPO. For every 200 training epochs, we switch the training, where we either fix the trained victim and train the attacker for the victim or fix the trained attacker and train the victim under attack. We continue this process for 10 iterations. 
\end{itemize}
Note that the messages are symmetric (of the same format), we shuffle the messages before feeding them into the policy network for both Vanilla and AT, to reduce the bias caused by agent order. We find that shuffling the messages helps the agent converge much faster (50\% fewer total steps). Note that \ours randomly selects \ksamples and thus messages are also shuffled.

\subsubsection{\marketenv}
\label{app:exp_imple:market}

\textbf{Implementation of Trainer}\quad 
As in the FoodCollector experiments, we use the PPO algorithm to train all agent action policy as well as adversarial agent communication policies. We use the same MLP-based policy and value networks as the FoodCollector but parameterize the policy as a Gaussian distribution. The PPO agent interacts with the environment for $50$ steps, and it is trained for $10000$ episodes. The learning rate is set to be 0.0003 for the policy network, and 0.001 for the value network.

\textbf{Attacker}\quad 
The attacker uses its observations to communicate malicious messages to  victim distributors, and its action space is the communication space of a benign agent.  We consider the following non-adaptive and adaptive attackers:
\begin{itemize}
    \item \textit{Non-adaptive Attacker}: The attacker's goal is to harm the victim agent by misreporting its observed demands so that the victim distributor under-estimates or over-estimates the restocking of products. In our experiments, we evaluate the effectiveness of defense strategies against the following attack strategies:
    \begin{itemize}
     \item{\permute}: The communication message is a random permutation of the true demand vector observed by the attacker.
    \item {\swap}: In order to construct a communication vector as different as possible from its observed demand, the attacker reports the most requested products as the least request ones and vice versa. Therefore, the highest demand among the products is interchanged with the lowest demand, the second highest demand is interchanged with the second lowest demand and so forth.
    \item{\flip:} Adversary $i$ modifies its observed demand $\mathbf{d}_i$ by mirroring it with respect to $\eta=\frac{1}{M}\sum_{j=1}^M d_{ij}$, such that products demanded less than $\eta$ are reported as being requested more, and conversely, highly demanded products are reported as less popular.
    \end{itemize}

    \item \textit{Adaptive Attacker}: The attacker communication policy is trained using the PPO algorithm, and its reward is set as the negative reward of the victim agent. The attacker uses a Gaussian policy with a softmax activation in the output layer to learn a adversarial probability distribution across products, which is then scaled by the total observed demands $\sum_{j=1}^M d_{ij}$ to construct the communication message.

\end{itemize}

\textbf{Implementation of Baselines}\quad 
\begin{itemize}
    \item \textit{Vanilla Learning}: In the vanilla training method with no defense mechanism against adversarial  communication, we train a shared policy network using agents' local observations and their received communication messages.
     \item \textit{Adversarial Training (AT)}: We alternate between training the agent action policy and the victim communication policy, both using PPO. For the first iteration we train the policies for $10000$ episodes, and then use $1000$ episodes for $5$ additional training alteration iterations for both the action policy and the adversarial communication policy. For more efficient adversarial training, we first shuffle the received communication messages before feeding them into the policy network. Consequently, the trained policy  treats communications received from different agents in a similar manner, and we are able to only train the policy with a fixed set of adversarial agents rather than training the network on all possible combinations of adversarial agents. 
\end{itemize}
For Vanilla and AT, we do not shuffle the communication messages being input to the policy, as we did not observe improved convergence, as in the FoodCollector environment. 


\subsection{Additional Results}
\label{app:exp_results}

In addition to Figure \ref{fig:res_k_dist}, we also provide the plot for hyper-parameter tests in discrete-action FoodCollector and \marketenv in Figure \ref{fig:res_hyper} below. 

\begin{figure}[h]
 \centering
 \begin{subfigure}[t]{0.24\textwidth}
 \centering
  \resizebox{\columnwidth}{!}{\input{figs/FoodCollector_Ablation_k_2adv_cont}}
  \vspace{-1.5em}
  \caption{FoodCollector}
  \label{sfig:hyper_k_food}
 \end{subfigure}
 \hfill
 \begin{subfigure}[t]{0.24\textwidth}
 \centering
  \resizebox{\columnwidth}{!}{\input{figs/MarketEnv/MarketEnv_Ablation_k_2adv}}
  \vspace{-1.5em}
  \caption{\marketenv}
  \label{sfig:hyper_k_market}
 \end{subfigure}
 \hfill
 \begin{subfigure}[t]{0.24\textwidth}
 \centering
  \resizebox{\columnwidth}{!}{\input{figs/FoodCollector_Ablation_k_sample_cont}}
  \vspace{-1.5em}
  \caption{FoodCollector}
  \label{sfig:hyper_d_food}
 \end{subfigure}
 \hfill
 \begin{subfigure}[t]{0.24\textwidth}
 \centering
  \resizebox{\columnwidth}{!}{\input{figs/MarketEnv/MarketEnv_Ablation_K_sampleD}}
  \vspace{-1.5em}
  \caption{\marketenv}
  \label{sfig:hyper_d_market}
 \end{subfigure}
\vspace{-0.5em}
\caption{Hyper-parameter tests of ablation size $k$ and sample size $D$. 
We show how natural reward and attacked performance change with \textbf{(a)} various $k$ in continuous-action FoodCollector, \textbf{(b)} various $D$ in continuous-action FoodCollector, \textbf{(c)} various $k$ in \marketenv, and \textbf{(d)} various $D$ in \marketenv. \\
Dashed green lines refer to the performance of Vanilla agent under $C=2$ attacks. }
\label{fig:res_hyper}
\end{figure}
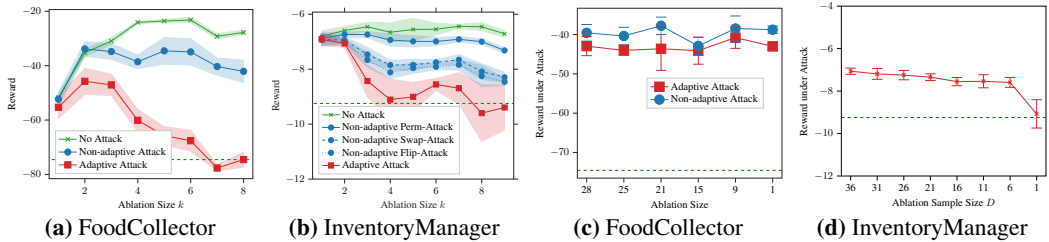

\textbf{Adversarial Training does not improve agent's robust performance.}
As we observed in Figure \ref{fig:res_all}, adversarial training (AT) does not improve the robustness of the agent in both discrete and continuous action space, although AT achieves good robustness againt $\ell_p$ attacks in vision tasks~\cite{madry2018towards,zhang2019theoretically}, and against observation attacks~\cite{zhang2021robust,sun2021strongest} and action attacks~\cite{pinto2017robust} in RL.
We hypothesize that it is due to (1) the large number of total agents,  (2) the uncertainty of adversarial message channels, and (3) the relatively large perturbation length. 
To be more specific, in related works~\cite{pinto2017robust,zhang2021robust,sun2021strongest}, an agent and an attacker are alternately trained, during which the agent learns to adapt to the learned attacker. However, in the threat model we consider, there are $C$ out of $N$ messages significantly perturbed, which it is hard for the agent to adapt to attacks during alternate training.

\textbf{Additional Results with QMIX}\quad 
\ours is a generic defense approach that can be used for any RL/MARL learning algorithm. In Figure~\ref{fig:res_qmix}, we show the results of \ours combined with an MARL algorithm QMIX~\cite{rashid2018qmix} in the discrete-action FoodCollector environment (QMIX does not work for continuous actions so it is not applicable in continuous-action FoodCollector and \marketenv). Compared to the vanilla QMIX algorithm, QMIX+\ours achieves much higher robustness and stable performance under various numbers of adversaries. 
\input{figs/QMIX_AME}

\section{Additional Experiment in MARL Image Classification}
\label{app:exp_mnist}

\subsection{Description of MARL-MNIST Environment}

We use the environment setup proposed by Mousavi et al.~\cite{mousavi2019multi}, where agents collaboratively classify an unknown image by their observations and inter-agent communication. 
More specifically, we use $N=9$ agents in the MNIST dataset of handwritten digits~\cite{lecun1998gradient}.  The dataset consists
of 60,000 training images and 10,000 testing images, where
each image has $28 \times 28$ pixels. 
There are $h=5$ steps in an episode. In the beginning, all agents start from a pre-determined spatial configuration. At every step, each agent observes a local $5 \times 5$ patch, performs some local data processing, and shares the result with neighboring agents (we use a fully connected communication graph). 
With given observation and communication, each agent outputs an action in $\{Up, Down, Left, Right\}$. By each movement, the agent is translated in the desired direction by $5$ pixels. In the end of an episode, agents make predictions, and all of them are rewarded by $- \mathrm{prediction\_loss}$.

\subsection{Implementation Details}

We use the same network architecture and hyperparameter setting as Mousavi et al.~\cite{mousavi2019multi}, which are implemented in~\cite{mnistgithub}. 

\textbf{Network Architecture}\quad
Concretely, at every step, the input of every agent contains 3 components: (1) an encoded observation with two convolutional layers followed by vectorization and a fully connected layer; (2) the average of all communication messages from other agents; (3) a position encoding computed by feeding the current position into a single linear layer.
These 3 components are concatenated and passed to two independent LSTM modules, one is for an acting policy, another is for a message generator. 
In the end of an episode, every agent uses its final cell state to generate a prediction using a 2-layer MLP. Then we take the average of the output logits of all agents, and use a softmax function to obtain the final probabilistic label prediction. The reward is the opposite number of the L2 difference between the prediction and the one-hot encoding of true image label.

\textbf{Hyperparameters}\quad
In our experiments, we follow the default hyperparameter setting in~\cite{mnistgithub}. We use $N=9$ agents. The size of LSTM belief state is 128. The hidden layers have size 160. The message size is set to be 32. The state encoding has size 8. We use an Adam optimizer with learning rate 1e-3. We train the agents in the MNIST dataset for 40 epochs.

\textbf{Attackers}\quad
Since the communication messages are learned by neural networks, we perturb the $C$ messages received by each agent. To make sure that the messages are not obviously detectable, we clip every dimension of the perturbed message into the range of $[-3,3]$. The non-adaptive attacker randomly generates a new message. The adaptive attacker learns a new message generator based on its own belief state, which is trained with learning rate 1e-3 for 50 epochs.

\subsection{Experiment Results}
Figure~\ref{fig:res_all_mnist} demonstrates the robust performance of the MARL algorithm proposed by Mousavi et al.~\cite{mousavi2019multi} with our \ours defense or baseline defenses (Vanilla and AT). We set $N=9$ and $k=2$ for all experiments.
Under learned adaptive attackers, the original MARL classifier (Vanilla)~\cite{mousavi2019multi}  without \ours suffers from significant performance drop in terms of both precision and recall. Defending with adversarial training (AT) does not achieve good robustness, either. But \ours considerably improves the robustness of agents across different numbers of attackers. \\
Under random attacks, we find that the original MARL classifier~\cite{mousavi2019multi} is moderately robust when a few communication signals are randomly perturbed. However, when noise exists in many communication channels (e.g. $C=6$), the performance decreases a lot. In contrast, our \ours still achieves high performance when $C=6$ communication messages are corrupted, even if the guarantee of ablation size $k$ only holds for $C\leq 2$. Therefore, we again emphasize the the theoretical guarantee considers the worst-case strong attack, while under a relatively weak attack, we can achieve better robustness beyond what the theory suggests.

\begin{figure}[!t]
\hspace{-1em}
 \centering
 \begin{subfigure}[t]{0.24\columnwidth}
  \resizebox{\columnwidth}{!}{\input{figs/mnist/MNIST_Adaptive_Attacker_Precision}}
  \vspace{-1.5em}
  \caption{Adaptive Attack}
 \end{subfigure}
 \hfill
 \begin{subfigure}[t]{0.24\textwidth}
 \centering
  \resizebox{\columnwidth}{!}{\input{figs/mnist/MNIST_Adaptive_Attacker_Recall}}
  \vspace{-1.5em}
  \caption{Adaptive Attack}
 \end{subfigure}
 \hfill
 \begin{subfigure}[t]{0.24\columnwidth}
  \resizebox{\columnwidth}{!}{\input{figs/mnist/MNIST_Random_Attacker_Precision}}
  \vspace{-1.5em}
  \caption{Non-adaptive Attack}
 \end{subfigure}
\hfill
 \begin{subfigure}[t]{0.24\textwidth}
 \centering
  \resizebox{\columnwidth}{!}{\input{figs/mnist/MNIST_Random_Attacker_Recall}}
  \vspace{-1.5em}
  \caption{Non-adaptive Attack}
 \end{subfigure}
\vspace{-0.5em}
\caption{\textbf{(MARL-MNIST)}: Precision and recall of MARL classification on MNIST without or with \ours, under learned adaptive attacks and non-adaptive random attacks. All results are averaged over 5 random seeds. }
\label{fig:res_all_mnist}
\end{figure}

%% file: figs/comm_dist.tex
\begin{tikzpicture}[scale=0.8]

\definecolor{color0}{rgb}{1,0.388235294117647,0.27843137254902}
\definecolor{color1}{rgb}{0,0.749019607843137,1}

\begin{axis}[
axis line style={white!15!black},
legend cell align={left},
legend style={fill opacity=1, draw opacity=1, text opacity=1, draw=white!80!black},
tick align=outside,
x grid style={white!80!black},
xmajorticks=true,
xmin=-0.38, xmax=1.38,
xtick style={color=white!15!black},
xtick={0,1},
xtick pos = lower,
xticklabels={Communication,No Communication},
y grid style={white!80!black},
ylabel={Reward},
ymajorticks=true,
ytick pos = left,
ymin=0, ymax=90.8775,
ytick style={color=white!15!black},
ytick={0,20,40,60,80,100},
yticklabels={-100,-80,-60,-40,-20,}
]
\draw[draw=black,fill=color0] (axis cs:-0.3,0) rectangle (axis cs:0.3,86.55);

\draw[draw=black,fill=color1] (axis cs:0.7,0) rectangle (axis cs:1.3,28.12);

\addplot [semithick, black, dashed, forget plot]
table {%
-0.38 19.98
1.38 19.98
};
\end{axis}

\end{tikzpicture}

%% file: figs/comm_cont.tex
\begin{tikzpicture}[scale=0.8]

\definecolor{color0}{rgb}{1,0.388235294117647,0.27843137254902}
\definecolor{color1}{rgb}{0,0.749019607843137,1}

\begin{axis}[
axis line style={white!15!black},
tick align=outside,
x grid style={white!80!black},
xmajorticks=true,
xmin=-0.38, xmax=1.38,
xtick style={color=white!15!black},
xtick={0,1},
xtick pos = lower,
xticklabels={Communication,No Communication},
y grid style={white!80!black},
ylabel={Reward},
ymajorticks=true,
ymin=0, ymax=82.299,
ytick style={color=white!15!black},
ytick={0,10,20,30,40,50,60,70,80,90},
yticklabels={-100,-90,-80,-70,-60,-50,-40,-30,-20,},
ytick pos = left,
]
\draw[draw=black,fill=color0] (axis cs:-0.3,0) rectangle (axis cs:0.3,78.38);

\draw[draw=black,fill=color1] (axis cs:0.7,0) rectangle (axis cs:1.3,40.14);

\addplot [semithick, black, dashed]
table {%
-0.38 9.4
1.38 9.4
};
\end{axis}

\end{tikzpicture}

%% file: figs/MarketEnv/MarketEnv_comm_dist.tex
\begin{tikzpicture}

\definecolor{darkgray176}{RGB}{176,176,176}
\definecolor{deepskyblue}{RGB}{0,191,255}
\definecolor{tomato}{RGB}{255,99,71}

\begin{axis}[
tick align=outside,
tick pos=left,
x grid style={darkgray176},
xmin=-0.325, xmax=1.325,
xtick style={color=black},
xtick={0,1},
xticklabels={Communication,No Communication},
y grid style={darkgray176},
ylabel={Reward},
ymin=0, ymax=2.5,
ytick style={color=black},
ytick={0.5,1.5,2.5},
yticklabels={-8,-7,-6}
]
\draw[draw=black,fill=tomato,line width=0.08pt] (axis cs:-0.25,0) rectangle (axis cs:0.25,1.85261747367);
\draw[draw=black,fill=deepskyblue,line width=0.08pt] (axis cs:0.75,0) rectangle (axis cs:1.25,0.962091208159715);
\path [draw=black, semithick]
(axis cs:0,1.75868214378157)
--(axis cs:0,1.94655280355844);

\addplot [semithick, black, mark=-, mark size=6, mark options={solid}, only marks]
table {%
0 1.75868214378157
};
\addplot [semithick, black, mark=-, mark size=6, mark options={solid}, only marks]
table {%
0 1.94655280355844
};
\path [draw=black, semithick]
(axis cs:1,0.775513325279025)
--(axis cs:1,1.14866909104041);

\addplot [semithick, black, mark=-, mark size=6, mark options={solid}, only marks]
table {%
1 0.775513325279025
};
\addplot [semithick, black, mark=-, mark size=6, mark options={solid}, only marks]
table {%
1 1.14866909104041
};
\end{axis}

\end{tikzpicture}

%% file: figs/FoodCollector_Ablation_k_2adv_cont.tex
\begin{tikzpicture}[scale=0.7]

\definecolor{color0}{rgb}{0.172549019607843,0.627450980392157,0.172549019607843}
\definecolor{color1}{rgb}{0.12156862745098,0.466666666666667,0.705882352941177}
\definecolor{color2}{rgb}{0.83921568627451,0.152941176470588,0.156862745098039}

\begin{axis}[
axis line style={white!15!black},
legend cell align={left},
legend style={
  fill opacity=1,
  draw opacity=1,
  text opacity=1,
  at={(0.03,0.03)},
  anchor=south west,
  draw=white!80!black
},
tick align=outside,
x grid style={white!80!black},
xlabel={Ablation Size \(\displaystyle k\)},
xmajorticks=true,
xmin=0.65, xmax=8.35,
xtick pos = lower,
xtick style={color=white!15!black},
y grid style={white!80!black},
ylabel={Reward},
ymajorticks=true,
ytick pos = left,
ymin=-81.557, ymax=-19.363,
ytick style={color=white!15!black}
]
\path [draw=color0, fill=color0, opacity=0.2]
(axis cs:1,-50.47)
--(axis cs:1,-52.97)
--(axis cs:2,-36.6)
--(axis cs:3,-32.26)
--(axis cs:4,-24.51)
--(axis cs:5,-24.2)
--(axis cs:6,-24.23)
--(axis cs:7,-30.22)
--(axis cs:8,-28.53)
--(axis cs:8,-27.05)
--(axis cs:8,-27.05)
--(axis cs:7,-28.18)
--(axis cs:6,-22.19)
--(axis cs:5,-22.94)
--(axis cs:4,-23.67)
--(axis cs:3,-29.64)
--(axis cs:2,-33.82)
--(axis cs:1,-50.47)
--cycle;

\path [draw=color1, fill=color1, opacity=0.2]
(axis cs:1,-49.7)
--(axis cs:1,-54.96)
--(axis cs:2,-36.49)
--(axis cs:3,-37.7)
--(axis cs:4,-41.09)
--(axis cs:5,-39.42)
--(axis cs:6,-39.81)
--(axis cs:7,-43.51)
--(axis cs:8,-46.22)
--(axis cs:8,-38)
--(axis cs:8,-38)
--(axis cs:7,-37.13)
--(axis cs:6,-30.07)
--(axis cs:5,-29.68)
--(axis cs:4,-36.05)
--(axis cs:3,-31.9)
--(axis cs:2,-31.21)
--(axis cs:1,-49.7)
--cycle;

\path [draw=color2, fill=color2, opacity=0.2]
(axis cs:1,-51.18)
--(axis cs:1,-59.4)
--(axis cs:2,-50.56)
--(axis cs:3,-50.98)
--(axis cs:4,-64.11)
--(axis cs:5,-69.01)
--(axis cs:6,-71.49)
--(axis cs:7,-78.73)
--(axis cs:8,-77.28)
--(axis cs:8,-71.82)
--(axis cs:8,-71.82)
--(axis cs:7,-76.55)
--(axis cs:6,-63.71)
--(axis cs:5,-62.47)
--(axis cs:4,-56.09)
--(axis cs:3,-43.18)
--(axis cs:2,-40.94)
--(axis cs:1,-51.18)
--cycle;

\addplot [semithick, color0, mark=x, mark size=3, mark options={solid}]
table {%
1 -51.72
2 -35.21
3 -30.95
4 -24.09
5 -23.57
6 -23.21
7 -29.2
8 -27.79
};
\addlegendentry{No Attack}
\addplot [semithick, color1, mark=*, mark size=3, mark options={solid}]
table {%
1 -52.33
2 -33.85
3 -34.8
4 -38.57
5 -34.55
6 -34.94
7 -40.32
8 -42.11
};
\addlegendentry{Non-adaptive Attack}
\addplot [semithick, color2, mark=square*, mark size=3, mark options={solid}]
table {%
1 -55.29
2 -45.75
3 -47.08
4 -60.1
5 -65.74
6 -67.6
7 -77.64
8 -74.55
};
\addlegendentry{Adaptive Attack}
\addplot [semithick, green!50!black, dashed]
table {%
0.75 -74.55
8.25 -74.55
};
\end{axis}

\end{tikzpicture}

%% file: figs/MarketEnv/MarketEnv_Ablation_k_2adv.tex
\begin{tikzpicture}

\definecolor{crimson2143940}{RGB}{214,39,40}
\definecolor{darkslategray38}{RGB}{38,38,38}
\definecolor{forestgreen4416044}{RGB}{44,160,44}
\definecolor{lightgray204}{RGB}{204,204,204}
\definecolor{steelblue31119180}{RGB}{31,119,180}

\begin{axis}[
axis line style={darkslategray38},
legend cell align={left},
legend style={
  fill opacity=1,
  draw opacity=1,
  text opacity=1,
  at={(0.03,0.03)},
  anchor=south west,
  draw=lightgray204
},
tick align=outside,
x grid style={lightgray204},
xlabel=\textcolor{darkslategray38}{Ablation Size \(\displaystyle k\)},
xmin=0.6, xmax=9.4,
xtick style={color=darkslategray38},
y grid style={lightgray204},
ylabel=\textcolor{darkslategray38}{Reward},
ymin=-12, ymax=-5.921281323553,
ytick style={color=darkslategray38}
]
\path [draw=forestgreen4416044, fill=forestgreen4416044, opacity=0.2]
(axis cs:1,-6.59783973870044)
--(axis cs:1,-7.0071927980605)
--(axis cs:2,-6.72688970455533)
--(axis cs:3,-6.62029947542658)
--(axis cs:4,-7.00323512405983)
--(axis cs:5,-6.95927351837719)
--(axis cs:6,-6.75759808545436)
--(axis cs:7,-6.51077553592448)
--(axis cs:8,-6.59921805067195)
--(axis cs:9,-6.83470780756119)
--(axis cs:9,-6.58696161670332)
--(axis cs:9,-6.58696161670332)
--(axis cs:8,-6.30947714856544)
--(axis cs:7,-6.36701242768733)
--(axis cs:6,-6.33262393916374)
--(axis cs:5,-6.14556300071715)
--(axis cs:4,-6.30969806788508)
--(axis cs:3,-6.29247015239239)
--(axis cs:2,-6.45339077905592)
--(axis cs:1,-6.59783973870044)
--cycle;

\path [draw=steelblue31119180, fill=steelblue31119180, opacity=0.2]
(axis cs:1,-6.68244055764541)
--(axis cs:1,-7.04729572572572)
--(axis cs:2,-6.83921086991393)
--(axis cs:3,-6.85366450766974)
--(axis cs:4,-7.15958292001317)
--(axis cs:5,-7.19288617713524)
--(axis cs:6,-7.1444769220135)
--(axis cs:7,-7.01043149906424)
--(axis cs:8,-7.10109415335005)
--(axis cs:9,-7.42123445029993)
--(axis cs:9,-7.21075853129977)
--(axis cs:9,-7.21075853129977)
--(axis cs:8,-6.89257832235377)
--(axis cs:7,-6.81333860008864)
--(axis cs:6,-6.83421944147124)
--(axis cs:5,-6.78054265825743)
--(axis cs:4,-6.71841804122908)
--(axis cs:3,-6.62321979247172)
--(axis cs:2,-6.63347963462225)
--(axis cs:1,-6.68244055764541)
--cycle;

\path [draw=steelblue31119180, fill=steelblue31119180, opacity=0.2]
(axis cs:1,-6.69727771867657)
--(axis cs:1,-7.13056044943666)
--(axis cs:2,-7.05648964808973)
--(axis cs:3,-7.73037223244322)
--(axis cs:4,-8.02165232443479)
--(axis cs:5,-8.04131848343492)
--(axis cs:6,-7.85526057275484)
--(axis cs:7,-7.83877393733835)
--(axis cs:8,-8.37576055620526)
--(axis cs:9,-8.47557689487488)
--(axis cs:9,-8.10887258725201)
--(axis cs:9,-8.10887258725201)
--(axis cs:8,-7.79890061391681)
--(axis cs:7,-7.47244168383737)
--(axis cs:6,-7.64864292530718)
--(axis cs:5,-7.64170943997633)
--(axis cs:4,-7.67422917429)
--(axis cs:3,-7.19963655913169)
--(axis cs:2,-6.82530273196134)
--(axis cs:1,-6.69727771867657)
--cycle;

\path [draw=steelblue31119180, fill=steelblue31119180, opacity=0.2]
(axis cs:1,-6.71948610091803)
--(axis cs:1,-7.15154788987751)
--(axis cs:2,-7.11739579643354)
--(axis cs:3,-7.92638387371537)
--(axis cs:4,-8.34355168691175)
--(axis cs:5,-8.14892552832822)
--(axis cs:6,-8.01931797959091)
--(axis cs:7,-8.03371079156682)
--(axis cs:8,-8.63524553988125)
--(axis cs:9,-8.59234024006521)
--(axis cs:9,-8.32385836235105)
--(axis cs:9,-8.32385836235105)
--(axis cs:8,-7.86265877164573)
--(axis cs:7,-7.64387468039364)
--(axis cs:6,-7.80747131987031)
--(axis cs:5,-7.76969068245778)
--(axis cs:4,-7.8855803217457)
--(axis cs:3,-7.39513674204925)
--(axis cs:2,-6.90513413806718)
--(axis cs:1,-6.71948610091803)
--cycle;

\path [draw=crimson2143940, fill=crimson2143940, opacity=0.2]
(axis cs:1,-6.69536332469966)
--(axis cs:1,-7.12915128823936)
--(axis cs:2,-7.21036745629226)
--(axis cs:3,-9.00910061524039)
--(axis cs:4,-9.85397516130134)
--(axis cs:5,-9.52630721171358)
--(axis cs:6,-8.77569342090652)
--(axis cs:7,-9.25861483356248)
--(axis cs:8,-10.631196544)
--(axis cs:9,-10.2128533002522)
--(axis cs:9,-8.57537223560712)
--(axis cs:9,-8.57537223560712)
--(axis cs:8,-8.56736808001069)
--(axis cs:7,-8.128674224737)
--(axis cs:6,-8.33445629071456)
--(axis cs:5,-8.48344975005529)
--(axis cs:4,-8.36241639275077)
--(axis cs:3,-7.85347219651206)
--(axis cs:2,-6.91766990559479)
--(axis cs:1,-6.69536332469966)
--cycle;

\addplot [semithick, forestgreen4416044, mark=x, mark size=2.5, mark options={solid}]
table {%
1 -6.80251626838047
2 -6.59014024180563
3 -6.45638481390948
4 -6.65646659597246
5 -6.55241825954717
6 -6.54511101230905
7 -6.43889398180591
8 -6.45434759961869
9 -6.71083471213225
};
\addlegendentry{No Attack}
\addplot [semithick, steelblue31119180, mark=*, mark size=2.5, mark options={solid}]
table {%
1 -6.86486814168557
2 -6.73634525226809
3 -6.73844215007073
4 -6.93900048062112
5 -6.98671441769633
6 -6.98934818174237
7 -6.91188504957644
8 -6.99683623785191
9 -7.31599649079985
};
\addlegendentry{Non-adaptive Perm-Attack}
\addplot [semithick, steelblue31119180, dashed, mark=*, mark size=2.5, mark options={solid}]
table {%
1 -6.91391908405661
2 -6.94089619002554
3 -7.46500439578746
4 -7.8479407493624
5 -7.84151396170562
6 -7.75195174903101
7 -7.65560781058786
8 -8.08733058506104
9 -8.29222474106344
};
\addlegendentry{Non-adaptive Swap-Attack}
\addplot [semithick, steelblue31119180, dotted, mark=*, mark size=2.5, mark options={solid}]
table {%
1 -6.93551699539777
2 -7.01126496725036
3 -7.66076030788231
4 -8.11456600432873
5 -7.959308105393
6 -7.91339464973061
7 -7.83879273598023
8 -8.24895215576349
9 -8.45809930120813
};
\addlegendentry{Non-adaptive Flip-Attack}
\addplot [semithick, crimson2143940, mark=square*, mark size=2.5, mark options={solid}]
table {%
1 -6.91225730646951
2 -7.06401868094352
3 -8.43128640587623
4 -9.10819577702605
5 -9.00487848088443
6 -8.55507485581054
7 -8.69364452914974
8 -9.59928231200537
9 -9.39411276792965
};
\addlegendentry{Adaptive Attack}
\addplot [semithick, green!50!black, dashed, forget plot]
table {%
0.65 -9.24562131226271
9.5 -9.24562131226271
};

\end{axis}

\end{tikzpicture}

%% file: figs/FoodCollector_Ablation_k_sample_cont.tex
\begin{tikzpicture}

\definecolor{color0}{rgb}{0.83921568627451,0.152941176470588,0.156862745098039}
\definecolor{color1}{rgb}{255,127,14}
\definecolor{steelblue31119180}{RGB}{31,119,180}

\begin{axis}[
axis line style={white!15!black},
legend cell align={left},
legend style={
  fill opacity=1,
  draw opacity=1,
  text opacity=1,
  at={(0.91,0.5)},
  anchor=east,
  draw=white!80!black
},
tick align=outside,
x grid style={white!80!black},
xlabel={Ablation Size},
xmin=0.75, xmax=6.25,
xtick pos=left,
xtick style={color=white!15!black},
xtick={6,5,4,3,2,1},
xticklabels={1,9,15,21,25,28},
y grid style={white!80!black},
ylabel={Reward under Attack},
ytick pos = left,
ymin=-76.514, ymax=-33.306,
ytick style={color=white!15!black}
]
\path [draw=color0, semithick]
(axis cs:6,-43.81)
--(axis cs:6,-42.21);

\path [draw=color0, semithick]
(axis cs:5,-43.51)
--(axis cs:5,-38.05);

\path [draw=color0, semithick]
(axis cs:4,-47.55)
--(axis cs:4,-40.65);

\path [draw=color0, semithick]
(axis cs:3,-49.12)
--(axis cs:3,-38.12);

\path [draw=color0, semithick]
(axis cs:2,-45.24)
--(axis cs:2,-42.82);

\path [draw=color0, semithick]
(axis cs:1,-45.37)
--(axis cs:1,-40.53);

\path [draw=color1, semithick]
(axis cs:6,-39.7)
--(axis cs:6,-37.82);

\path [draw=color1, semithick]
(axis cs:5,-41.65)
--(axis cs:5,-35.27);

\path [draw=color1, semithick]
(axis cs:4,-45.1)
--(axis cs:4,-40.72);

\path [draw=color1, semithick]
(axis cs:3,-39.97)
--(axis cs:3,-35.57);

\path [draw=color1, semithick]
(axis cs:2,-42.57)
--(axis cs:2,-38.15);

\path [draw=color1, semithick]
(axis cs:1,-41.68)
--(axis cs:1,-37.44);

\addplot [semithick, color0, mark=square*, mark size=4.5, mark options={solid}]
table {%
6 -43.01
5 -40.78
4 -44.1
3 -43.62
2 -44.03
1 -42.95
};
\addlegendentry{Adaptive Attack}
\addplot [semithick, steelblue31119180, mark=*, mark size=4.5, mark options={solid}]
table {%
6 -38.76
5 -38.46
4 -42.91
3 -37.77
2 -40.36
1 -39.56
};
\addlegendentry{Non-adaptive Attack}

\addplot [semithick, color0, mark=-, mark size=5, mark options={solid}, only marks]
table {%
6 -43.81
5 -43.51
4 -47.55
3 -49.12
2 -45.24
1 -45.37
};
\addplot [semithick, color0, mark=-, mark size=5, mark options={solid}, only marks]
table {%
6 -42.21
5 -38.05
4 -40.65
3 -38.12
2 -42.82
1 -40.53
};
\addplot [semithick, steelblue31119180, mark=-, mark size=5, mark options={solid}, only marks]
table {%
6 -39.7
5 -41.65
4 -45.1
3 -39.97
2 -42.57
1 -41.68
};
\addplot [semithick, steelblue31119180, mark=-, mark size=5, mark options={solid}, only marks]
table {%
6 -37.82
5 -35.27
4 -40.72
3 -35.57
2 -38.15
1 -37.44
};
\addplot [semithick, green!50!black, dashed, forget plot]
table {%
0.75 -74.55
6.25 -74.55
};

\end{axis}

\end{tikzpicture}

%% file: figs/MarketEnv/MarketEnv_Ablation_K_sampleD.tex
\begin{tikzpicture}

\definecolor{crimson2143940}{RGB}{214,39,40}
\definecolor{darkgray176}{RGB}{176,176,176}
\definecolor{green01270}{RGB}{0,127,0}

\begin{axis}[
tick align=outside,
tick pos=left,
x grid style={darkgray176},
xlabel={Ablation Sample Size \(\displaystyle D\)},
xmin=0.65, xmax=8.35,
xtick style={color=black},
xtick={8,7,6,5,4,3,2,1},
xticklabels={1,6,11,16,21,26,31,36},
y grid style={darkgray176},
ylabel={Reward under Attack},
ymin=-12, ymax=-4,
ytick style={color=black}
]
\path [draw=crimson2143940, semithick]
(axis cs:8,-9.74059914597625)
--(axis cs:8,-8.40480946885644);

\path [draw=crimson2143940, semithick]
(axis cs:7,-7.82767881580901)
--(axis cs:7,-7.36011852029993);

\path [draw=crimson2143940, semithick]
(axis cs:6,-7.84636882641228)
--(axis cs:6,-7.23198591180835);

\path [draw=crimson2143940, semithick]
(axis cs:5,-7.74616133715478)
--(axis cs:5,-7.36050787953725);

\path [draw=crimson2143940, semithick]
(axis cs:4,-7.50347219631257)
--(axis cs:4,-7.1878295609426);

\path [draw=crimson2143940, semithick]
(axis cs:3,-7.46801121498131)
--(axis cs:3,-7.02703671201468);

\path [draw=crimson2143940, semithick]
(axis cs:2,-7.44726709902166)
--(axis cs:2,-6.93645933422067);

\path [draw=crimson2143940, semithick]
(axis cs:1,-7.21036745629226)
--(axis cs:1,-6.91766990559479);

\addplot [semithick, crimson2143940, mark=-, mark size=5, mark options={solid}, only marks]
table {%
8 -9.74059914597625
7 -7.82767881580901
6 -7.84636882641228
5 -7.74616133715478
4 -7.50347219631257
3 -7.46801121498131
2 -7.44726709902166
1 -7.21036745629226
};
\addplot [semithick, crimson2143940, mark=-, mark size=5, mark options={solid}, only marks]
table {%
8 -8.40480946885644
7 -7.36011852029993
6 -7.23198591180835
5 -7.36050787953725
4 -7.1878295609426
3 -7.02703671201468
2 -6.93645933422067
1 -6.91766990559479
};
\addplot [semithick, green01270, dashed]
table {%
0.65 -9.24562131226271
8.35 -9.24562131226271
};
\addplot [semithick, crimson2143940, mark=x, mark size=2.5, mark options={solid}]
table {%
8 -9.07270430741634
7 -7.59389866805447
6 -7.53917736911031
5 -7.55333460834601
4 -7.34565087862759
3 -7.24752396349799
2 -7.19186321662117
1 -7.06401868094352
};
\end{axis}

\end{tikzpicture}

%% file: figs/QMIX_AME.tex
\begin{figure}[!h]
\vspace{-0.5em}
 \centering
 \begin{subfigure}[t]{0.3\columnwidth}
  \resizebox{\textwidth}{!}{\input{figs/QMIX_non_adaptive}}
  \vspace{-1.5em}
  \caption{Discrete \& Non-adaptive}
 \end{subfigure}
 \hfill
 \begin{subfigure}[t]{0.3\columnwidth}
   \resizebox{\textwidth}{!}{\input{figs/QMIX_adaptive}}
  \vspace{-1.5em}
  \caption{Discrete \& Adaptive}
 \end{subfigure}
 \vspace{-0.5em}
\caption{Reward comparison between original QMIX and QMIX combined with our \ours in FoodCollector with discrete action space, under no attacker or non-adaptive/adaptive attacks under varying numbers of adversaries. For \ours, the ablation size $k$ is set as 2.}
\vspace{-0.5em}
\label{fig:res_qmix}
\end{figure}

%% file: figs/QMIX_non_adaptive.tex
\begin{tikzpicture}

\definecolor{color0}{rgb}{0.56078431372549,0.737254901960784,0.56078431372549}
\definecolor{color1}{rgb}{1,0.388235294117647,0.27843137254902}

\begin{axis}[
axis line style={white!15!black},
legend cell align={left},
legend style={fill opacity=1, draw opacity=1, text opacity=1, draw=none, fill=none, at={(1,1)}, anchor=north east},
tick align=outside,
x grid style={white!80!black},
xtick style={color=white!15!black},
xmin=-0.37, xmax=3.37,
xtick={0,1,2,3},
xticklabels={No Attacker,C=1,C=2,C=3},
xtick pos = lower,
y grid style={white!80!black},
ymin=0, ymax=100,
ytick style={color=black},
ytick={0,20,40,60,80,100},
yticklabels={-100,-80,-60,-40,-20,0},
ytick pos = left,
ylabel = {\textbf{Reward}}
]
\draw[draw=black,fill=color0] (axis cs:-0.2,0) rectangle (axis cs:0,88);
\draw[draw=black,fill=color0] (axis cs:0.8,0) rectangle (axis cs:1,53);
\draw[draw=black,fill=color0] (axis cs:1.8,0) rectangle (axis cs:2,49);
\draw[draw=black,fill=color0] (axis cs:2.8,0) rectangle (axis cs:3,45);
\draw[draw=black,fill=color1] (axis cs:-1.38777878078145e-17,0) rectangle (axis cs:0.2,77);
\draw[draw=black,fill=color1] (axis cs:1,0) rectangle (axis cs:1.2,76);
\draw[draw=black,fill=color1] (axis cs:2,0) rectangle (axis cs:2.2,76);
\draw[draw=black,fill=color1] (axis cs:3,0) rectangle (axis cs:3.2,74);
\path [draw=black, semithick]
(axis cs:-0.1,87)
--(axis cs:-0.1,89);

\path [draw=black, semithick]
(axis cs:0.9,50)
--(axis cs:0.9,56);

\path [draw=black, semithick]
(axis cs:1.9,46)
--(axis cs:1.9,52);

\path [draw=black, semithick]
(axis cs:2.9,41)
--(axis cs:2.9,49);

\path [draw=black, semithick]
(axis cs:0.1,76)
--(axis cs:0.1,78);

\path [draw=black, semithick]
(axis cs:1.1,72)
--(axis cs:1.1,80);

\path [draw=black, semithick]
(axis cs:2.1,71)
--(axis cs:2.1,81);

\path [draw=black, semithick]
(axis cs:3.1,68)
--(axis cs:3.1,80);

\addplot [semithick, black, mark=-, mark size=6, mark options={solid}, only marks, forget plot]
table {%
-0.1 87
0.9 50
1.9 46
2.9 41
};
\addplot [semithick, black, mark=-, mark size=6, mark options={solid}, only marks, forget plot]
table {%
-0.1 89
0.9 56
1.9 52
2.9 49
};
\addplot [semithick, black, mark=-, mark size=6, mark options={solid}, only marks, forget plot]
table {%
0.1 76
1.1 72
2.1 71
3.1 68
};
\addplot [semithick, black, mark=-, mark size=6, mark options={solid}, only marks, forget plot]
table {%
0.1 78
1.1 80
2.1 81
3.1 80
};
\addplot [semithick, green!50!black, mark=*, mark size=3, mark options={solid}]
table {%
-0.1 88
0.9 53
1.9 49
2.9 45
};
\addlegendentry{QMIX}
\addplot [semithick, red, mark=square*, mark size=3, mark options={solid}]
table {%
0.1 77
1.1 76
2.1 76
3.1 74
};
\addlegendentry{QMIX+AME (ours)}
\end{axis}

\end{tikzpicture}

%% file: figs/QMIX_adaptive.tex
\begin{tikzpicture}

\definecolor{color0}{rgb}{0.56078431372549,0.737254901960784,0.56078431372549}
\definecolor{color1}{rgb}{1,0.388235294117647,0.27843137254902}

\begin{axis}[
axis line style={white!15!black},
legend cell align={left},
legend style={fill opacity=1, draw opacity=1, text opacity=1, draw=none, fill=none, at={(1,1)}, anchor=north east},
tick align=outside,
x grid style={white!80!black},
xtick style={color=white!15!black},
xmin=-0.37, xmax=3.37,
xtick={0,1,2,3},
xticklabels={No Attacker,C=1,C=2,C=3},
xtick pos = lower,
y grid style={white!80!black},
ymin=0, ymax=100,
ytick style={color=black},
ytick={0,20,40,60,80,100},
yticklabels={-100,-80,-60,-40,-20,0},
ytick pos = left,
ylabel = {\textbf{Reward}}
]
\draw[draw=black,fill=color0] (axis cs:-0.2,0) rectangle (axis cs:0,88);
\draw[draw=black,fill=color0] (axis cs:0.8,0) rectangle (axis cs:1,50);
\draw[draw=black,fill=color0] (axis cs:1.8,0) rectangle (axis cs:2,28);
\draw[draw=black,fill=color0] (axis cs:2.8,0) rectangle (axis cs:3,24);
\draw[draw=black,fill=color1] (axis cs:-1.38777878078145e-17,0) rectangle (axis cs:0.2,77);
\draw[draw=black,fill=color1] (axis cs:1,0) rectangle (axis cs:1.2,72);
\draw[draw=black,fill=color1] (axis cs:2,0) rectangle (axis cs:2.2,70);
\draw[draw=black,fill=color1] (axis cs:3,0) rectangle (axis cs:3.2,58);
\path [draw=black, semithick]
(axis cs:-0.1,87)
--(axis cs:-0.1,89);

\path [draw=black, semithick]
(axis cs:0.9,46)
--(axis cs:0.9,54);

\path [draw=black, semithick]
(axis cs:1.9,25)
--(axis cs:1.9,31);

\path [draw=black, semithick]
(axis cs:2.9,19)
--(axis cs:2.9,29);

\path [draw=black, semithick]
(axis cs:0.1,76)
--(axis cs:0.1,78);

\path [draw=black, semithick]
(axis cs:1.1,70)
--(axis cs:1.1,74);

\path [draw=black, semithick]
(axis cs:2.1,65)
--(axis cs:2.1,75);

\path [draw=black, semithick]
(axis cs:3.1,54)
--(axis cs:3.1,62);

\addplot [semithick, black, mark=-, mark size=6, mark options={solid}, only marks, forget plot]
table {%
-0.1 87
0.9 46
1.9 25
2.9 19
};
\addplot [semithick, black, mark=-, mark size=6, mark options={solid}, only marks, forget plot]
table {%
-0.1 89
0.9 54
1.9 31
2.9 29
};
\addplot [semithick, black, mark=-, mark size=6, mark options={solid}, only marks, forget plot]
table {%
0.1 76
1.1 70
2.1 65
3.1 54
};
\addplot [semithick, black, mark=-, mark size=6, mark options={solid}, only marks, forget plot]
table {%
0.1 78
1.1 74
2.1 75
3.1 62
};
\addplot [semithick, green!50!black, mark=*, mark size=3, mark options={solid}]
table {%
-0.1 88
0.9 50
1.9 28
2.9 24
};
\addlegendentry{QMIX}
\addplot [semithick, red, mark=square*, mark size=3, mark options={solid}]
table {%
0.1 77
1.1 72
2.1 70
3.1 58
};
\addlegendentry{QMIX+AME (ours)}
\end{axis}

\end{tikzpicture}

%% file: figs/mnist/MNIST_Adaptive_Attacker_Recall.tex
\begin{tikzpicture}

\definecolor{darkgray176}{RGB}{176,176,176}
\definecolor{darkseagreen}{RGB}{143,188,143}
\definecolor{deepskyblue}{RGB}{0,191,255}
\definecolor{green01270}{RGB}{0,127,0}
\definecolor{lightgray204}{RGB}{204,204,204}
\definecolor{tomato}{RGB}{255,99,71}

\begin{axis}[
legend cell align={left},
legend style={fill opacity=1, draw opacity=1, text opacity=1, draw=lightgray204},
tick align=outside,
tick pos=both,
x grid style={darkgray176},
xmin=-0.48, xmax=3.48,
xtick style={color=black},
xtick={0,1,2,3},
xticklabels={No Attacker,C=1,C=2,C=3},
y grid style={darkgray176},
ylabel={Recall},
ymin=0.4, ymax=1.15,
ytick style={color=black}
]
\draw[draw=black,fill=darkseagreen] (axis cs:-0.3,0) rectangle (axis cs:-0.1,0.9436);
\draw[draw=black,fill=darkseagreen] (axis cs:0.7,0) rectangle (axis cs:0.9,0.916);
\draw[draw=black,fill=darkseagreen] (axis cs:1.7,0) rectangle (axis cs:1.9,0.836);
\draw[draw=black,fill=darkseagreen] (axis cs:2.7,0) rectangle (axis cs:2.9,0.594);
\draw[draw=black,fill=deepskyblue] (axis cs:-0.1,0) rectangle (axis cs:0.1,0.9354);
\draw[draw=black,fill=deepskyblue] (axis cs:0.9,0) rectangle (axis cs:1.1,0.898);
\draw[draw=black,fill=deepskyblue] (axis cs:1.9,0) rectangle (axis cs:2.1,0.766);
\draw[draw=black,fill=deepskyblue] (axis cs:2.9,0) rectangle (axis cs:3.1,0.47);
\draw[draw=black,fill=tomato] (axis cs:0.1,0) rectangle (axis cs:0.3,0.9306);
\draw[draw=black,fill=tomato] (axis cs:1.1,0) rectangle (axis cs:1.3,0.928);
\draw[draw=black,fill=tomato] (axis cs:2.1,0) rectangle (axis cs:2.3,0.9242);
\draw[draw=black,fill=tomato] (axis cs:3.1,0) rectangle (axis cs:3.3,0.9028);
\path [draw=black, semithick]
(axis cs:-0.2,0.938064658998761)
--(axis cs:-0.2,0.949135341001239);

\path [draw=black, semithick]
(axis cs:0.8,0.893550055679356)
--(axis cs:0.8,0.938449944320644);

\path [draw=black, semithick]
(axis cs:1.8,0.791909184629903)
--(axis cs:1.8,0.880090815370097);

\path [draw=black, semithick]
(axis cs:2.8,0.470452438308156)
--(axis cs:2.8,0.717547561691844);

\path [draw=black, semithick]
(axis cs:0,0.928035218944191)
--(axis cs:0,0.942764781055809);

\path [draw=black, semithick]
(axis cs:1,0.890516685226452)
--(axis cs:1,0.905483314773548);

\path [draw=black, semithick]
(axis cs:2,0.684366673472166)
--(axis cs:2,0.847633326527834);

\path [draw=black, semithick]
(axis cs:3,0.351509494051211)
--(axis cs:3,0.588490505948789);

\path [draw=black, semithick]
(axis cs:0.2,0.920933563221125)
--(axis cs:0.2,0.940266436778876);

\path [draw=black, semithick]
(axis cs:1.2,0.91633809621031)
--(axis cs:1.2,0.939661903789691);

\path [draw=black, semithick]
(axis cs:2.2,0.913685248457524)
--(axis cs:2.2,0.934714751542476);

\path [draw=black, semithick]
(axis cs:3.2,0.892247275233382)
--(axis cs:3.2,0.913352724766618);

\addplot [semithick, black, mark=-, mark size=6, mark options={solid}, only marks, forget plot]
table {%
-0.2 0.938064658998761
0.8 0.893550055679356
1.8 0.791909184629903
2.8 0.470452438308156
};
\addplot [semithick, black, mark=-, mark size=6, mark options={solid}, only marks, forget plot]
table {%
-0.2 0.949135341001239
0.8 0.938449944320644
1.8 0.880090815370097
2.8 0.717547561691844
};
\addplot [semithick, black, mark=-, mark size=6, mark options={solid}, only marks, forget plot]
table {%
0 0.928035218944191
1 0.890516685226452
2 0.684366673472166
3 0.351509494051211
};
\addplot [semithick, black, mark=-, mark size=6, mark options={solid}, only marks, forget plot]
table {%
0 0.942764781055809
1 0.905483314773548
2 0.847633326527834
3 0.588490505948789
};
\addplot [semithick, black, mark=-, mark size=6, mark options={solid}, only marks, forget plot]
table {%
0.2 0.920933563221125
1.2 0.91633809621031
2.2 0.913685248457524
3.2 0.892247275233382
};
\addplot [semithick, black, mark=-, mark size=6, mark options={solid}, only marks, forget plot]
table {%
0.2 0.940266436778876
1.2 0.939661903789691
2.2 0.934714751542476
3.2 0.913352724766618
};
\addplot [semithick, green01270, mark=*, mark size=3, mark options={solid}]
table {%
-0.2 0.9436
0.8 0.916
1.8 0.836
2.8 0.594
};
\addlegendentry{Vanilla}
\addplot [semithick, blue, mark=x, mark size=3, mark options={solid}]
table {%
0 0.9354
1 0.898
2 0.766
3 0.47
};
\addlegendentry{AT}
\addplot [semithick, red, mark=square*, mark size=3, mark options={solid}]
table {%
0.2 0.9306
1.2 0.928
2.2 0.9242
3.2 0.9028
};
\addlegendentry{AME (ours)}
\end{axis}

\end{tikzpicture}

%% file: figs/mnist/MNIST_Random_Attacker_Precision.tex
\begin{tikzpicture}

\definecolor{darkgray176}{RGB}{176,176,176}
\definecolor{darkseagreen}{RGB}{143,188,143}
\definecolor{deepskyblue}{RGB}{0,191,255}
\definecolor{green01270}{RGB}{0,127,0}
\definecolor{lightgray204}{RGB}{204,204,204}
\definecolor{tomato}{RGB}{255,99,71}

\begin{axis}[
legend cell align={left},
legend style={fill opacity=1, draw opacity=1, text opacity=1, draw=lightgray204},
tick align=outside,
tick pos=both,
x grid style={darkgray176},
xmin=-0.48, xmax=3.48,
xtick style={color=black},
xtick={0,1,2,3},
xticklabels={No Attacker,C=2,C=4,C=6},
y grid style={darkgray176},
ylabel={Precision},
ymin=0.75, ymax=1.02,
ytick style={color=black}
]
\draw[draw=black,fill=darkseagreen] (axis cs:-0.3,0) rectangle (axis cs:-0.1,0.9428);
\draw[draw=black,fill=darkseagreen] (axis cs:0.7,0) rectangle (axis cs:0.9,0.944);
\draw[draw=black,fill=darkseagreen] (axis cs:1.7,0) rectangle (axis cs:1.9,0.914);
\draw[draw=black,fill=darkseagreen] (axis cs:2.7,0) rectangle (axis cs:2.9,0.834);
\draw[draw=black,fill=deepskyblue] (axis cs:-0.1,0) rectangle (axis cs:0.1,0.935);
\draw[draw=black,fill=deepskyblue] (axis cs:0.9,0) rectangle (axis cs:1.1,0.932);
\draw[draw=black,fill=deepskyblue] (axis cs:1.9,0) rectangle (axis cs:2.1,0.894);
\draw[draw=black,fill=deepskyblue] (axis cs:2.9,0) rectangle (axis cs:3.1,0.82);
\draw[draw=black,fill=tomato] (axis cs:0.1,0) rectangle (axis cs:0.3,0.9324);
\draw[draw=black,fill=tomato] (axis cs:1.1,0) rectangle (axis cs:1.3,0.93);
\draw[draw=black,fill=tomato] (axis cs:2.1,0) rectangle (axis cs:2.3,0.922);
\draw[draw=black,fill=tomato] (axis cs:3.1,0) rectangle (axis cs:3.3,0.9);
\path [draw=black, semithick]
(axis cs:-0.2,0.937925576957219)
--(axis cs:-0.2,0.947674423042782);

\path [draw=black, semithick]
(axis cs:0.8,0.939101020514434)
--(axis cs:0.8,0.948898979485566);

\path [draw=black, semithick]
(axis cs:1.8,0.899033370452904)
--(axis cs:1.8,0.928966629547096);

\path [draw=black, semithick]
(axis cs:2.8,0.786)
--(axis cs:2.8,0.882);

\path [draw=black, semithick]
(axis cs:0,0.927624364434166)
--(axis cs:0,0.942375635565834);

\path [draw=black, semithick]
(axis cs:1,0.924516685226452)
--(axis cs:1,0.939483314773548);

\path [draw=black, semithick]
(axis cs:2,0.886)
--(axis cs:2,0.902);

\path [draw=black, semithick]
(axis cs:3,0.797196491498017)
--(axis cs:3,0.842803508501983);

\path [draw=black, semithick]
(axis cs:0.2,0.924460226703488)
--(axis cs:0.2,0.940339773296512);

\path [draw=black, semithick]
(axis cs:1.2,0.921055728090001)
--(axis cs:1.2,0.938944271909999);

\path [draw=black, semithick]
(axis cs:2.2,0.904795349465915)
--(axis cs:2.2,0.939204650534085);

\path [draw=black, semithick]
(axis cs:3.2,0.891055728090001)
--(axis cs:3.2,0.908944271909999);

\addplot [semithick, black, mark=-, mark size=6, mark options={solid}, only marks, forget plot]
table {%
-0.2 0.937925576957219
0.8 0.939101020514434
1.8 0.899033370452904
2.8 0.786
};
\addplot [semithick, black, mark=-, mark size=6, mark options={solid}, only marks, forget plot]
table {%
-0.2 0.947674423042782
0.8 0.948898979485566
1.8 0.928966629547096
2.8 0.882
};
\addplot [semithick, black, mark=-, mark size=6, mark options={solid}, only marks, forget plot]
table {%
0 0.927624364434166
1 0.924516685226452
2 0.886
3 0.797196491498017
};
\addplot [semithick, black, mark=-, mark size=6, mark options={solid}, only marks, forget plot]
table {%
0 0.942375635565834
1 0.939483314773548
2 0.902
3 0.842803508501983
};
\addplot [semithick, black, mark=-, mark size=6, mark options={solid}, only marks, forget plot]
table {%
0.2 0.924460226703488
1.2 0.921055728090001
2.2 0.904795349465915
3.2 0.891055728090001
};
\addplot [semithick, black, mark=-, mark size=6, mark options={solid}, only marks, forget plot]
table {%
0.2 0.940339773296512
1.2 0.938944271909999
2.2 0.939204650534085
3.2 0.908944271909999
};
\addplot [semithick, green01270, mark=*, mark size=3, mark options={solid}]
table {%
-0.2 0.9428
0.8 0.944
1.8 0.914
2.8 0.834
};
\addlegendentry{Vanilla}
\addplot [semithick, blue, mark=x, mark size=3, mark options={solid}]
table {%
0 0.935
1 0.932
2 0.894
3 0.82
};
\addlegendentry{AT}
\addplot [semithick, red, mark=square*, mark size=3, mark options={solid}]
table {%
0.2 0.9324
1.2 0.93
2.2 0.922
3.2 0.9
};
\addlegendentry{AME (ours)}
\end{axis}

\end{tikzpicture}

%% file: figs/mnist/MNIST_Random_Attacker_Recall.tex
\begin{tikzpicture}

\definecolor{darkgray176}{RGB}{176,176,176}
\definecolor{darkseagreen}{RGB}{143,188,143}
\definecolor{deepskyblue}{RGB}{0,191,255}
\definecolor{green01270}{RGB}{0,127,0}
\definecolor{lightgray204}{RGB}{204,204,204}
\definecolor{tomato}{RGB}{255,99,71}

\begin{axis}[
legend cell align={left},
legend style={fill opacity=1, draw opacity=1, text opacity=1, draw=lightgray204},
tick align=outside,
tick pos=both,
x grid style={darkgray176},
xmin=-0.48, xmax=3.48,
xtick style={color=black},
xtick={0,1,2,3},
xticklabels={No Attacker,C=2,C=4,C=6},
y grid style={darkgray176},
ylabel={Recall},
ymin=0.75, ymax=1.02,
ytick style={color=black}
]
\draw[draw=black,fill=darkseagreen] (axis cs:-0.3,0) rectangle (axis cs:-0.1,0.9436);
\draw[draw=black,fill=darkseagreen] (axis cs:0.7,0) rectangle (axis cs:0.9,0.944);
\draw[draw=black,fill=darkseagreen] (axis cs:1.7,0) rectangle (axis cs:1.9,0.914);
\draw[draw=black,fill=darkseagreen] (axis cs:2.7,0) rectangle (axis cs:2.9,0.794);
\draw[draw=black,fill=deepskyblue] (axis cs:-0.1,0) rectangle (axis cs:0.1,0.9354);
\draw[draw=black,fill=deepskyblue] (axis cs:0.9,0) rectangle (axis cs:1.1,0.932);
\draw[draw=black,fill=deepskyblue] (axis cs:1.9,0) rectangle (axis cs:2.1,0.894);
\draw[draw=black,fill=deepskyblue] (axis cs:2.9,0) rectangle (axis cs:3.1,0.81);
\draw[draw=black,fill=tomato] (axis cs:0.1,0) rectangle (axis cs:0.3,0.9306);
\draw[draw=black,fill=tomato] (axis cs:1.1,0) rectangle (axis cs:1.3,0.932);
\draw[draw=black,fill=tomato] (axis cs:2.1,0) rectangle (axis cs:2.3,0.918);
\draw[draw=black,fill=tomato] (axis cs:3.1,0) rectangle (axis cs:3.3,0.894);
\path [draw=black, semithick]
(axis cs:-0.2,0.938064658998761)
--(axis cs:-0.2,0.949135341001239);

\path [draw=black, semithick]
(axis cs:0.8,0.939101020514434)
--(axis cs:0.8,0.948898979485566);

\path [draw=black, semithick]
(axis cs:1.8,0.899033370452904)
--(axis cs:1.8,0.928966629547096);

\path [draw=black, semithick]
(axis cs:2.8,0.715107668306736)
--(axis cs:2.8,0.872892331693264);

\path [draw=black, semithick]
(axis cs:0,0.928035218944191)
--(axis cs:0,0.942764781055809);

\path [draw=black, semithick]
(axis cs:1,0.924516685226452)
--(axis cs:1,0.939483314773548);

\path [draw=black, semithick]
(axis cs:2,0.882)
--(axis cs:2,0.906);

\path [draw=black, semithick]
(axis cs:3,0.78243190249582)
--(axis cs:3,0.83756809750418);

\path [draw=black, semithick]
(axis cs:0.2,0.920933563221125)
--(axis cs:0.2,0.940266436778876);

\path [draw=black, semithick]
(axis cs:1.2,0.924516685226452)
--(axis cs:1.2,0.939483314773548);

\path [draw=black, semithick]
(axis cs:2.2,0.897603921945629)
--(axis cs:2.2,0.938396078054371);

\path [draw=black, semithick]
(axis cs:3.2,0.882)
--(axis cs:3.2,0.906);

\addplot [semithick, black, mark=-, mark size=6, mark options={solid}, only marks, forget plot]
table {%
-0.2 0.938064658998761
0.8 0.939101020514434
1.8 0.899033370452904
2.8 0.715107668306736
};
\addplot [semithick, black, mark=-, mark size=6, mark options={solid}, only marks, forget plot]
table {%
-0.2 0.949135341001239
0.8 0.948898979485566
1.8 0.928966629547096
2.8 0.872892331693264
};
\addplot [semithick, black, mark=-, mark size=6, mark options={solid}, only marks, forget plot]
table {%
0 0.928035218944191
1 0.924516685226452
2 0.882
3 0.78243190249582
};
\addplot [semithick, black, mark=-, mark size=6, mark options={solid}, only marks, forget plot]
table {%
0 0.942764781055809
1 0.939483314773548
2 0.906
3 0.83756809750418
};
\addplot [semithick, black, mark=-, mark size=6, mark options={solid}, only marks, forget plot]
table {%
0.2 0.920933563221125
1.2 0.924516685226452
2.2 0.897603921945629
3.2 0.882
};
\addplot [semithick, black, mark=-, mark size=6, mark options={solid}, only marks, forget plot]
table {%
0.2 0.940266436778876
1.2 0.939483314773548
2.2 0.938396078054371
3.2 0.906
};
\addplot [semithick, green01270, mark=*, mark size=3, mark options={solid}]
table {%
-0.2 0.9436
0.8 0.944
1.8 0.914
2.8 0.794
};
\addlegendentry{Vanilla}
\addplot [semithick, blue, mark=x, mark size=3, mark options={solid}]
table {%
0 0.9354
1 0.932
2 0.894
3 0.81
};
\addlegendentry{AT}
\addplot [semithick, red, mark=square*, mark size=3, mark options={solid}]
table {%
0.2 0.9306
1.2 0.932
2.2 0.918
3.2 0.894
};
\addlegendentry{AME (ours)}
\end{axis}

\end{tikzpicture}

%% file: a4_hyper.tex
\section{Additional Discussion on the Choice of Ablation Size $k$}
\label{app:hyper}

Our \ours defensive mechanism proposed in Section~\ref{sec:algo_overview} requires an extra hyperparameter: the ablation size $k$. (The sample size $D$ is only for the partial-sample variant introduced in Section~\ref{sec:algo_more}, not needed in the original form of \ours.) In this section, we discuss the selection of $k$ in practice. We start by discussing the relationship between $k$ and the required theoretical conditions.


\textbf{I. (For Continuous Action Space) Condition~\ref{cond:k} and $\boldsymbol{k}$}

We first decompose Equation~\eqref{eq:k_cond} as below.
\begin{align}
    \binom{N-1-C}{k} &> \frac{1}{2} \binom{N-1}{k} \\
    \frac{(N-1-C)!}{(N-1-C-k)!k!} &> \frac{(N-1)!}{2(N-1-k)!k!} \\
    1 &> \frac{(N-1)\cdots(N-C)}{2(N-k-1)\cdots(N-k-C)}
\end{align}
Therefore, Equation~\eqref{eq:k_cond} is equivalent to
\begin{equation}
\label{eq:k_1}
    (N-k-1)\cdots(N-k-C) > \frac{1}{2} (N-1)\cdots(N-C)
\end{equation}

When $k=1$, the above inequality becomes
\begin{align}
    (N-2)\cdots(N-C)(N-C-1) &> \frac{1}{2}(N-1)(N-2)\cdots(N-C) \\
    (N-C-1) &> \frac{1}{2}(N-1) \\
    \frac{1}{2}(N-1) &> C,
\end{align}
which holds under Assumption~\ref{assum:num}. Therefore, \textbf{$\boldsymbol{k=1}$ is always a feasible solution for Condition~\ref{cond:k}.}

Then, when $C$ is fixed and $k$ goes up from 1, the LHS of Equation~\eqref{eq:k_1} goes down while the RHS does not change. Therefore, for a given number of agents ($N$) and a fixed number of adversaries ($C$), there exists an integer $k_0$ such that any $k \leq k_0$ satisfies Condition~\ref{cond:k}. In practice, if we have an estimate of the number of adversarial messages that we would like to defend against, then we could choose the maximum $k$ satisfying Equation \eqref{eq:k_cond}.

On the other hand, when $k$ is fixed, there exists a $C_0$ such that any $C \leq C_0$ can let Equation~\eqref{eq:k_1} hold. Therefore, if $C$ is unknown, for any selection of $k$, Equation~\eqref{eq:k_1} can justify the maximum number of adversaries for the current selection.

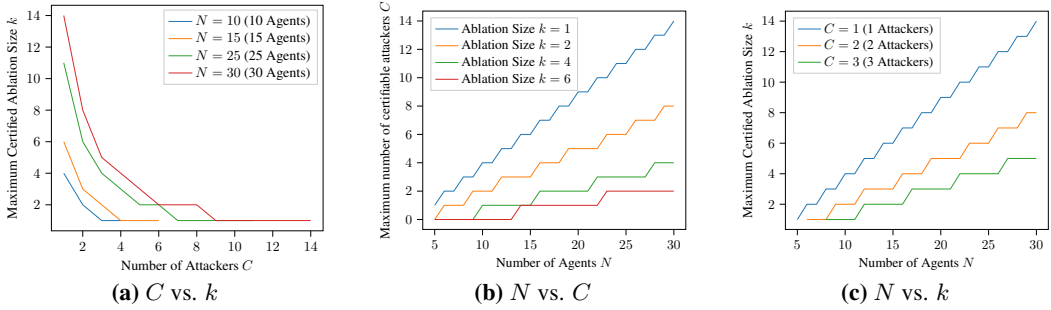
\begin{figure*}[h]
 \centering
    \begin{subfigure}[t]{0.31\textwidth}
    \centering
     \resizebox{\textwidth}{!}{\input{figs/Ablation_C_K}}
     \vspace{-0.5cm}
      \subcaption{$C$ vs. $k$}
      \label{sfig:C&K}
    \end{subfigure}
    \hfill
    \begin{subfigure}[t]{0.30\textwidth}
     \centering
     \resizebox{\textwidth}{!}{\input{figs/Ablation_N_C}}
     \label{sfig:N&C}
     \vspace{-0.5cm}
      \subcaption{$N$ vs. $C$}
    \end{subfigure}
    \hfill
    \begin{subfigure}[t]{0.30\textwidth}
     \centering
     \resizebox{\textwidth}{!}{\input{figs/Ablation_N_K}}
     \label{sfig:N&K}
     \vspace{-0.5cm}
     \subcaption{$N$ vs. $k$}
    \end{subfigure}
\caption{Relationship between the total number of agents $N$, number of attackers $C$, and ablation size $k$. \textbf{(a)}: Maximum certifiable ablation size $k$ under different number of attackers. \textbf{(b)}: Maximum defensible number of attackers $C$ with different total number of agents $N$. \textbf{(c)}: Maximum certifiable ablation size $k$ with different total number of agents $N$.}
\label{fig:ablation_cnk}
\end{figure*}

From Equation~\eqref{eq:k_1}, we can see the interdependence between three parameters: total number of agents $N$, number of adversaries $C$, and the ablation size $k$. In Figure \ref{fig:ablation_cnk} we visualize their relationship by fixing one variable at a time. From these figures, we can see an obvious trade-off between $C$ and $k$ --- if there are more adversaries, $k$ has to be set smaller. But when $C$ is small, e.g. $C=1$, $k$ can be relatively large, so the agent does not need to compromise much natural performance to achieve robustness. 

\textbf{II. (For Discrete Action Space) Condition~\ref{cond:cons} and $\boldsymbol{k}$}

Different from Condition~\ref{cond:k} which is independent of the environment, Condition~\ref{cond:cons} required in a discrete action space is related to the environment and the communication quality. Intuitively, Condition~\ref{cond:cons} can be satisfied if the benign messages can reach some ``consensus'', i.e., there are enough purely benign \ksamples voting for the same action. This can be achieved when the environment is relatively deterministic (e.g., there is a certainly optimal direction to go). Fortunately, Condition~\ref{cond:cons} can be checked during training time, and we can train the \ablname and the communication policy to increase $u_{\max}$ as much as possible. 

On the other hand, Condition~\ref{cond:cons} is also related to the selection of ablation size $k$. The ratio of contaminated votes is $\frac{\binom{N-1}{k}-\binom{N-1-C}{k}}{\binom{N-1}{k}}$, and it is easy to show that this rate increases as $k$ when $k \leq N-C-1$. Therefore, when $k$ is smaller, it is relatively easier to satisfy Condition~\ref{cond:cons} as the adversarial messages can take over a smaller proportion of the total number of votes.

\textbf{Selecting $\boldsymbol{k}$ to Balance between Natural Performance and Robustness}

The above analysis shows that a smaller $k$ makes the agent more robust, while the natural performance may be sacrificed as the \ablname makes decisions based on less benign information. Such a trade-off between natural performance and robustness is common in the literature of adversarial learning~\cite{tsipras2018robustness,zhang2019theoretically}. \\
In practice, we suggest setting $k$ to be the largest integer satisfying Equation~\eqref{eq:k_cond}. If higher robustness is needed, then $k$ can be further decreased. If robustness is not the major concern while higher natural performance is required, one can increase $k$. Note that if $k=N-1$, \ours degenerates to the original vanilla policy without defense.

\textbf{What If Conditions Are Not Satisfied?}
Even if Condition~\ref{cond:k} and Condition~\ref{cond:cons} are not satisfied, the agent can still be robust under attacks as verified in our experiments (\ours with $k=2$ still achieves relatively robust performance under $C=3$ which exceeds the theoretically largest number of certifiable adversaries). Because these conditions are needed for the certificates which consider the theoretically worst-case attacks. However, in practice, an attacker has restricted power and knowledge (e.g., it does not know the victim policy/reward, and does not know the environment dynamics as prior), and is likely to be even weaker than the learned adaptive white-box attacker we use in experiments. As a result, even if a larger $k$ may break the conditions, it can still improve the empirical robustness of an agent in practice.

\textbf{Extension: Adaptive Defense with Different $k$'s}\quad
Moreover, to allow higher flexibility, one can train multiple \ablnames with different selections of $k$'s during training. Then, an adaptive strategy can be used in test time. For example, if $u_{\max}$ is too small, we can use a larger $k$ with the corresponding trained \ablname.

\textbf{Extension: Gaining both Natural Performance and Robustness by Attack Detection}\quad
From the analysis of the relation between $N$, $C$ and $k$, we can see that when the number of adversaries is large, the corresponding ablation size $k$ is supposed to be smaller. This is reasonable because a more conservative defense is needed against a stronger attacker. But if we can identify some adversarial messages and rule them out before message ablation and ensemble, we can still defend with guarantees using a relatively large $k$. For example, if we have identified $c$ adversarial messages, then we only need to deal with the remaining $C-c$ adversarial messages out of $N-1-c$ messages. By Equation~\eqref{eq:k_cond}, a larger $k$ can be used compared to defending $C$ adversarial messages out of $N-1$ messages. We also provide an adversary detection algorithm in Appendix~\ref{app:exp_detect} using a similar idea of \ours.

%% file: figs/Ablation_C_K.tex
\begin{tikzpicture}[scale=0.7]

\definecolor{color0}{rgb}{0.12156862745098,0.466666666666667,0.705882352941177}
\definecolor{color1}{rgb}{1,0.498039215686275,0.0549019607843137}
\definecolor{color2}{rgb}{0.172549019607843,0.627450980392157,0.172549019607843}
\definecolor{color3}{rgb}{0.83921568627451,0.152941176470588,0.156862745098039}

\begin{axis}[
legend cell align={left},    
legend style={fill opacity=1, draw opacity=1, text opacity=1, draw=white!80!black},
tick align=outside,
tick pos=left,
x grid style={white!69.0196078431373!black},
xlabel={Number of Attackers \(\displaystyle C\)},
xmin=0.35, xmax=14.65,
xtick pos = lower,
xtick style={color=black},
y grid style={white!69.0196078431373!black},
ylabel={Maximum Certified Ablation Size \(\displaystyle k\)},
ymin=0.35, ymax=14.65,
ytick style={color=black},
ytick pos = left,
ytick = {2,4,6,8,10,12,14}
]
\addplot [semithick, color0]
table {%
1 4
2 2
3 1
4 1
};
\addlegendentry{$N=10$ (10 Agents)}
\addplot [semithick, color1]
table {%
1 6
2 3
3 2
4 1
5 1
6 1
};
\addlegendentry{$N=15$ (15 Agents)}
\addplot [semithick, color2]
table {%
1 11
2 6
3 4
4 3
5 2
6 2
7 1
8 1
9 1
10 1
11 1
};
\addlegendentry{$N=25$ (25 Agents)}
\addplot [semithick, color3]
table {%
1 14
2 8
3 5
4 4
5 3
6 2
7 2
8 2
9 1
10 1
11 1
12 1
13 1
14 1
};
\addlegendentry{$N=30$ (30 Agents)}
\end{axis}

\end{tikzpicture}

%% file: figs/Ablation_N_C.tex
\begin{tikzpicture}[scale=0.7]

\definecolor{color0}{rgb}{0.12156862745098,0.466666666666667,0.705882352941177}
\definecolor{color1}{rgb}{1,0.498039215686275,0.0549019607843137}
\definecolor{color2}{rgb}{0.172549019607843,0.627450980392157,0.172549019607843}
\definecolor{color3}{rgb}{0.83921568627451,0.152941176470588,0.156862745098039}

\begin{axis}[
legend cell align={left},
legend style={
  fill opacity=1,
  draw opacity=1,
  text opacity=1,
  at={(0.03,0.97)},
  anchor=north west,
  draw=white!80!black
},
tick align=outside,
tick pos=left,
x grid style={white!69.0196078431373!black},
xlabel={Number of Agents \(\displaystyle N\)},
xmin=3.75, xmax=31.25,
xtick style={color=black},
xtick pos = lower,
y grid style={white!69.0196078431373!black},
ylabel={Maximum number of certifiable attackers \(\displaystyle C\)},
ymin=-0.7, ymax=14.7,
ytick style={color=black},
ytick pos = left,
ytick = {0, 2,4,6,8,10,12,14}
]
\addplot [semithick, color0]
table {%
5 1
6 2
7 2
8 3
9 3
10 4
11 4
12 5
13 5
14 6
15 6
16 7
17 7
18 8
19 8
20 9
21 9
22 10
23 10
24 11
25 11
26 12
27 12
28 13
29 13
30 14
};
\addlegendentry{Ablation Size $k=1$}
\addplot [semithick, color1]
table {%
5 0
6 1
7 1
8 1
9 2
10 2
11 2
12 3
13 3
14 3
15 3
16 4
17 4
18 4
19 5
20 5
21 5
22 5
23 6
24 6
25 6
26 7
27 7
28 7
29 8
30 8
};
\addlegendentry{Ablation Size $k=2$}
\addplot [semithick, color2]
table {%
5 0
6 0
7 0
8 0
9 0
10 1
11 1
12 1
13 1
14 1
15 1
16 2
17 2
18 2
19 2
20 2
21 2
22 3
23 3
24 3
25 3
26 3
27 3
28 4
29 4
30 4
};
\addlegendentry{Ablation Size $k=4$}
\addplot [semithick, color3]
table {%
5 0
6 0
7 0
8 0
9 0
10 0
11 0
12 0
13 0
14 1
15 1
16 1
17 1
18 1
19 1
20 1
21 1
22 1
23 2
24 2
25 2
26 2
27 2
28 2
29 2
30 2
};
\addlegendentry{Ablation Size $k=6$}
\end{axis}

\end{tikzpicture}

%% file: figs/Ablation_N_K.tex
\begin{tikzpicture}[scale=0.7]

\definecolor{color0}{rgb}{0.12156862745098,0.466666666666667,0.705882352941177}
\definecolor{color1}{rgb}{1,0.498039215686275,0.0549019607843137}
\definecolor{color2}{rgb}{0.172549019607843,0.627450980392157,0.172549019607843}

\begin{axis}[
legend cell align={left},
legend style={
  fill opacity=1,
  draw opacity=1,
  text opacity=1,
  at={(0.03,0.97)},
  anchor=north west,
  draw=white!80!black
},
tick align=outside,
tick pos=left,
unbounded coords=jump,
x grid style={white!69.0196078431373!black},
xlabel={Number of Agents \(\displaystyle N\)},
xmin=3.75, xmax=31.25,
xtick style={color=black},
xtick pos = lower,
y grid style={white!69.0196078431373!black},
ylabel={Maximum Certified Ablation Size \(\displaystyle k\)},
ymin=0.35, ymax=14.65,
ytick style={color=black},
ytick pos = left,
ytick = {2,4,6,8,10,12,14}
]
\addplot [semithick, color0]
table {%
5 1
6 2
7 2
8 3
9 3
10 4
11 4
12 5
13 5
14 6
15 6
16 7
17 7
18 8
19 8
20 9
21 9
22 10
23 10
24 11
25 11
26 12
27 12
28 13
29 13
30 14
};
\addlegendentry{$C=1$ (1 Attackers)}
\addplot [semithick, color1]
table {%
5 nan
6 1
7 1
8 1
9 2
10 2
11 2
12 3
13 3
14 3
15 3
16 4
17 4
18 4
19 5
20 5
21 5
22 5
23 6
24 6
25 6
26 7
27 7
28 7
29 8
30 8
};
\addlegendentry{$C=2$ (2 Attackers)}
\addplot [semithick, color2]
table {%
5 nan
6 nan
7 nan
8 1
9 1
10 1
11 1
12 2
13 2
14 2
15 2
16 2
17 3
18 3
19 3
20 3
21 3
22 4
23 4
24 4
25 4
26 4
27 5
28 5
29 5
30 5
};
\addlegendentry{$C=3$ (3 Attackers)}
\end{axis}

\end{tikzpicture}

%% file: a5_detect.tex
\section{Discussion: Detecting Malicious Messages with Ablation}
\label{app:exp_detect}

As discussed in Appendix~\ref{app:hyper}, to defend against a larger $C$, one has to choose a relatively small $k$ for certifiable performance. However, Figure \ref{fig:res_hyper} suggests that a small $k$ also sacrifices the natural performance of the agent to obtain higher robustness. This is known as the trade-off between robustness and accuracy~\cite{tsipras2018robustness,zhang2019theoretically}. Can we achieve better robustness while not sacrificing much natural performance, or obtain higher natural reward while not losing robustness?

We point out that with our proposed \ours defense, it is possible to choose a larger $k$ than what is required by Condition~\ref{cond:k} without sacrificing robust performance by identifying the malicious messages beforehand. 
The idea is to detect the adversarial messages and to rule out them before message ablation and ensemble, during the test time.

\begin{figure}[h]
 \centering
\input{figs/Attacker_Identification.tex}
\caption{Attacker identification based on action bias $\nu_j$. Adv1 and Adv2 stands for two agents hacked by the attacker. B1 up to B6 stands for six other benign agents.}
\label{fig:detect_adv}
\end{figure}
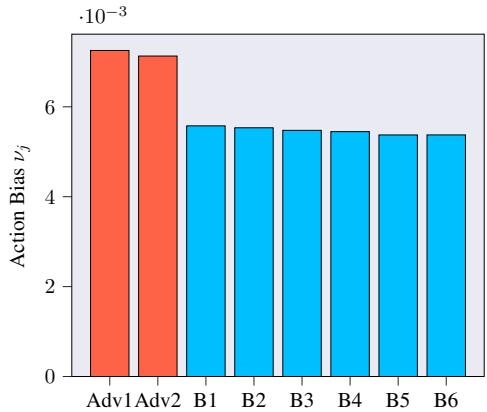

We hypothesize that given a well-trained victim policy, malicious messages tend to mislead the victim agent to take an action that is "far away" from a "good" action that the victim is supposed to take. 
To verify this hypothesis, we first train a \ablname $\ablpolicy_i$ with $k=1$ for agent $i$. Then for every communication message that another agent $j$ sends, we compute the action $a_j$ that the victim policy $\ablpolicy_i$ chooses based on all message subsets containing message $m_{j\to i}$, i.e. $a_j = \ablpolicy(\tau_i, m_{j\to i})$ (note that $k=1$ so $\ablpolicy$ only takes in one message at a time). We then define the \emph{action bias} as $\beta_j=\|a_j-\funmed\{a_k\}_{k=1}^{N}\|_1$. Based on our hypothesis, an agent which has been hacked by attackers should induce a significantly larger action bias since they are trying to mislead the victim to take a completely different action. Here, we execute the policy of two hacked agents together with six other good agents for twenty episodes and calculate the average action bias for each agent.
As shown in in Figure \ref{fig:detect_adv}, the agents hacked by attackers indeed induce a larger action bias compared to other benign agents, which suggests the effectiveness of identifying the malicious messages by action bias.

After filtering out $c$ messages, one could compute the required $k$ by Equation~\eqref{eq:k_cond} based on $C-c$ malicious messages and $N-1-c$ total messages, which can be larger than the largest $k$ induced by $C$ malicious messages out of $N-1$ total messages. For example, if $N=30$, $C=3$, then the largest $k$ satisfying Equation~\eqref{eq:k_cond} is 5, but when 1 adversarial message is filtered out, the largest $k$ that can defend against the remaining 2 adversarial messages is 8. Although the adversary identification is not theoretically guaranteed to be accurate, Figure \ref{fig:detect_adv} demonstrates the effectiveness of the adversarial message detection, which, combined with \ours with larger $k$'s, has the potential to achieve high natural performance and strong robustness in practice.

%% file: figs/Attacker_Identification.tex
\begin{tikzpicture}[scale=0.8]

\definecolor{color0}{rgb}{0.917647058823529,0.917647058823529,0.949019607843137}
\definecolor{color1}{rgb}{1,0.388235294117647,0.27843137254902}
\definecolor{color2}{rgb}{0,0.749019607843137,1}

\begin{axis}[
axis background/.style={fill=color0},
axis line style={black},
tick align=outside,
x grid style={white},
xmajorticks=true,
xmin=-0.79, xmax=7.79,
xtick style={color=white!15!black},
xtick={0,1,2,3,4,5,6,7},
xtick pos = lower,
xticklabels={Adv1, Adv2, B1, B2, B3, B4, B5, B6},
y grid style={white},
ylabel={Action Bias $\nu_j$},
ymajorticks=true,
ymin=0, ymax=0.007615881,
ytick style={color=white!15!black},
ytick pos = left
]
\draw[draw=black,fill=color1] (axis cs:-0.4,0) rectangle (axis cs:0.4,0.00725322);
\draw[draw=black,fill=color1] (axis cs:0.6,0) rectangle (axis cs:1.4,0.00712916);
\draw[draw=black,fill=color2] (axis cs:1.6,0) rectangle (axis cs:2.4,0.00557617);
\draw[draw=black,fill=color2] (axis cs:2.6,0) rectangle (axis cs:3.4,0.00553338);
\draw[draw=black,fill=color2] (axis cs:3.6,0) rectangle (axis cs:4.4,0.00547642);
\draw[draw=black,fill=color2] (axis cs:4.6,0) rectangle (axis cs:5.4,0.00544767);
\draw[draw=black,fill=color2] (axis cs:5.6,0) rectangle (axis cs:6.4,0.00537401);
\draw[draw=black,fill=color2] (axis cs:6.6,0) rectangle (axis cs:7.4,0.00537477);
\end{axis}

\end{tikzpicture}

%% file: neurips/a8_impact.tex
\section{Additional Discussion on Societal Impacts and Limitations}
\label{app:discuss}

This paper proposes a defense algorithm for multi-agent sequential decision making problems. The proposed algorithm, \ours, is certifiably robust during test-time under communication attacks that perturbs up to $C$ out of $N-1$ messages. Since our goal is to improve the robustness of any decision-making agent against corrupted communication, our algorithm will benefit many real-world applications. Therefore, the societal impacts of this work will be positive.

The method discussed in this paper is both theoretically grounded and empirically effective. But there are still some limitations. For example, the theoretical guarantee requires some conditions that may be broken in some challenging environments, as discussed in Appendix~\ref{app:hyper}, although the algorithm still performs well without these conditions in practice. Our future work includes relaxing these assumptions and conditions and obtain more universal certificates.
Moreover, we focus on the case where communication messages from different agents are symmetric, such that the \ablname can make decisions based on randomly sampled $k$ messages without considering the order of messages. But our algorithm can be extended to cover various message spaces (i.e., different agents may send messages with different formats).